\documentclass{article}

\usepackage{microtype}
\usepackage{colortbl}  %彩色表格需要加载的宏包
\usepackage{xcolor}
\usepackage{graphicx}
\usepackage{subfigure}
\usepackage{booktabs} % for professional tables

\usepackage{hyperref}
\usepackage[accepted]{icml2024}

\usepackage{float}
\usepackage{amsmath}
\usepackage{amssymb}
\usepackage{mathtools}
\usepackage{amsthm}

\usepackage[capitalize,noabbrev]{cleveref}

\theoremstyle{plain}

\theoremstyle{definition}

\theoremstyle{remark}

\usepackage[textsize=tiny]{todonotes}

\icmltitlerunning{DPN: \underline{D}ecoupling \underline{P}artition and \underline{N}avigation for Neural Solvers of Min-max Vehicle Routing Problems}

\begin{document}

\twocolumn[
\icmltitle{DPN: \underline{D}ecoupling \underline{P}artition and \underline{N}avigation for Neural Solvers \\of Min-max Vehicle Routing Problems}

\icmlsetsymbol{equal}{*}

\begin{icmlauthorlist}
\icmlauthor{Zhi Zheng}{equal,sustech}
\icmlauthor{Shunyu Yao}{equal,cityu,sustech-sdim}
\icmlauthor{Zhenkun Wang}{sustech-sdim}
\icmlauthor{Xialiang Tong}{hw}
\icmlauthor{Mingxuan Yuan}{hw}
\icmlauthor{Ke Tang}{sustech}

\end{icmlauthorlist}

\icmlaffiliation{sustech-sdim}{School of System Design and Intelligent Manufacturing, Southern University of Science and Technology, Shenzhen, China}
\icmlaffiliation{sustech}{Department of Computer Science and Engineering, Southern University of Science and Technology, Shenzhen, China}
\icmlaffiliation{cityu}{Department of Computer Science, City University of Hong Kong, Hong Kong SAR.}
\icmlaffiliation{hw}{China Huawei Noah’s Ark Lab, Shenzhen, China.}
\icmlcorrespondingauthor{Zhenkun Wang}{wangzhenkun90@gmail.com}

\icmlkeywords{Neural Combinatorial Optimization, Vehicle Routing Problems, Min-max Vehicle Routing Problems, Reinforcement Learning}

\vskip 0.3in
]

\printAffiliationsAndNotice{\icmlEqualContribution} % otherwise use the standard text.

\begin{abstract}
The min-max vehicle routing problem (min-max VRP) traverses all given customers by assigning several routes and aims to minimize the length of the longest route. Recently, reinforcement learning (RL)-based sequential planning methods have exhibited advantages in solving efficiency and optimality. However, these methods fail to exploit the problem-specific properties in learning representations, resulting in less effective features for decoding optimal routes. This paper considers the sequential planning process of min-max VRPs as two coupled optimization tasks: customer partition for different routes and customer navigation in each route (i.e., partition and navigation). To effectively process min-max VRP instances, we present a novel attention-based Partition-and-Navigation encoder (P\&N Encoder) that learns distinct embeddings for partition and navigation. Furthermore, we utilize an inherent symmetry in decoding routes and develop an effective agent-permutation-symmetric (APS) loss function. Experimental results demonstrate that the proposed Decoupling-Partition-Navigation (DPN) method significantly surpasses existing learning-based methods in both single-depot and multi-depot min-max VRPs. Our code is available at \footnote{\url{https://github.com/CIAM-Group/NCO_code/tree/main/single_objective/DPN-minmaxVRP-master}}.
\end{abstract}

\section{Introduction}
The vehicle routing problem (VRP) aims to determine a set of routes that satisfies the demand of all customers by minimizing a distance-based objective function \citep{toth2002overview, toth2014vehicle}. As a variant of VRP, the min-max vehicle routing problem (min-max VRP), instead of minimizing the total length of all routes (i.e., min-sum), seeks to reduce the length of the longest one among all the routes (i.e., min-max). Many real-world applications, such as transportation planning \citep{delgado2001minmax}, disaster management \citep{cheikhrouhou2021comprehensive}, and robotics \citep{faigl2016application,david2021multi} are more consistent with the min-max VRP due to their operational requirements. Although the min-max VRP is of great significance and has attracted widespread attention, solving it is still very challenging \citep{kumar2012survey}. The min-max VRP emphasizes balancing the lengths between different routes, thereby it considers not only the order of customers within each route but also the partition of customers \citep{francca1995m}. Efficient solvers should simultaneously address both a customer partition task and a navigation task for customers assigned to each route (i.e., partition and navigation tasks) \citep{arkin2006approximations,carlsson2009solving,narasimha2013ant,vandermeulen2019balanced}. 

%\sy{The min-max VRP emphasizes balancing the lengths of different routes which requires the consideration of both the division of customers among routes (i.e., partition) and the order of customers within each route (i.e., navigation) \citep{francca1995m}. Therefore, efficient solvers should address these two aspects simultaneously \citep{carlsson2009solving,vandermeulen2019balanced}.}

\begin{figure*}[htbp]
    \centering
    \subfigure[Four examples of min-max VRPs]{\label{four}\includegraphics[width = 0.315\hsize]{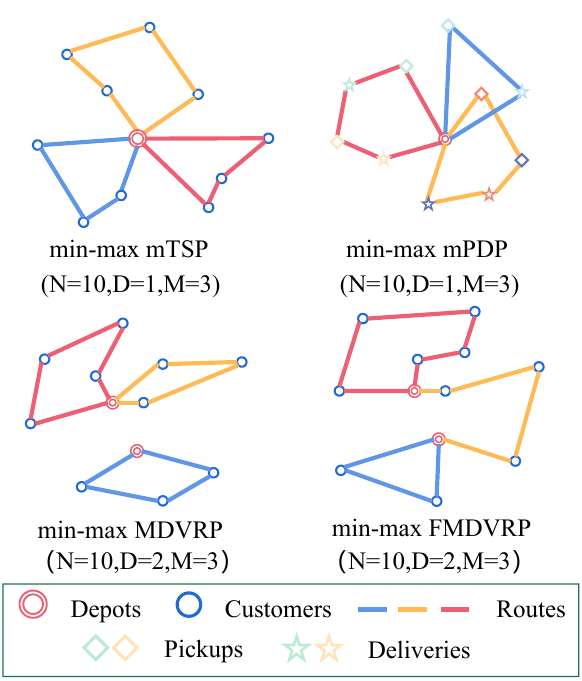}}
    \subfigure[Existing framework]{\label{eq}\includegraphics[width = 0.2928\hsize]{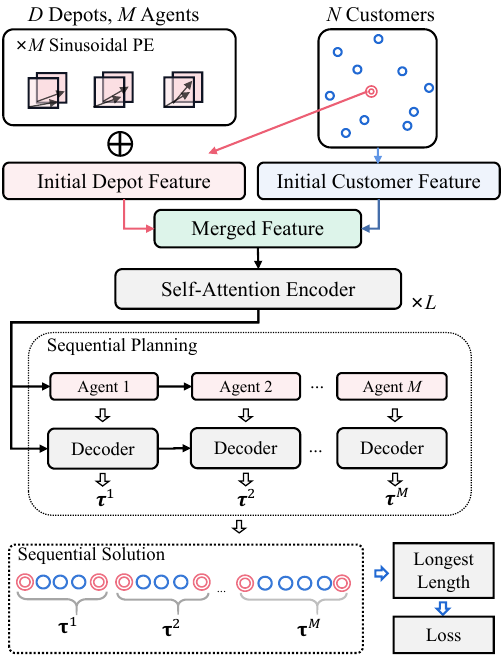}}
    \subfigure[DPN (Ours)]{\label{dpn-small}\includegraphics[width = 0.380\hsize]{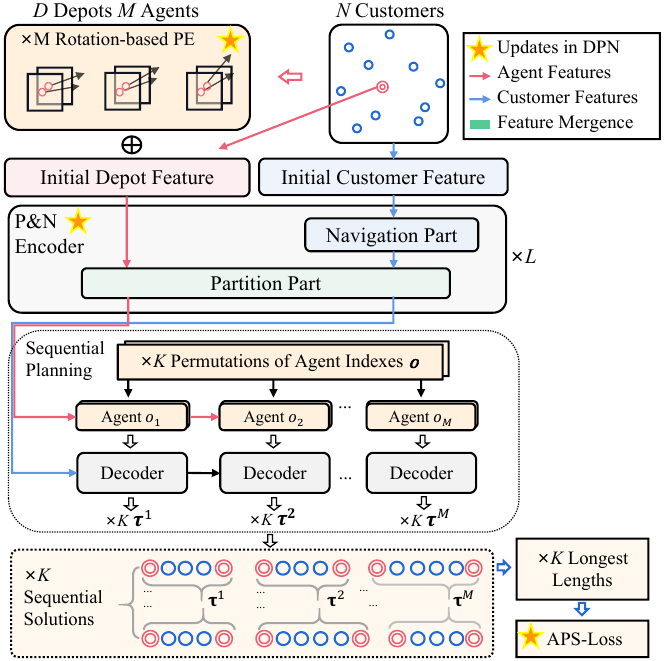}}
     \caption{(a) The instances and solutions of four involved min-max VRPs. (b) The sequential planning framework proposed in \citet{son2023solving}. (c) The sequential planning framework of the proposed DPN. To decouple the partition and navigation features and improve the representation ability, we conduct modifications to the existing framework by proposing a novel P\&N Encoder, utilizing agent-permutation-symmetries (APS) in loss calculation, and introducing a Roataion-based positional encoding (Rotation-based PE) for agent representations.}
    \label{fig:example}
\end{figure*}

% \sy{Recently, reinforcement learning (RL)-based neural solvers have shown promising performance in terms of efficiency and optimality on classical min-sum VRPs \citep{bello2017neural, nazari2018reinforcement, pan2023h, gao2023towards}. Many learning-based methods are developed to handle some min-max VRP variants, \textcolor{red}{e.g., the min-max multi-agent traveling salesman problem (min-max mTSP),} including two-stage methods \citep{kaempfer2018learning,hu2020reinforcement,liang2023splitnet}, learning improvement heuristics \citep{kim2022learning}, parallel planning methods \citep{cao2021dan,gao2023amarl,park2021schedulenet}, and sequential planning methods \citep{son2023solving}. Two-stage methods and learning improvement heuristics rely on the handcrafted heuristic operators. Parallel planning methods employ multiple decentralized models to construct the set of routes in parallel, in which the complexity of the joint action space hinders optimization \textcolor{red}{(training efficiency or what?)}. Sequential planning methods, on the other hand, sequentially construct the set of routes with a single model. \textcolor{red}{This approach reduces the decision space complexity, facilitating easier convergence \citep{son2023solving}}.}
Recently, reinforcement learning (RL)-based neural solvers have been successfully applied to classical min-sum VRPs such as the traveling salesman problem (TSP) \citep{bello2017neural} and the capacitated vehicle routing problem (CVRP) \citep{nazari2018reinforcement}. These solvers show competitive performances in efficiency and accuracy compared to traditional heuristic methods \citep{pan2023h,gao2023towards}. Meanwhile, some learning-based methods have emerged to handle some min-max VRP variants. These methods include two-stage methods \citep{kaempfer2018learning,hu2020reinforcement,liang2023splitnet}, learning improvement heuristics \citep{kim2022learning}, parallel planning methods \citep{cao2021dan,park2021schedulenet,gao2023amarl}, and sequential planning methods \citep{son2023solving}. Two-stage methods and learning improvement heuristics highly rely on handcrafted heuristic algorithms \citep{liang2023splitnet}. Parallel planning methods employ multiple decentralized models to construct the set of routes in parallel, which derives a complex joint action space and causes difficulties in training \citep{son2023solving}. Sequential planning methods, on the other hand, sequentially construct the set of routes with a single model. This approach significantly reduces the decision space complexity in training, facilitating the exploration of the optimal solution.

% 写法改动2：1.为什么并行不行，需要solid; 2.sequential方法的缺陷，不明显提出Equity-Transformer
% Equity-Transformer \citep{son2023solving} is the state-of-the-art sequential planning method. %describe Equity-Transformer,有重心的去介绍它。以便后面引出motivation。
% Existing methods often ignore the following valuable properties of min-max VRPs: 1) the partition and navigation should be considered separately. 2) the optimality of the solution independent from the order of the set of routes. 3) the depot has a huge impact on the results, especially generalization.

As shown in Figure \ref{eq}, existing sequential planning methods process min-max VRP instances with an encoder-decoder model structure and utilize multiple agent embeddings to construct the routes one by one. However, as specialized solvers of min-max VRPs, existing methods fail to leverage problem-specific properties, potentially impairing their representation ability. Firstly, directly processing the merged feature through several self-attention-based encoder layers leads to ambiguous representations for customers and agents. This ambiguity blurs the distinction between the partition and navigation tasks, potentially resulting in routes with imbalanced lengths. Moreover, the decoding strategy that employs agents to generate routes in a fixed order (i.e., from the first agent to the last one) could potentially trap the solver in local minima. Lastly, the adopted sinusoidal positional encoding (PE) is crucial for learning partition, but excluding the depot coordinates in its formula might weaken the model's generalization ability across different depot locations.

This paper aims to fully exploit the problem-specific properties of the min-max VRP, particularly the requirements of partition and navigation. To preserve independent representations for partition and navigation, we propose the Decoupling-Partition-Navigation (DPN) method. As shown in Figure \ref{dpn-small}, it adopts a novel attention-based Partition-and-Navigation encoder (P\&N Encoder) consisting of a navigation part and a partition part in each layer. Moreover, we propose an agent-permutation-symmetric (APS) loss function, leveraging the equivalence of optima to improve training efficiency. Beyond that, we use a depot-location-aware Rotation-based PE to enhance partition-related representations. We conduct experiments on the four min-max VRPs exhibited in Figure \ref{four}, i.e., min-max multi-agent traveling salesman problem (min-max mTSP), min-max multi-agent pickup and delivery problem (min-max mPDP), min-max multi-depot VRP (min-max MDVRP), and min-max flexible multi-depot VRP (min-max FMDVRP). Experimental results indicate that DPN can significantly outperform the other neural solvers on all four min-max VRPs. 

The contributions of this paper can be summarized as follows: \textbf{1):} We propose an RL-based sequential planning method for min-max VRPs that utilizes a novel attention-based P\&N Encoder to process decoupled representations of the partition and navigation tasks. \textbf{2):} The proposed DPN leverages symmetries in the permutation of agents and presents a novel loss function that speeds up the convergence in the RL training. \textbf{3):} Experiments demonstrate that the proposed DPN can significantly outperform existing neural solvers on four representative min-max VRPs.

\section{Preliminaries}

\subsection{Sequential Planning for Solving Min-max VRPs}\label{sequantialplanning}

A min-max VRP instance with $M$ routes (generated by $M$ agents), $D$ depots, and $N$ customers is defined over a graph $\mathcal{G}=\{\mathcal{V},\mathcal{E}\}$. Each $v_i\in \mathcal{V}$ represents a depot or a customer and $e_{ij}\in \mathcal{E}$ represents the edge between node $v_i$ and $v_j$. The solution (i.e., a set of routes) $\mathcal{T}$ is formed by node indexes in $\mathcal{V}$, and each customer in $\mathcal{V}$ can only get visited once. Moreover, the number of routes in solution $\mathcal{T}$ is restricted to $M$ (i.e., $\mathcal{T} = \{\boldsymbol{\tau}^1,\ldots,\boldsymbol{\tau}^M\}$). Each route $\boldsymbol{\tau}^i$ for $i\in\{1,\ldots, M\}$ only starts and ends at a depot. The objective function of the min-max VRP can be formulated as
\begin{equation}
\begin{aligned}
    &\underset{\mathcal{T}\in {\Omega}}{\mbox{minimize}}\quad f(\mathcal{T})=\underset{i\in\{1,\ldots,M\}}{\max}L(\boldsymbol{\tau}^i),
\end{aligned}
\label{ob1}
\end{equation}
where ${\Omega}$ is a set consisting of all feasible solutions, and $L(\boldsymbol{\tau}^i)$ calculates the Euclidean length of route $\boldsymbol{\tau}^i$.

Unlike the parallel planning method that simultaneously constructs the $M$ routes, the sequential planning methods generate the $M$ routes one by one. During the construction of each route, the sequential planning model gradually selects the next customer to extend the route and finally returns to the depot. The sequential planning process can be formalized as a Markov Decision Process (MDP) $\mathcal{M}=\{\mathcal{S},\mathcal{A},\boldsymbol{r},\mathcal{P}\}$ \citep{son2023solving}. At the $t$-th step, the action $a_t\in \mathcal{A}$ represents the index of the selected node from unvisited customers or the set of depots, and the state $s_t\in \mathcal{S}$ records the partial solution currently constructed, the instance $\mathcal{G}$, and the number of agents $M$. The terminal state $s_T$ contains a feasible solution $\mathcal{T}\in {\Omega}$ with a min-max reward $r_T=-f(\mathcal{T})$ formulated in Eq. \eqref{ob1}. Parameterized by $\theta$, the policy $p_\theta$ for sampling the set of routes $\tau$ can be calculated as 
\begin{align}
\!\! p_\theta(\mathcal{T}|\mathcal{G},M)=\prod_{t=1}^{T-1}p_\theta(a_t|s_t)=\prod_{i=1}^{M}p(\boldsymbol{\tau}^i|\boldsymbol{\tau}^{1:i},\mathcal{G},\theta),
\label{sq1}
\end{align}
where $\boldsymbol{\tau}^{1:i}$ represents all the routes generated before $\boldsymbol{\tau}^i$. More details of the MDP are presented in Appendix \ref{sequential} and a comprehensive literature review about neural solvers is in Appendix \ref{related}.

\begin{figure*}
    \centering
    \includegraphics[width = \hsize]{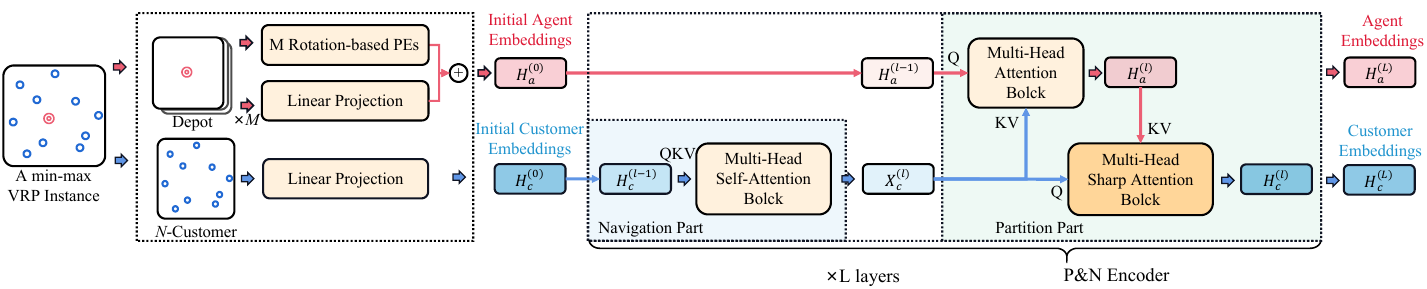}
    \caption{Initial embeddings and the proposed P\&N Encoder.}
    \label{fig:net}
\end{figure*}

\subsection{Partition and Navigation in Min-max VRPs}
The tasks of assigning customers to $M$ routes (i.e., partition) and optimizing the routing of customers assigned to each route (i.e., navigation) are considered simultaneously in min-max solvers \citep{carlsson2009solving,narasimha2013ant}. These two tasks can be defined as follows:

\textbf{Partition.} Customers and depots assigned to the route $\boldsymbol{\tau}^i$ for $i\in\{1,\ldots, M\}$ forms a partition of $\mathcal{G}$. Each sub-graph is denoted as $\mathcal{G}^i$. The partition function $P_{\theta,M}(\mathcal{G})=\{\mathcal{G}^1,\ldots,\mathcal{G}^M\}$ with parameter $\theta$ generates sub-graph partitions with
\begin{align}
    \bigcap_{i\in\{1,\ldots, M\}}\mathcal{G}^i\ \subseteq\  \text{Depots},\bigcup_{i\in\{1,\ldots, M\}}\mathcal{G}^i\ =\ \mathcal{G}.
\label{partition}
\end{align}

\textbf{Navigation.} The navigation task (generally being TSP) optimizes routings in each sub-graph $\mathcal{G}^i$ for $i\in\{1,\ldots, M\}$. For sequential planning methods, if the number of nodes in $\mathcal{G}^i$ is $n^i$, the navigation policy $\pi_\theta$ of $\mathcal{G}^i$ can written as follows:
%\sy{Notation. }
\begin{align}
\pi_\theta(\boldsymbol{\tau^i}|\mathcal{G}^i)=\prod_{t=1}^{n^i}p(\boldsymbol{\tau}^i(t)|\boldsymbol{\tau}^i(1:t),\mathcal{G}^i,\theta),\label{navigation-policy}
\end{align}
where $\boldsymbol{\tau}^i(t)$ is the t-th node index in $\boldsymbol{\tau}^i$ and $\boldsymbol{\tau}^i(1:t)$ represents indexes of nodes selected before $\boldsymbol{\tau}^i(t)$. Integrating the partition and navigation policy into Eq. \eqref{ob1}, the optimal parameter $\theta^*$ can be obtained equivalently as follows:
\begin{align}
    \theta^*=\underset{\theta}{\arg\min} \underset{G^i\in P_{\theta,M}(\mathcal{G})}{\max}\mathbb{E}_{\pi_\theta(\boldsymbol{\tau^i}|\mathcal{G}^i)}\big[L(\boldsymbol{\tau}^i)\big].
\end{align}

\subsection{Transformer Blocks}

The Transformer block \citep{vaswani2017attention} is widely adopted in sequential planning models. It contains a multi-head-attention function and a feed-forward function.

\textbf{Attention Mechanism.} The attention mechanism is a fundamental component in deep learning models \citep{niu2021review}. The classical attention function allows two embeddings to share information, and it can be expressed as
\begin{equation}
\begin{aligned}
    Y = \text{Attn}(X,C) = \text{Softmax}\left( \frac{ X W_Q (C W_K)^{\intercal}}{\sqrt{d}}\right)C W_V,
\end{aligned}
\label{attn}
\end{equation}
where $X \in R^{n\times d}$ and $C \in R^{m\times d}$ are inputted embeddings. $W_Q, W_K \in R^{d\times d_k}$, and $ W_V \in R^{d\times d_v}$ are trainable query, key, and value parameters, respectively. The  Softmax function is applied independently across each row, and the output of the attention function is a matrix $Y \in R^{n\times d}$. As a special case, the self-attention function calculates the attention function with two same inputs (i.e., Attn$(X', X')$). For more efficiency, the multi-head attention function applies the attention function in Eq. \eqref{attn} on $H$ different ``heads'' with independent parameters, i.e.,
\begin{equation}
\begin{aligned}
    &\text{MHA}(X,C)=\text{Concat}(Y_1,\ldots,Y_H)W_p,\\
    &\quad \text{where}\quad Y_i=\text{Attn}_i(X,C), \forall i\in\{1,\ldots,H\},
\end{aligned}
\label{mha}
\end{equation}
where $d_k=d_v= \frac{d}{H}$ in each $\text{Attn}_i$. $W_p\in R^{d\times d}$ denotes a trainable square projection matrix that combines different attention heads. The transformer blocks \citep{vaswani2017attention,luo2023neural} also include the feed-forward function, i.e.,
\begin{align}
    \quad \text{FF}(X)=\left(\text{ReLU}(XW_1)\right)W_2,\label{ff}
\end{align}
where FF and ReLU represent the feed-forward function and the ReLU activation function \citep{nair2010rectified}, respectively. $W_1 \in R^{d\times d_h}$, and $ W_2 \in R^{d_h\times d}$ are trainable projections and $d_h$ represents the hidden space dimension.

\section{Methodology}

As shown in Figure \ref{dpn-small}, DPN adopts the general framework of the sequential planning method with a multi-layer encoder and a single-layer decoder. The $L$-layer encoder processes the initial embeddings of customer coordinates to customer embeddings and encodes the initial embeddings of depot coordinates and the $M$-route constraint (represented by PEs) to agent embeddings. By handling these embeddings and a contextual representation of the MDP state $s_t$, the decoder is repeatedly employed to generate actions. The proposed DPN integrates problem-specific properties in three key components of the original sequential planning methods. Firstly, DPN presents a bi-part attention-based P\&N Encoder, which utilizes separate structures to capture decoupled features for partition and navigation. Secondly, we develop a novel agent-permutation-symmetric loss (APS-Loss) function that leverages symmetries in decoding routes for effective explorations in customer partition. Finally, we adopt the Rotation-based PE to incorporate the coordinates of depots into the PE calculation to achieve superior representations for customer partition. 

The proposed DPN can efficiently address both single-depot and multi-depot min-max VRPs. For clarity, this Section only illustrates the details of DPN in solving the single-depot ones (e.g., the min-max mTSP). The P\&N Encoder for multi-depot min-max VRPs can be found in Section \ref{PN2}, with additional implementation details, motivations, and necessity statements provided in Appendix \ref{Methods2}.

\subsection{P\&N Encoder}

Figure \ref{fig:net} depicts the input and architecture of the proposed P\&N Encoder. Linear projections and several PEs convert an instance into initial embeddings, and each layer of the P\&N Encoder comprises two main components: a navigation part and a partition part.

\textbf{Initial Embeddings.} A single-depot instance with $N$ customers and $M$ routes (i.e., number of agents) is mapped to initial agent embeddings $H_a^{(0)}\in \mathcal{R}^{M\times d}$ and initial customer embeddings $H_c^{(0)}\in \mathcal{R}^{N\times d}$ before being fed to the P\&N Encoder. In processing these embeddings, the coordinate of the depot $\boldsymbol{x}_d\in\mathcal{R}^{1 \times 2}$ and the coordinates of customers $\boldsymbol{x}\in\mathcal{R}^{N\times 2}$ are embedded into $H_a^{(0)}$ and $H_c^{(0)}$ via linear projections. $M$ various $d$-dimensional PEs are also added in $H_a^{(0)}$ to distinguish the multiple agents.

\textbf{Motivation.} As Figure \ref{eq}, the previous sequential planning methods adopt multi-head self-attention to process the merged features of agents and customers (i.e., compute multi-head self-attention to $\text{Concat}[H_a^{(0)}, H_c^{(0)}]$). However, as further discussed in Appendix \ref{modelstructure}, these calculations can be decomposed into four distinct kinds of relations, and the Softmax function in self-attention functions directly normalizes the customer-customer relations and the agent-customer relations, potentially confusing the heterogeneous representations for partition and navigation. To achieve decoupled representations, the proposed P\&N Encoder structurally separates the representation encoding process of these two tasks into navigation and partition parts, respectively. In the navigation part, customer embeddings conduct self-attention for navigation features. In the partition part, two attention blocks are calculated between agents and customers to enhance representations for the customer partition task.

\textbf{Navigation Part.} The partition part in the $l$-th layer P\&N Encoder calculates customer embeddings $H_c^{(l-1)}$ independently by a multi-head self-attention block. This part focuses on customer routing based on $H_c^{(l-1)}$ and excludes the impact of learning partitions. It consists of an MHA function, a feed-forward function (FF), and two skip connections with trainable parameters $\alpha_1$ and $\alpha_2$ (Rezero normalization \citep{bachlechner2021rezero}), i.e.,
\begin{equation}
\begin{aligned}
    &\hat{X}_c^{(l)}=\alpha_1*\text{MHA}(H_c^{(l-1)},H_c^{(l-1)}))+H_c^{(l-1)},\\
    &X_c^{(l)}=\alpha_2*\text{FF}(\hat{X}_c^{(l)})\quad\quad\quad\quad\quad\quad+\hat{X}_c^{(l)}.
\end{aligned}
\end{equation}
\textbf{Partition Part.} The following partition part is designed to process representations related to the customer partition task. Agent features $H_a^{(l-1)}$ and customer features from the navigation part $X_c^{(l)}$ are fused through two cascaded attention blocks. In this part, agents are supposed to integrate features from their assigned customers (i.e., sub-graph $\mathcal{G}_i$ for $i$-th agent), while customers can derive representations about their sub-graph assignments as well. Moreover, according to Eq. \eqref{partition}, each customer can only be assigned to one sub-graph, so each customer should only integrate the embeddings of one among the $M$ agents. To meet this requirement, we remove the Softmax temperature in the Attn function (Eq. \eqref{attn}) for fast convergence and design a multi-head sharp attention function (denoted as MHSA) as
\begin{equation}
\begin{aligned}
    \text{MHSA}(X,C)&=\text{Concat}(S_1,\ldots,S_H)W_P,\\
     \text{where} \quad  \quad S_i&= \text{Softmax}\left(X W_Q (C W_K)^{\intercal}\right)CW_V.
\end{aligned}
\label{sharp-attn}
\end{equation}
%Multi-head sharp-attention has also been used for entity recognition \citep{yan2019tener}.
The total function of the partition part employs four residual links with parameters $\alpha_3,\ldots,\alpha_6$, i.e.,
\begin{align}
    &\hat{H}_a^{(l)}=\alpha_3*\text{MHA}(H_a^{(l-1)},X_c^{(l)})+H_a^{(l-1)},\label{place}\\
    &H_a^{(l)}=\alpha_4*\text{FF}(\hat{H}_a^{(l)})\quad\quad\quad\quad\ \ +\hat{H}_a^{(l)},\\
    &\hat{H}_c^{(l)}=\alpha_5*\text{MHSA}(X_c^{(l)},H_a^{(l)})\ \  +X_c^{(l)},\label{place2}\\
    &H_c^{(l)}=\alpha_6*\text{FF}(\hat{H}_c^{(l)})\quad\quad\quad\quad\ \ +\hat{H}_c^{(l)}.
\end{align}

\subsection{APS-Loss}

Leveraging symmetry can effectively promote the training process of neural combination optimization solvers \citep{kwon2020pomo,kwon2021matrix,kim2022sym}. Correspondingly, as another important contribution, DPN develops a novel agent-permutation-symmetry (APS) in solving min-max VRPs. %which is formalized similarly to the definition of problem symmetries in \citet{kim2022sym}.
For an instance $\mathcal{G}$, and an index order of agents $\boldsymbol{o}=(o_1,o_2,\ldots,o_M)$ (i.e., permutation) for generating the $M$ routes in solution $\mathcal{T}$, the APS maintains an equivalence optimal objective functions for any permutations $\boldsymbol{o}$. The APS holds because, for all the $\boldsymbol{o}=(o_1,\ldots,o_M)$, and $\{\mathcal{G}^{i}|i\in\{1,\ldots,M\}\}$ being optimal sub-graph partitions of $\mathcal{G}$, the set of routes in the optimal solution $\mathcal{T^*}= \{\boldsymbol{\tau^*}^i\sim \pi_\theta(\cdot|\mathcal{G}^{o_i})$ for $\ i \in \{1\ldots, M\}\}$ keeps unordered. Hence, the objective function (i.e., the longest route length) is identical for any permutations. $\pi_\theta(\cdot|\mathcal{G}^{o_i})$ represents the policy derived from Eq. \eqref{navigation-policy}.
%and features of $\mathcal{G}^{o_i}$ are represented by the $i$-th agent embeddings.

As shown in Figure \ref{dpn-small}, DPN leverages the APS for a better baseline with $K$ sampled permutations $\boldsymbol{o}^{(1)}$ to $\boldsymbol{o}^{(K)}$. The APS-equipped baseline $b(\mathcal{G})$ is the average objective function of the $K$ generated solutions estimated as
\begin{equation}
\begin{aligned}
b(\mathcal{G})=\frac{1}{K}\sum_{k=1}^K\underset{i\in\{1,\ldots,M\}}{\max}L(\boldsymbol{\tau}^i\sim p_\theta(\cdot|\mathcal{G}^{o^{(k)}_i})).
\end{aligned}
\end{equation}
APS loss function $\mathcal{L}(\mathcal{G})$ (i.e., APS-Loss) is processed by the REINFORCE \citep{williams1992simple} algorithm as:
\begin{align}
\pi'_\theta(\mathcal{T}|\mathcal{G},M&,\boldsymbol{o}^{(k)})=\prod_{i=1}^{M}\pi_\theta(\boldsymbol{\tau}^{i}|\mathcal{G}^{o^{(k)}_i}),\label{total-loss}\\
\nabla_\theta\mathcal{L}(\mathcal{G})=&-\frac{1}{K}\sum_{k=1}^K\mathbb{E}_{\pi'_\theta(\mathcal{T}|\mathcal{G},M,\boldsymbol{o^{(k)}})}\\
&\Big[(f(\mathcal{T})-b(\mathcal{G}))\nabla_\theta\text{log}\ \pi'_\theta(\mathcal{T}|\mathcal{G},M,\boldsymbol{o^{(k)}}) \Big],\notag
\end{align}
where features of the sub-graph $\mathcal{G}^{o^{(k)}_i}$ are supposed to be represented in the $o^{(k)}_i$-th agent embedding in DPN. The APS-Loss can promote explorations in learning representations and alleviate the bias in embeddings caused by predetermined decoding orders.

\subsection{Rotation-based PE}

In sequential planning methods \citep{son2023solving}, different agents starting from the same depot are distinguished by positional encodings. Existing sequential planning methods adopt the sinusoidal PE \citep{vaswani2017attention} for agent distinction. The $d$-dimensional sinusoidal PE (denoted as SPE $\in \mathcal{R}^{M\times d}$) in the case is defined as
\begin{equation}
\begin{aligned}
    \text{SPE}(m,q)&=\left\{
    \begin{aligned}
    & \sin(m/10,000 ^{\frac{\lfloor q/2\rfloor}{d}}),q\equiv 0(mod\ 2)\\
    & \cos(m/10,000 ^{\frac{\lfloor q/2\rfloor}{d}}),q\equiv 1(mod\ 2)
    \end{aligned}
    \right..
\end{aligned}
\end{equation}
In Appendix \ref{PEmotivation}, we illustrate how a PE conveys the angle-based information when representing the sub-graphs in the partition part of the P\&N Encoder. The angle-based interval between sub-graphs is highly correlated with the depot location \citep{francca1995m}, so SPE excluding the depot coordinates will harm the partition optimality. To introduce the depot coordinates to PEs, we refer to the concept of Rotary PE \citep{su2024roformer} and design the Rotation-based PE $\in \mathcal{R}^{M\times d}$ for a depot-aware agent distinction as 
\begin{equation}
\begin{aligned}
    \text{PE}(m,q)&=\text{Re}\big[(\boldsymbol{x}_dW_{a}+b_{a})\exp(im/1,000^{\frac{\lfloor q/2\rfloor}{d}})\big],
\end{aligned}
\end{equation}
where $i$ is an imaginary unit (i.e., $i^2=-1$) and $\boldsymbol{x}_dW_{a}+b_{a}$ is a trainable linear projection of the depot coordinate $\boldsymbol{x}_d\in \mathcal{R}^2$. PE is calculated in the complex vector space, and Re represents the real part of a complex number. Additionally, Appendix \ref{implementationpe} provides an intuitive way of calculating PE.

\begin{table*}[htbp]
\caption{The average objective function values(i.e., Obj.), gaps to the best algorithm (i.e., Gap), and testing times (i.e., Time) obtained by each algorithm on 12 datasets. Each set contains 100 randomly generated medium-scale instances. The overall best results are highlighted in bold, and the algorithm that produces the best results among all five learning-based results is highlighted in a gray background.}
\centering
\resizebox{\textwidth}{!}{
\renewcommand\arraystretch{1}
\setlength{\tabcolsep}{3.5mm}
\small{
\begin{tabular}{lcclcclccl}
\toprule[0.5mm]
\multicolumn{10}{c}{Min-max mTSP50($N$=49,$D$=1)}                                                                                                                                                                                   \\ \hline
\multicolumn{1}{c|}{$M$=}                              & \multicolumn{3}{c}{3}                                  & \multicolumn{3}{c}{5}                                  & \multicolumn{3}{c}{7}                                  \\ \hline
\multicolumn{1}{c|}{Methods}                         & Obj.            & Gap       & \multicolumn{1}{c}{Time} & Obj.            & Gap       & \multicolumn{1}{c}{Time} & Obj.            & Gap       & \multicolumn{1}{c}{Time} \\ \hline
\multicolumn{1}{l|}{HGA}                             & \textbf{2.4184}          & -         & 7m                       & \textbf{2.0154}          & -         & 6m                       & \textbf{1.9386}          & -         & 4m                       \\
\multicolumn{1}{l|}{LKH3}                            & 2.4301          & 0.4816\%  & 2m                       & 2.0182          & 0.1351\%  & 4m                       & 1.9408          & 0.1157\%  & 7m                       \\
\multicolumn{1}{l|}{OR-Tools(600s)}                  & 2.5878          & 7.0030\%  & 98s                      & 2.1584          & 7.0938\%  & 3m                       & 2.0992          & 8.2849\%  & 3m                       \\ \hline
\multicolumn{1}{l|}{DAN}                             & 2.9899          & 23.629\% & 22s                      & 2.3225          & 15.238\% & 23s                      & 2.1492          & 10.864\% & 30s                      \\
\multicolumn{1}{l|}{Equity-Transformer}              & 2.5643          & 6.0301\%  & \textless{}1s            & 2.0798          & 3.1935\%  & \textless{}1s            & 1.9618          & 1.1964\%  & \textless{}1s            \\
\multicolumn{1}{l|}{Equity-Transformer-$\times$8aug} & 2.4901          & 2.9641\%  & \textless{}1s            & 2.0399          & 1.2139\%  & \textless{}1s            & 1.9465          & 0.4076\%  & \textless{}1s            \\
\multicolumn{1}{l|}{DPN}                            & 2.5149          & 3.9888\%  & \textless{}1s            & 2.0545          & 1.9405\%  & \textless{}1s            & 1.9549          & 0.8413\%  & \textless{}1s            \\
\multicolumn{1}{l|}{DPN-$\times$8aug}               & 2.4654          & 1.9433\%  & \textless{}1s            & 2.0337          & 0.9087\%  & \textless{}1s            & 1.9460          & 0.3804\%  & \textless{}1s            \\
\multicolumn{1}{l|}{DPN-$\times$8aug-$\times$16per} & \cellcolor[HTML]{D9D9D9}2.4637 & \cellcolor[HTML]{D9D9D9}1.8728\%  & 1s                       & \cellcolor[HTML]{D9D9D9}2.0324 & \cellcolor[HTML]{D9D9D9}0.8416\%  & 1s                       & \cellcolor[HTML]{D9D9D9}1.9454 & \cellcolor[HTML]{D9D9D9}0.3503\%  & 1s                       \\ \midrule
\multicolumn{10}{c}{Min-max mTSP100($N$=99,$D$=1)}                                                                                                                                                                                   \\ \hline
\multicolumn{1}{c|}{$M$=}                              & \multicolumn{3}{c}{5}                                  & \multicolumn{3}{c}{7}                                  & \multicolumn{3}{c}{10}                                 \\ \hline
\multicolumn{1}{c|}{Methods}                         & Obj.            & Gap       & \multicolumn{1}{c}{Time} & Obj.            & Gap       & \multicolumn{1}{c}{Time} & Obj.            & Gap       & \multicolumn{1}{c}{Time} \\ \hline
\multicolumn{1}{l|}{HGA}                             & \textbf{2.1893}          & -         & 20m                      & 1.9963          & 0.1240\%  & 16m                      & 1.9507          & 0.0273\%  & 14m                      \\
\multicolumn{1}{l|}{LKH3}                            & 2.1924          & 0.1410\%  & 16m                      & \textbf{1.9939}          & -         & 17m                      & \textbf{1.9502}          & -         & 17m                      \\
\multicolumn{1}{l|}{OR-Tools(600s)}                  & 2.3477          & 7.2346\%  & 5m                       & 2.1627          & 8.4671\%  & 6m                       & 2.1465          & 10.068\% & 7m                       \\ \hline
\multicolumn{1}{l|}{DAN}                             & 2.6995          & 23.305\% & 40s                      & 2.3115          & 15.930\% & 42s                      & 2.1556          & 10.534\% & 46s                      \\
\multicolumn{1}{l|}{Equity-Transformer}              & 2.3042          & 5.2456\%  & \textless{}1s            & 2.0487          & 2.7480\%  & \textless{}1s            & 1.9583          & 0.4153\%  & \textless{}1s            \\
\multicolumn{1}{l|}{Equity-Transformer-$\times$8aug} & 2.2563          & 3.0577\%  & \textless{}1s            & 2.0225          & 1.4345\%  & \textless{}1s            & 1.9534          & 0.1652\%  & \textless{}1s            \\
\multicolumn{1}{l|}{DPN}                            & 2.2704          & 3.7017\%  & \textless{}1s            & 2.0335          & 1.9860\%  & \textless{}1s            & 1.9587          & 0.4377\%  & \textless{}1s            \\
\multicolumn{1}{l|}{DPN-$\times$8aug}               & 2.2346          & 2.0703\%  & \textless{}1s            & 2.0143          & 1.0223\%  & \textless{}1s            & 1.9534          & 0.1671\%  & \textless{}1s            \\
\multicolumn{1}{l|}{DPN-$\times$8aug-$\times$16per} & \cellcolor[HTML]{D9D9D9}2.2314 & \cellcolor[HTML]{D9D9D9}1.9240\%  & 1s                       & \cellcolor[HTML]{D9D9D9}2.0126 & \cellcolor[HTML]{D9D9D9}0.9388\%  & 1s                       & \cellcolor[HTML]{D9D9D9}1.9532 & \cellcolor[HTML]{D9D9D9}0.1542\%  & 1s                       \\ \midrule[0.5mm]
\multicolumn{10}{c}{Min-max mPDP50($N$=50,$D$=1)}                                                                                                                                                                                   \\ \hline
\multicolumn{1}{c|}{$M$=}                              & \multicolumn{3}{c}{3}                                  & \multicolumn{3}{c}{5}                                  & \multicolumn{3}{c}{7}                                  \\ \hline
\multicolumn{1}{c|}{Methods}                         & Obj.            & Gap       & \multicolumn{1}{c}{Time} & Obj.            & Gap       & \multicolumn{1}{c}{Time} & Obj.            & Gap       & \multicolumn{1}{c}{Time} \\ \hline
\multicolumn{1}{l|}{OR-Tools(600s)}                  & 3.6796          & 6.1250\%    & 20m                      & 2.9924          & 7.8586\%    & 23m                      & 2.7867          & 10.663\%   & 29m                      \\
\multicolumn{1}{l|}{Equity-Transformer}              & 5.1494          & 48.515\%   & \textless{}1s            & 3.7159          & 33.937\%   & \textless{}1s            & 3.1586          & 25.431\%   & \textless{}1s            \\
\multicolumn{1}{l|}{Equity-Transformer-$\times$8aug} & 4.5857          & 32.258\%   & 1s                       & 3.3566          & 20.986\%   & 1s                       & 2.8764          & 14.227\%   & 1s                       \\
\multicolumn{1}{l|}{DPN}                            & 3.6247          & 4.5413\%    & \textless{}1s            & 2.8946          & 4.3347\%    & \textless{}1s            & 2.5889          & 2.8081\%    & \textless{}1s            \\
\multicolumn{1}{l|}{DPN-$\times$8aug}               & 3.4744          & 0.2082\%    & 1s                       & 2.7803          & 0.2141\%    & 1s                       & 2.5223          & 0.1651\%    & 1s                       \\
\multicolumn{1}{l|}{DPN-$\times$8aug-$\times$16per} & \cellcolor[HTML]{D9D9D9}\textbf{3.4672} & \cellcolor[HTML]{D9D9D9}-         & 1s                       & \cellcolor[HTML]{D9D9D9}\textbf{2.7744} & \cellcolor[HTML]{D9D9D9}-         & 1s                       & \cellcolor[HTML]{D9D9D9}\textbf{2.5182} & \cellcolor[HTML]{D9D9D9}-         & 1s                       \\ \midrule
\multicolumn{10}{c}{Min-max mPDP100($N$=100,$D$=1)}                                                                                                                                                                                 \\ \hline
\multicolumn{1}{c|}{$M$=}                              & \multicolumn{3}{c}{5}                                  & \multicolumn{3}{c}{7}                                  & \multicolumn{3}{c}{10}                                 \\ \hline
\multicolumn{1}{c|}{Methods}                         & Obj.            & Gap       & \multicolumn{1}{c}{Time} & Obj.            & Gap       & \multicolumn{1}{c}{Time} & Obj.            & Gap       & \multicolumn{1}{c}{Time} \\ \hline
\multicolumn{1}{l|}{OR-Tools(600s)}                  & 14.315         & 309.48\%  & 4h                       & 14.486         & 378.96\%  & 5h                       & 14.500         & 438.07\%  & 5h                       \\
\multicolumn{1}{l|}{Equity-Transformer}              & 5.5571          & 58.958\%   & \textless{}1s            & 4.4831          & 48.234\%   & \textless{}1s            & 3.7483          & 39.092\%   & \textless{}1s            \\
\multicolumn{1}{l|}{Equity-Transformer-$\times$8aug} & 5.0735          & 45.123\%   & 1s                       & 4.1152          & 36.069\%   & 1s                       & 3.4411          & 27.690\%   & 1s                       \\
\multicolumn{1}{l|}{DPN}                            & 3.6500          & 4.4054\%    & \textless{}1s            & 3.1404          & 3.8364\%    & \textless{}1s            & 2.7998          & 3.8933\%    & \textless{}1s            \\
\multicolumn{1}{l|}{DPN-$\times$8aug}               & 3.5123          & 0.4673\%    & 1s                       & 3.0430          & 0.6165\%    & 1s                       & 2.7101          & 0.5647\%    & 1s                       \\
\multicolumn{1}{l|}{DPN-$\times$8aug-$\times$16per} & \cellcolor[HTML]{D9D9D9}\textbf{3.4960} & \cellcolor[HTML]{D9D9D9}-         & 2s                       & \cellcolor[HTML]{D9D9D9}\textbf{3.0244} & \cellcolor[HTML]{D9D9D9}-         & 2s                       & \cellcolor[HTML]{D9D9D9}\textbf{2.6949} & \cellcolor[HTML]{D9D9D9}-         & 2s                       \\ 
\bottomrule[0.5mm]
\end{tabular}
}}\label{50100}
\end{table*}

\begin{table*}[htbp]
\caption{The average objective function values (i.e., Obj.) on 100 randomly generated larger-scale min-max mTSP or min-max mPDP instances. The overall best results and the best learning-based results are highlighted in bold and gray backgrounds, respectively.}
\centering
\resizebox{\textwidth}{!}{
\renewcommand\arraystretch{1}
\setlength{\tabcolsep}{3.85mm}
\small{\begin{tabular}{l|ccc|ccc|ccc}
\toprule[0.5mm]
                                  & \multicolumn{3}{c|}{Min-max mTSP200}                & \multicolumn{3}{c|}{Min-max mTSP500}                & \multicolumn{3}{c}{Min-max mTSP1,000}                \\ \hline
\multicolumn{1}{c|}{$M$=}           & 10              & 15              & 20              & 30              & 40              & 50              & 50              & 75              & 100             \\ \hline
\multicolumn{1}{c|}{Methods}      & Obj.            & Obj.            & Obj.            & Obj.            & Obj.            & Obj.            & Obj.            & Obj.            & Obj.            \\ \hline
HGA                               & 1.9861          & \textbf{1.9628} & \textbf{1.9627} & \textbf{2.0061} & 2.0061          & 2.0061          & \textbf{2.0448} & 2.0448 &            2.0448 \\
LKH3                              & \textbf{1.9817} & 1.9628          & 1.9628          & 2.0061          & 2.0061          & 2.0061          & 2.0448          & 2.0448          & 2.0448          \\
OR-Tools(600s)                    & 2.3711          & 2.3687          & 2.3687          & 8.9338          & 8.9356          & 8.9308          & 16.436         & 16.436         & 16.436         \\ \hline
NCE*                              & 2.07            & 1.97            & 1.96            & 2.07            & 2.01            & 2.01            & 2.13            & 2.07            & 2.05            \\
DAN                               & 2.3586          & 2.1732          & 2.1151          & 2.2345          & 2.1610          & 2.1465          & 2.3390          & 2.2544          & 2.2394          \\
ScheduleNet*                      & 2.35            & 2.13            & 2.07            & 2.16            & 2.12            & 2.09            & 2.26            & 2.17            & 2.16            \\
Equity-Transformer-F-$\times$8aug & 2.0500          & 1.9688          & 1.9631          & 2.0165          & 2.0084          & 2.0068          & 2.0634          & 2.0531          & 2.0488          \\
DPN-F-$\times$8aug               & 2.0030          & 1.9647          & 1.9628          & 2.0065          & 2.0061          & 2.0061          & 2.0452          & 2.0448          & 2.0448          \\
DPN-F-$\times$8aug-$\times$16per & \cellcolor[HTML]{D9D9D9}1.9993          & \cellcolor[HTML]{D9D9D9}1.9640          & \cellcolor[HTML]{D9D9D9}1.9628          & \cellcolor[HTML]{D9D9D9}2.0061          & \cellcolor[HTML]{D9D9D9}\textbf{2.0061} & \cellcolor[HTML]{D9D9D9}\textbf{2.0061} & \cellcolor[HTML]{D9D9D9}2.0450 & \cellcolor[HTML]{D9D9D9}\textbf{2.0448} & \cellcolor[HTML]{D9D9D9}\textbf{2.0448} \\ \midrule[0.5mm]
\multicolumn{1}{c|}{}             & \multicolumn{3}{c|}{Min-max mPDP200}                & \multicolumn{3}{c|}{Min-max mPDP500}                & \multicolumn{3}{c}{Min-max mPDP1,000}                \\ \hline
\multicolumn{1}{c|}{$M$=}           & 10              & 15              & 20              & 30              & 40              & 50              & 50              & 75              & 100             \\ \hline
\multicolumn{1}{c|}{Methods}      & Obj.            & Obj.            & Obj.            & Obj.            & Obj.            & Obj.            & Obj.            & Obj.            & Obj.            \\ \hline
OR-Tools(600s)                    & 45.299         & 45.387         & 45.131         & 140.85        & 140.92        & 140.79        & 280.22        & 280.19        & 280.14        \\
Equity-Transformer-F-$\times$8aug & 4.9143          & 3.8186          & 3.3417          & 4.4619          & 3.7723          & 3.4455          & 4.9328          & 3.9198          & 3.5241          \\
Equity-Transformer-F-sample*      & 4.68            & 3.65            & 3.18            & 4.11            & 3.52            & 3.23            & 4.73            & 3.77            & 3.38            \\ \hline
DPN-F-$\times$8aug               & 3.3227          & 2.8630          & 2.6735          & 3.1615          & 3.0264          & 2.9379          & 3.2802          & 3.0673          & 3.0000          \\
DPN-F-$\times$8aug-$\times$16per & \cellcolor[HTML]{D9D9D9}\textbf{3.2959} & \cellcolor[HTML]{D9D9D9}\textbf{2.8363} & \cellcolor[HTML]{D9D9D9}\textbf{2.6519} & \cellcolor[HTML]{D9D9D9}\textbf{3.0878} & \cellcolor[HTML]{D9D9D9}\textbf{2.9510} & \cellcolor[HTML]{D9D9D9}\textbf{2.8690} & \cellcolor[HTML]{D9D9D9}\textbf{3.2263} & \cellcolor[HTML]{D9D9D9}\textbf{2.9811} & \cellcolor[HTML]{D9D9D9}\textbf{2.9114} \\ 
\bottomrule[0.5mm]
\end{tabular}
}}\label{more}
\end{table*}

\section{Experiments: Single-depot Min-max VRPs}
To evaluate the effectiveness of the proposed DPN method on single-depot min-max VRPs, we implement it on min-max mTSP and min-max mPDP. Detailed formulations of the two problems are provided in Appendix \ref{defination}.

\textbf{Training Settings.} For both the min-max mTSP and min-max mPDP, we conduct training from scratch on 50-scale (i.e., min-max mTSP50 and min-max mPDP50) and 100-scale problems respectively, and then conduct fine-tuning for larger problem scales (200-, 500-, and 1,000-scale). For all the scales involved, the DPN utilizes a 6-layer P\&N Encoder, a uniform sampling of the number of agents $M$ from 2 to 10, and a fixed number of permutations $K$ to 60. The MHA function comprises 8 heads $H=8$, with the embedding dimension $d=128$ and the hidden dimension $d_h=512$. The Adam optimizer \citep{Adam} is employed with weight decay $\beta=1$, and the learning rate $\alpha=1e^{-4}$ in training models from scratch and being $\alpha=1e^{-5}$ for the fine-tuning. Moreover, the hyperparameters for various problems and scales are provided in Appendix \ref{Seetings}. With utilizing an NVIDIA Tesla V100S GPU, the training duration for the DPN targeting min-max mTSP100 and min-max mPDP100 span three and five days, respectively, and the fine-tuning for larger scale is completed within a few hours.

\subsection{Performance Evaluation}
\textbf{Baselines.} We select representative heuristic and learning-based methods as baselines. For heuristic methods, the Lin-Kernighan-Helsgaun3 algorithm (LKH3) \citep{lin1973effective} and the hybrid genetic algorithm (HGA) \citep{mahmoudinazlou2024hybrid} are adopted, and these algorithms conduct 10 runs for each instance. As an outstanding operations research (OR) solver, OR-Tools has demonstrated excellent generalization ability. We report its search results within 600 seconds. For learning-based methods, we adopt a parallel planning method DAN \citep{cao2021dan} and a sequential planning method Equity-Transformer \citep{son2023solving}. For efficient yet unavailable neural methods such as ScheduleNet \citep{park2021schedulenet} and NCE \citep{kim2022learning}, we list the reported results as well. The min-max mPDP has also been investigated, but heuristic methods can hardly be implemented on Euclidean coordinates \citep{lau2022multi}. Therefore, our comparative algorithm only considers OR-Tools and the learning-based method Equity-Transformer. 

\textbf{Performance on Medium Scales.} We first examine the performance of DPN on medium-scale problems, including min-max mTSP and min-max mPDP of 50- and 100-scale. For each scale, we conduct tests on three datasets distinct with various $M$ values. Each dataset contains 100 randomly generated instances, and the distribution of $M$ follows the settings in \citet{kim2022learning}. All the involved learning-based models are trained from scratch at the testing scale. The $\times$8aug data augmentation method is available as a variant, which conducts equivalent transformations to augment the original instance and reports the best results among eight equivalent instances \citep{kwon2020pomo}. For DPN, we additionally provide a $\times$16per data augmentation method which reports the best solution among 16 agent permutations (i.e., $K=16$ in testing). Table \ref{50100} exhibits the comparison results on 12 datasets, and it presents the average objective function values, the gap from the best algorithm, and the testing time for each dataset. The best result of each dataset is highlighted in bold, while the best learning-based result is highlighted in a gray background. Results demonstrate that the proposed DPN-$\times$8aug-$\times$16per version achieves the best performance among the learning methods in all the 12 datasets and significantly stands out in min-max mPDPs. Compared to heuristic methods such as HGA and LKH3, the DPN variants can achieve competitive results in a much shorter time. The proposed DPN method consistently outperforms the Equity-Transformer under the same settings, with an average reduction of 0.72\% in the optimality gap of min-max mTSP and 33.7\% for min-max mPDP.

\begin{figure}[t]
    \centering
    \subfigure[min-max mTSP]{\includegraphics[width = 0.235\textwidth]{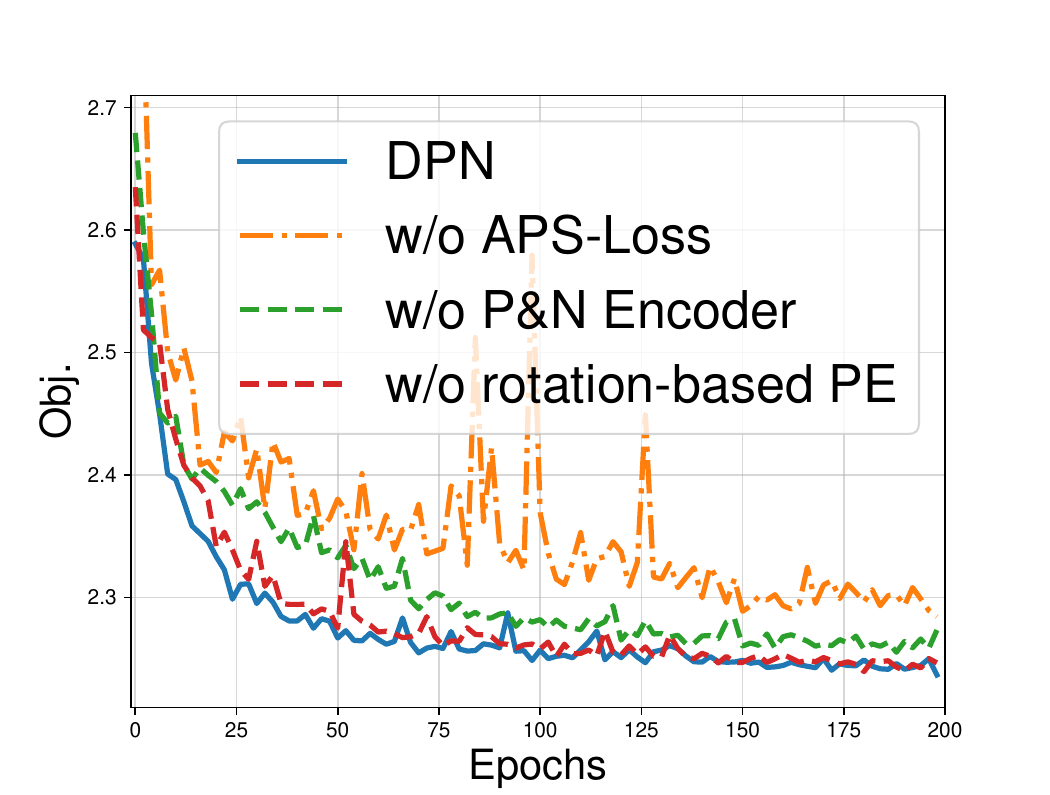}}
    \subfigure[min-max mPDP]{\includegraphics[width = 0.235\textwidth]{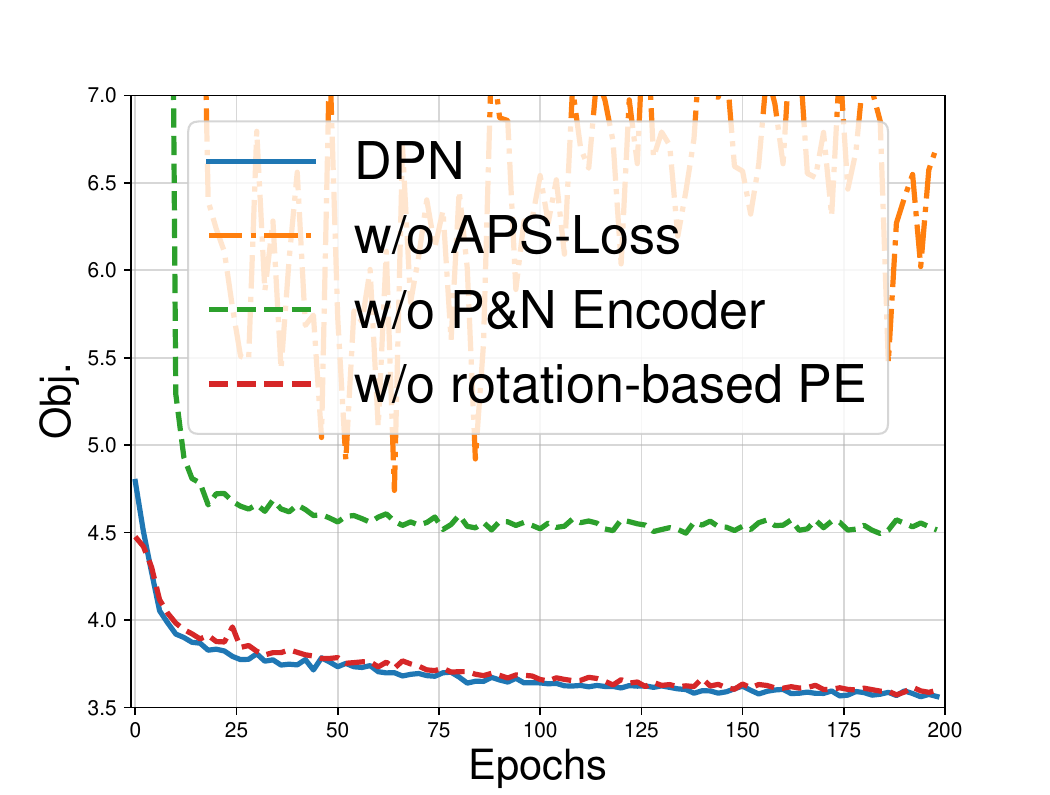}}\caption{Training curves for ablation study ($M$=5).}\label{fig:curve}
\end{figure}

\begin{table}[t]
\caption{Ablation study on proposed components in DPN.}
\centering
\renewcommand\arraystretch{0.96}
\setlength{\tabcolsep}{3.85mm}
\small{\begin{tabular}{l|cc}
\hline\toprule
                         \multicolumn{1}{c|}{Min-max}                              & \multicolumn{1}{l}{mTSP100} & \multicolumn{1}{l}{mPDP100} \\ \hline 
\multicolumn{1}{c|}{$M$=}                                  & 5                                   & 5                                   \\ \hline
w/o P\&N Encoder                                         & 2.2406                              & 4.5119                              \\
w/o APS-Loss                                             & 2.2621                              & 5.1915                              \\
w/o Rotation-based PE                                    & 2.2347                              & 3.5158                              \\ \hline
DPN-$\times$8aug-$\times$16per & \textbf{2.2314}                     & \textbf{3.4960}                     \\ \bottomrule\hline
\end{tabular}}\label{component}
\end{table}

\begin{table*}[htbp]
\caption{The average objective function values (i.e., Obj.) and testing times (i.e., Time) on 12 multi-depot min-max VRP datasets.}
\centering
\renewcommand\arraystretch{1}
\setlength{\tabcolsep}{1.8mm}
\small{\begin{tabular}{l|cccccc|cccccc}
\hline\toprule
                                  & \multicolumn{6}{c|}{MDVRP50($N$=50),$D$=6}                                                              & \multicolumn{6}{c}{MDVRP100($N$=100),$D$=8}                                  \\ \cline{2-13} 
\multicolumn{1}{c|}{$M$=}           & \multicolumn{2}{c}{3}           & \multicolumn{2}{c}{5}           & \multicolumn{2}{c|}{7}          & \multicolumn{2}{c}{5}  & \multicolumn{2}{c}{7}  & \multicolumn{2}{c}{10} \\ \hline
\multicolumn{1}{c|}{Methods}      & Obj.            & Time          & Obj.            & Time          & Obj.            & Time          & Obj.            & Time & Obj.            & Time & Obj.            & Time \\ \hline
CE*                               & 2.25            & 40m           & 1.53            & 17m           & 1.28            & 11m           & 1.85            & 50m  & 1.43            & 1h   & 1.18            & 1h   \\
OR-Tools*                         & 2.64            & 4m            & 1.68            & 5m            & 1.36            & 5m            & 2.17            & 6h   & 1.60            & 3h   & 1.29            & 2h   \\ \hline
NCE*                              & 2.25            & 3m            & 1.53            & 4m            & 1.28            & 5m            & 1.86            & 19m  & 1.43            & 20m  & 1.18            & 26m  \\
DPN-$\times$8aug-$\times$16per   & 2.1491          & \textless{}1s & 1.4431          & \textless{}1s & 1.2012          & \textless{}1s & 1.8056          & 1s   & 1.4099          & 1s   & 1.1527          & 1s   \\
DPN-F-$\times$8aug-$\times$16per & \cellcolor[HTML]{D9D9D9}\textbf{2.1404} & 1s            & \cellcolor[HTML]{D9D9D9}\textbf{1.4394} & 1s            & \cellcolor[HTML]{D9D9D9}\textbf{1.1969} & 1s            & \cellcolor[HTML]{D9D9D9}\textbf{1.7936} & 2s   & \cellcolor[HTML]{D9D9D9}\textbf{1.4001} & 2s   & \cellcolor[HTML]{D9D9D9}\textbf{1.1429} & 2s   \\ \midrule[0.5mm]
                                  & \multicolumn{6}{c|}{FMDVRP50($N$=50),$D$=6}                                                             & \multicolumn{6}{c}{FMDVRP100($N$=100),$D$=8}                                 \\ \cline{2-13} 
\multicolumn{1}{c|}{$M$=}           & \multicolumn{2}{c}{3}           & \multicolumn{2}{c}{5}           & \multicolumn{2}{c|}{7}          & \multicolumn{2}{c}{5}  & \multicolumn{2}{c}{7}  & \multicolumn{2}{c}{10} \\ \hline
\multicolumn{1}{c|}{Methods}      & Obj.            & Time          & Obj.            & Time          & Obj.            & Time          & Obj.            & Time & Obj.            & Time & Obj.            & Time \\ \hline
CE*                               & 2.07            & 35m           & 1.41            & 15m           & 1.19            & 9m            & \textbf{1.74}   & 6h   & \textbf{1.34}   & 4h   & \textbf{1.09}   & 1h   \\
OR-Tools*                         & 2.39            & 4m            & 1.56            & 4m            & 1.27            & 4m            & 2.00            & 51m  & 1.51            & 54m  & 1.20            & 57m  \\ \hline
NCE*                              & 2.08            & 2m            & 1.40            & 3m            & 1.19            & 4m            &  \cellcolor[HTML]{D9D9D9}1.75            & 11m  &  \cellcolor[HTML]{D9D9D9}\textbf{1.34}   & 16m  &  \cellcolor[HTML]{D9D9D9}\textbf{1.09}   & 22m  \\
ScheduleNet*                      & 2.61            & 9m            & 1.86            & 9m            & 1.57            & 10m           & 2.32            & 1h   & 1.86            & 1h   & 1.54            & 1h   \\
DPN-$\times$8aug-$\times$16per   & 2.0471          & \textless{}1s & 1.3869          & \textless{}1s & \cellcolor[HTML]{D9D9D9}\textbf{1.1619}          & \textless{}1s & 1.7694          & 1s   & 1.3708          & 1s   & 1.1028          & 1s   \\
DPN-F-$\times$8aug-$\times$16per & \cellcolor[HTML]{D9D9D9}\textbf{2.0429} & \textbf{1s}   & \cellcolor[HTML]{D9D9D9}\textbf{1.3856} & 1s            & 1.1649 & 1s            &1.7638          & 2s   & 1.3642          & 2s   & 1.1012          & 2s   \\ \bottomrule\hline
\end{tabular}}\label{mdvrp}
\end{table*}

\textbf{Performance on Larger Scales.} Neural solvers of min-max VRPs also demonstrate their ability to solve larger-scale problems \citep{son2023solving}. To evaluate the performance of DPN on large-scale problems, we generate datasets with 100 instances with scales of 200, 500, and 1,000, and the number of agents $M$ follows the same setting in \citet{son2023solving}. Both Equity-Transformer and DPN conduct fine-tunings at scales of 200 and 500. These fine-tuned models are denoted as F in Table \ref{more}, and the 1,000-scale results are obtained using the 500-scale fine-tuned model. Results marked with ``*'' indicate results reported from \citet{son2023solving} on the same test set. As exhibited in Table \ref{more}, the proposed DPN-F-$\times$8aug-$\times$16per acquires the best objective functions on four larger-scale min-max mTSP datasets and all nine larger-scale min-max mPDP datasets. In a total of 18 datasets, variants of DPN consistently significantly beat the Equity-Transformer and other learning-based solvers, demonstrating DPN's adaptability to large-scale scenarios. 

Experiments on both medium and large scales exhibit the effectiveness of DPN in single-depot min-max VRPs while also demonstrating the significant value of separately considering partition and navigation in sequential planning. Experiments in Appendix \ref{Experiment2} further demonstrate the good generalization performance of DPN.

\subsection{Analyses}\label{ablation}
To validate the necessity of the components in DPN, we conduct a series of ablation studies. As detailed in Table \ref{component}, we remove the proposed components (i.e., P\&N Encoder, APS-Loss, and Rotation-based PE) and obtain the three ablation models. The 'w/o' (without) notation in Table \ref{component} and Figure \ref{fig:curve} indicates degrading the component by using the corresponding components proposed in Equity-Transformer. Three ablation models are evaluated with both the $\times$8aug and $\times$16per on min-max mTSP100 and min-max mPDP100 with $M$=5 agents. Results show that all three components of DPN make substantial contributions, with the P\&N Encoder and APS-Loss playing more significant roles. The full-version DPN demonstrates notable advantages in convergence speed over the ablation models. Ablation studies on $M$=10 are presented in Appendix \ref{ablationm10} where Appendix \ref{ablationpe} further suggests that the Rotation-based PE is less vulnerable to the depot locations.

\section{Experiments: Multi-depot Min-max VRPs}

This section further evaluates DPN on two representative min-max multi-depot VRPs including min-max MDVRP and min-max FMDVRP. Previous sequential planning neural solvers for min-max VRP are all limited to single-depot problems, and DPN is the first to solve multi-depot ones.

\subsection{P\&N Encoder for Multi-depot}\label{PN2}
In solving single-depot min-max VRP, P\&N Encoder only processes the agent embeddings and customer embeddings and directly attaches the depot coordinates to agent embeddings. However, the $D$ depots in multi-depot problems necessitate additional structures to process the embeddings of depots. To represent the depot assignment features in each route, DPN for multi-depot min-max VRPs additionally employs two additional structures in the partition part of each layer to represent depot-related partitions. The specific structure is described in detail in Appendix \ref{mdvrpmodel}.

\subsection{Performance Evaluation}
When training DPN for multi-depot min-max VRPs with both 50- and 100-scales, the number of depots $D$ is uniformly sampled from 2 to 10. In testing, the number of depots is set to 6 for 50-scale datasets and 8 for 100-scale ones. Training on the corresponding number of depots of test sets can obtain better results, so we further provide a fine-tuned version consistently denoted as F in Table \ref{mdvrp}. We conduct experiments on 12 100-instance datasets of min-max MDVRP and min-max FMDVRP. A heuristic algorithm CROSS exchange (CE) \citep{taillard1997tabu}, the OR-Tools, a neural solver ScheduleNet \citep{park2021schedulenet}, and the most efficient among all existing learning-based neural solver of multi-depot min-max VRPs Neuro CE (NCE) \citep{kim2022learning} are employed as baselines. Since some of these methods are unavailable, we provide the reported results in \citet{kim2022learning}. Based on the results in Table \ref{mdvrp}, we can conclude that DPN is well applied to multi-depot min-max VRP. Especially, DPN-F-$\times$8aug-$\times$16per achieves significant advantages on min-max MDVRP and 50-scale min-max FMDVRP. CE and NCE outperform the proposed DPN on 100-scale min-max FMDVRP.

\section{Conclusion}
In this paper, we have introduced a novel Decoupling-Partition-Navigation (DPN) method for solving min-max VRPs. It is the first attempt to decouple the partition and navigation in sequential-planning-based min-max VRP solvers. DPN consists of a novel P\&N Encoder, a novel APS-Loss, and a Rotation-based PE. Experimental results on four min-max VRPs have demonstrated that DPN significantly outperforms existing learning-based methods. In addition, comprehensive ablation studies have been performed to verify the effectiveness of each component of DPN. 

\textbf{Limitation and Future Work.} Although DPN efficiently processes constraints (e.g., route number $M$) and representations in min-max VRP, it cannot be directly applied to general VRPs (further discussed in \ref{discussap}). In the future, we plan to extend the idea of decoupled representation in DPN to general VRPs, pursuing better representations of each route.
%Although DPN achieves good overall performance, learning the optimal sub-graph partition remains challenging. In the future, we plan to introduce other problem-specific properties to approach the global partition optimum.
\section*{Acknowledge}
This work was supported by the National Key Research and Development Program of China (Grant No. 2022YFA1004102), the National Natural Science Foundation of China (Grant No. 62106096), the Natural Science Foundation of Guangdong Province (Grant No. 2024A1515011759), the National Natural Science Foundation of Shenzhen (Grant No. JCYJ20220530113013031).

\section*{Impact Statement}

This paper presents work whose goal is to advance the field of Machine Learning. There are many potential societal consequences of our work, none which we feel must be specifically highlighted here.

\bibliography{example_paper}
\bibliographystyle{icml2024}

%%%%%%%%%%%%%%%%%%%%%%%%%%%%%%%%%%%%%%%%%%%%%%%%%%%%%%%%%%%%%%%%%%%%%%%%%%%%%%%
%%%%%%%%%%%%%%%%%%%%%%%%%%%%%%%%%%%%%%%%%%%%%%%%%%%%%%%%%%%%%%%%%%%%%%%%%%%%%%%
% APPENDIX
%%%%%%%%%%%%%%%%%%%%%%%%%%%%%%%%%%%%%%%%%%%%%%%%%%%%%%%%%%%%%%%%%%%%%%%%%%%%%%%
%%%%%%%%%%%%%%%%%%%%%%%%%%%%%%%%%%%%%%%%%%%%%%%%%%%%%%%%%%%%%%%%%%%%%%%%%%%%%%%
\newpage
\appendix
\onecolumn
\section{Related Work}\label{related}

\subsection{Learning-based Methods for TSP and CVRP}

Currently, plenty of learning-based methods have been applied to TSP and CVRP. Classified by the solution generation process, these methods can be divided into two main categories \citep{bengio2021machine}. Learning improvement heuristic methods execute neural-parameter-assisted operators to modify solutions iteratively \citep{wu2021learning,zheng2023pareto}, while the learning constructive heuristic methods directly convert the input coordinates to near-optimal solutions in an end-to-end manner \citep{drakulic2023bqnco}. Although advanced learning improvement heuristic methods show advantages in small-scale instances with adequate running time \citep{ma2021learning,zheng2023pareto}, learning constructive heuristic methods are considered to have better generalization ability and more application value \citep{liu2022good,jiang2022learning,wang2023multiobjective}. Generally, most RL-based constructive methods including the famous Attention Model \citep{kool2019attention} employ a unified network structure including several multi-head attention encoder layers and a single-layer decoder. Advanced RL-based construction methods typically propose modifications to the original network structure \citep{jin2023pointerformer} and decoding methods \citep{sun2024learning}. LEHD \citep{luo2023neural} and ELG \citep{gao2023towards} propose modified network structures, while POMO \citep{kwon2020pomo} and Sym-NCO \citep{kim2022sym} utilize symmetry for better reinforcement learning baselines. 

Besides learning constructive heuristic methods, inspired by a similar idea of partition and navigation compared to DPN, two-stage methods \citep{hou2023generalize} are also compelling in solving TSP and CVRP. To divide and conquer large-scale problems, some works such as TAM \citep{hou2023generalize}, RBG \citep{zong2022rbg}, H-TSP \citep{pan2023h-tsp}, and GLOP \citep{ye2024glop} use two sets of independent neural networks (or heuristics) to generate a large-scale solution, which are respectively responsible for problem decomposition and sub-problem solving. These two-stage solving methods have achieved high-quality solutions for large-scale problems such as TSP with 10,000 nodes. Nevertheless, we want to clarify that thought inspired by a similar idea, as a one-stage constructive decoding method, the proposed DPN is different from these two-stage solving methods.

\subsection{Learning-based Methods for Min-max VRPs}\label{related_mmvrp}
The learning-based methods for solving min-max VRP can be classified into four categories: two-stage methods, learning improvement heuristics, parallel planning methods, and sequential planning methods \citep{son2023solving}. The application of learning-based neural solvers begins with \citet{kaempfer2018learning} and \citet{hu2020reinforcement}, which use imitation learning and RL respectively to learn a two-stage solution generation process for min-max mTSP. SplitNet \citep{liang2023splitnet} also adopted the two-stage planning process. Neural solvers in these two-stage methods are only used in one stage, dividing the customers into sub-graphs, while the TSP solver is introduced in another stage for customer navigation. As a learning improvement heuristic method, NCE \citep{kim2022learning} learns a neural-parameter-assisted CROSS exchange (CE) \citep{taillard1997tabu} operator and demonstrates zero-shot generalization ability among different min-max VRPs. The following two manners implement learning constructive heuristics for min-max VRPs. As discussed in Section \ref{sequantialplanning}, parallel planning methods DAN \citep{cao2021dan} and SchedulingNet \citep{park2021schedulenet} employ multiple networks to cooperatively process a limited number of routes ($M$ routes), while sequential planning methods like Equity-Transformer \citep{son2023solving} process the features of both customer and agents by a unified encoder and distinguish these two parts of representations through PE \citep{vaswani2017attention}. In the decoding phase, sequential planning methods sequentially generate the $M$ routes in a solution one after another, while parallel planning methods extend the $M$ routes simultaneously. Compared to parallel planning methods, sequential planning methods substantially reduce the decision space complexity \citep{son2023solving}.

In experiments, sequential planning methods demonstrate the inherent superiority among all the learning-based methods. From our perspective, except for the sequential planning method, all the other three kinds of methods, can be explained as employing different entities or procedures to represent features for learning partition and navigation, respectively. However, the Softmax function \citep{vaswani2017attention} in the attention-based encoder for sequential planning methods \citep{son2023solving} confuses the features for learning partition and navigation in its calculation. It appeals to an efficient sequential planning method that focuses on problem-specific properties especially decoupling the representations of partition and navigation in min-max VRPs.

\newpage
\section{Additional Defination}\label{defination}

\subsection{Sequential Planning in Min-max VRPs}\label{sequential}

Regarding solution generations, we generally follow the sequential planning process proposed in \citet{son2023solving}. From the multi-agent reinforcement learning (MARL) perspective, the min-max objective function is a cooperative task for multiple routes \citep{gao2023amarl}. Parallel planning methods use multiple networks to control the decoding of multiple routes separately. This framework is intuitive but has a more complex optimization space than sequential planning methods. Besides min-max VRPs, the sequential planning strategy is also proficient in general cooperative MARL tasks \citep{wen2022multi,ye2022towards}.

\begin{figure}[H]
    \centering
    \includegraphics[width = 0.85\hsize]{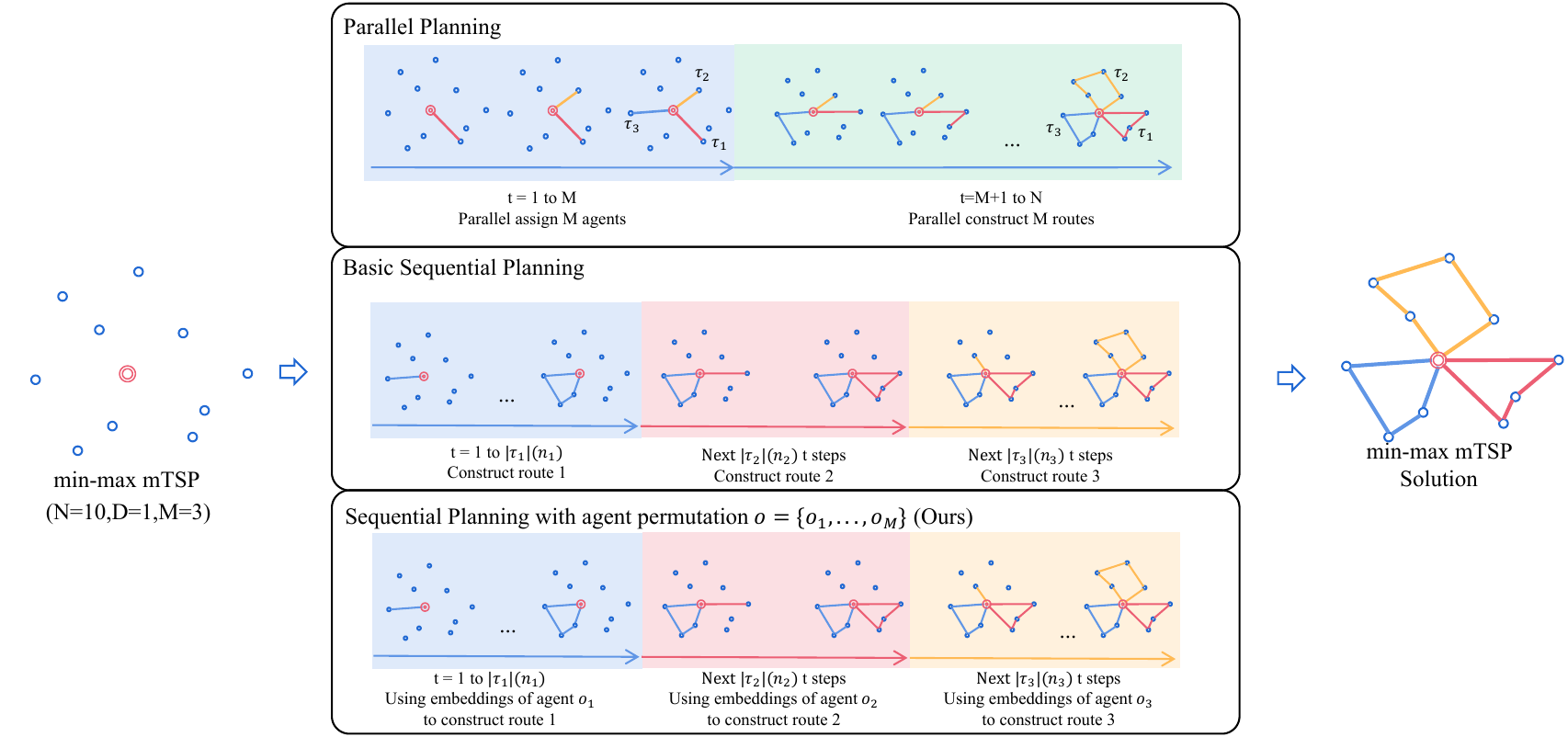}
    \caption{An min-max mTSP example of constructive-based planning methods for neural solvers of min-max VRP.}
    \label{fig:mdp}
\end{figure}

The Section \ref{sequantialplanning} briefly displays the policy of sequential planning, the MDP $\mathcal{M}=\{\mathcal{S},\mathcal{A},\boldsymbol{r},\mathcal{P}\}$ of sequential planning is defined in detail as follows:
\begin{itemize}
    \item \textbf{State.} The state $s_t\in \mathcal{S}$ consists of three parts, including the current partial solution, the instance $\mathcal{G}$, and the number of agents $M$. The partial solution in the initial state $s_0$ is an empty set and the terminal state $s_T$ contains a feasible solution including a set of routes $\mathcal{T}=\{\boldsymbol{\tau}^1,\ldots,\boldsymbol{\tau}^M\}$.
    \item \textbf{Action.} An action $a_t\in \mathcal{A}$ represents a selected index from candidate nodes in $\mathcal{V}$ which is then used to extend the current solution in $s_t$. Since all customers can only be accessed once, so $a_t$ can only select from unvisited customers or the set of depots in $\mathcal{V}$. If $a_t$ is a depot node, the following state $s_{t+1}$ will end the current route and start a new one.
    \item \textbf{Reward.} Sequential planning methods only assign rewards to the final state, $r_T=-f(\mathcal{T})$.
    \item \textbf{Policy.} The policy of sequentially constructing the set of routes is provided in Eq. \eqref{sq1}. With the sub-graphs constrained in Eq. \eqref{partition}, the policy for generating nodes in routes one by one is formulated with Eq. \eqref{navigation-policy} as follows:
    \begin{align}
        p_\theta(\mathcal{T}|\mathcal{G},M)=\prod_{t=1}^{T-1}p_\theta(a_t|s_t)=\prod_{i=1}^{M}p(\boldsymbol{\tau}^i|\boldsymbol{\tau}^{1:i},\mathcal{G},\theta)=\prod_{i=1}^{M}\prod_{j=1}^{n^i}\pi_\theta(\boldsymbol{\tau}^{i}(j)|\boldsymbol{\tau}^{i}(1:j),\mathcal{G}^{i}),
    \end{align}
\end{itemize}
where $\mathcal{G}^{i}\in P_{\theta,M}(\mathcal{G})$. Figure \ref{fig:mdp} exhibits a sketch map of the parallel planning process, the basic sequential planning method in \citet{son2023solving}, and the sequential planning method with agent permutation implemented in the proposed DPN. The proposed agent permutation only makes changes at the transaction to $s_{t+1}$ when $a_t$ is a depot node. When the construction of each route concludes, the agent for the next route's decoding is selected based on a sampled agent permutation $\boldsymbol{o}$. The detailed policy introducing a permutation $\boldsymbol{o}$ is provided in Eq. \eqref{total-loss}. The sequential planning process in Equity-Transformer can be regarded as a special situation of the implemented sequential planning with agent permutation, with $\boldsymbol{o}=(1,2,\ldots,n)$, and the sampling of agent permutations in DPN is implemented by randomly shuffling the special permutation $\boldsymbol{o}=(1,2,\ldots,n)$ for 100 times.

\newpage
\subsection{Min-max Multi-agent Travel Salesman Problem}
\label{sub:mTSP_form}
mTSP is a multi-agent extension of TSP with a single depot and a restricted number of routes. Min-max mTSP is a variant of mTSP problem with optimizing a min-max objective function. Min-max mTSP comprises the set of nodes (customers and depot) $\mathcal{V}$, the number of routes $M$, and the set of Depot $\mathcal{D}$ ($|\mathcal{D}|=D=1$). We define $d_{ij}$ as the length between node $i$ and $j$, and the decision variable $x_{ijm}$ which denotes whether the edge between node $i$ and $j$ are taken by agent $m$. The other decision variable $u_{ik}$ is an integer representing the position (i.e., occur place) of node $i$ in route $m$. For each customer node, it is positive in only one route. The MILP formulation of mTSP is given as follows \citep{zha2020cnf}:
\begin{align}
\label{mTSP_form}
&\text{minimize}&& L&&\\
&\text{subject to.}& & \sum_{i \in \mathcal{V}}\sum_{j \in \mathcal{V}} d_{ij}x_{ijm} \leq L, && \forall m\in \{1,\ldots,M\}, i\neq j, & \\
&& & \sum_{j \in \mathcal{V}, i \neq j} x_{ijm} = 1, && \forall m \in  \{1,\ldots,M\}, \forall i \in \mathcal{D},& \\
&& &  \sum_{i \in \mathcal{V}, j\neq i} \sum_{m \in  \{1,\ldots,M\}} x_{ijm} = 1, &&\forall j \in \mathcal{V} \setminus \mathcal{D},&\\
&& & \sum_{i \in \mathcal{V}, i\neq j} x_{ijm}  -  \sum_{h \in \mathcal{V}, h\neq j} x_{jhm} = 0, &&\forall m \in  \{1,\ldots,M\},\forall j \in \mathcal{V} \setminus \mathcal{D},&\\
&& & u_{im} - u_{jm} + |\mathcal{V}| x_{ijm} \leq |\mathcal{V}|-1, && \forall m \in  \{1,\ldots,M\}, i,j \in \mathcal{V} \setminus \mathcal{D}, i \neq j,&\\
&& & 0 \leq u_{im} \leq |\mathcal{V}|-1 && \forall m \in  \{1,\ldots,M\}, i \in \mathcal{V} \setminus \mathcal{D} ,&\\
&& & x_{ijm} \in \{0,1\}, && \forall m \in  \{1,\ldots,M\}, i,j \in \mathcal{V},& \\
&& & u_{im} \in \mathbb{Z}, && \forall m \in  \{1,\ldots,M\}, i \in \mathcal{V}, & 
\end{align}
where $L$ denotes the distance of the longest route (i.e., makespan) among the set of $M$ routes. These formulations are generally adopted from \citet{kim2022learning}, $u_{im}=0$ for node $i$ not in route $m\in\{1,\ldots,M\}$.

The tasks of customer partition for different routes and customer navigation in each route of min-max mTSP are mentioned frequently in the main text. The optimal partition of min-max mTSPs assumes that the navigation task obtains a route with the optimal TSP length of any sub-graph $\mathcal{G}_i$ \citep{vandermeulen2019balanced}. Without affecting optimality, we ignore the impact of agent permutation and assume that $\mathcal{T}= \{\boldsymbol{\tau}^i\sim \text{TSP}(\cdot|\mathcal{G}^{i})\}$ for $\ i \in \{1\ldots, M\}$, the objective function of the partition task in min-max mTSP is as follows \citep{carlsson2009solving}:

\begin{equation}
    \begin{aligned}
        &\underset{\theta}{\text{minimize}} &&\lambda\\
        &\text{subject to.} &&L(\text{TSP}(\cdot|\mathcal{G}^{i}))\le \lambda, \forall i\in\{1,\ldots, M\},\\
        & &&\bigcap_{i\in\{1,\ldots, M\}}\mathcal{G}^i=\text{Depots},\quad\bigcup_{i\in\{1,\ldots, M\}}\mathcal{G}^i=\mathcal{G},\\
        & &&\{\mathcal{G}^1,\ldots,\mathcal{G}^M\}=P_{\theta,M}(\mathcal{G}),
    \end{aligned}\label{partition-mtsp}
\end{equation}

where $\text{TSP}(\cdot|\mathcal{G}^{i})$ represents the optimal TSP solution of the sub-graph $\mathcal{G}^{i}$. The navigation task optimizes the routing performance to approach the optimal TSP solution as follows:

\begin{equation}
\begin{aligned}
        &\underset{\theta}{\text{minimize}} &&L(\boldsymbol{\tau}^i)=\sum_{i=2}^{|\boldsymbol{\tau}^i|}\Vert \boldsymbol{x}_i-\boldsymbol{x}_{i-1}\Vert_2+\Vert \boldsymbol{x}_0-\boldsymbol{x}_{|\boldsymbol{\tau}^i|}\Vert_2\\
        &\text{subject to.} &&\boldsymbol{\tau}^i = \pi_\theta(\cdot|\mathcal{G}^i),
\end{aligned}\label{navigation-mtsp}
\end{equation}
where $i\in \{1,\ldots,M\}$, and $\boldsymbol{x}_i$ represents the 2-dimensional coordinate of the node with index i. $|\boldsymbol{\tau}^i|$ is adopted to indicate the number of nodes in $\boldsymbol{\tau}^i$

\newpage

\subsection{Min-max Multi-agent Pickup and Delivery Problem}
\label{sub:mPDP_form}
Min-max mPDP is another single-depot min-max VRP considered in DPN. The mPDP is a multi-agent extension of the famous pickup and delivery problem (PDP). The difference between mPDP and mTSP is that the $N$ customers involved in mPDP are classified to $\frac{N}{2}$ (i.e., $\frac{|\mathcal{V}|-1}{2}$) pairs of corresponding pickup customers and delivery customers. For $2 \le i\le\frac{N}{2}+1$, $v_i$ and $v_{i+\frac{N}{2}}$ are a pair of pickup and delivery customers. There are two additional constraints defined for mPDP, pickup customers must get visited prior to their corresponding delivery customers (i.e., prior constraints), and each route must only contain paired pickup and delivery customers (i.e., pairing constraint) \citep{lu2004exact}. The MILP formulation of mPDP is given as follows:
\begin{align}
\label{mPDP_form}
&\text{minimize}&& L&&\\
&\text{subject to.}& & \sum_{i \in \mathcal{V}}\sum_{j \in \mathcal{V}} d_{ij}x_{ijm} \leq L, && \forall m\in \{1,\ldots,M\}, i\neq j, & \\
&& & \sum_{j \in \mathcal{V}, i \neq j} x_{ijm} = 1, && \forall m \in  \{1,\ldots,M\}, \forall i \in \mathcal{D},& \\
&& &  \sum_{i \in \mathcal{V}, j\neq i} \sum_{m \in  \{1,\ldots,M\}} x_{ijm} = 1, && \forall j \in \mathcal{V} \setminus \mathcal{D},&\\
&& & \sum_{i \in \mathcal{V}, i\neq j} x_{ijm}  -  \sum_{h \in \mathcal{V}, h\neq j} x_{jhm} = 0, &&\forall m \in  \{1,\ldots,M\},\forall j \in \mathcal{V} \setminus \mathcal{D},&\\
&& & u_{im} - u_{jm} + |\mathcal{V}| x_{ijm} \leq |\mathcal{V}|-1, && \forall m \in  \{1,\ldots,M\}, i,j \in \mathcal{V} \setminus \mathcal{D}, i \neq j,&\\
&& & 0 \leq u_{im} \leq |\mathcal{V}|-1 && \forall m \in  \{1,\ldots,M\}, i \in \mathcal{V} \setminus \mathcal{D} ,&\\
&& & x_{ijm} \in \{0,1\}, && \forall m \in  \{1,\ldots,M\}, i,j \in \mathcal{V},& \\
&& & u_{v_i m} \le u_{v_{i+\frac{N}{2}}m}, && \forall m \in  \{1,\ldots,M\},\  2 \le i\le\frac{N}{2}+1,& \label{eq:mpdp:const1}\\
&& & (u_{v_i m}-1)(u_{v_{i+\frac{N}{2}}m}-1)> 0 , && \forall m \in  \{1,\ldots,M\}, 2 \le i\le\frac{N}{2}+1,& \label{eq:mpdp:const2}
\end{align}

where Eq. \eqref{eq:mpdp:const1} is the prior constraint \citep{lu2004exact}, which forces the pickup node to be visited before its delivery node. Eq. \eqref{eq:mpdp:const2} is the pairing constraint, which ensures paired pickups and deliveries cannot occur in different routes. 

In min-max mPDPs, the navigation task considers the optimization of a PDP solution to sub-graphs, and the sub-graphs in the partition task are additionally subject to a validity constraint as follows:

\begin{equation}
    \begin{aligned}
        &\underset{\theta}{\text{minimize}} &&\lambda\\
        &\text{subject to.} &&L(\text{PDP}(\cdot|\mathcal{G}^{i}))\le \lambda, \forall i\in\{1,\ldots, M\},\\
        & &&\bigcap_{i\in\{1,\ldots, M\}}\mathcal{G}^i=\text{Depots},\quad\bigcup_{i\in\{1,\ldots, M\}}\mathcal{G}^i=\mathcal{G},\\
        & &&\{\mathcal{G}^1,\ldots,\mathcal{G}^M\}=P_{\theta,M}(\mathcal{G}),\\
        & && \text{PDP}(\cdot|\mathcal{G}^{i})\ne \emptyset\quad \forall i \in \{1,\ldots, M\},
    \end{aligned}
\end{equation}
where the last constraint means that sub-graphs must be feasible to be solved with at least one valid PDP route.
% & && \mathcal{G}^i=\{\mathcal{D},\mathcal{\mathcal{V}}_p^i,\mathcal{\mathcal{V}}_d^i\},\quad \forall i \in \{1,\ldots, M\}\\
% & && |\mathcal{\mathcal{V}}_p^i|=|\mathcal{\mathcal{V}}_d^i|,\quad \forall i \in \{1,\ldots, M\}\\
% & && v_{j+\frac{|\mathcal{V}|-1}{2}}\in \mathcal{\mathcal{V}}_d^i,\quad \forall i \in \{1,\ldots, M\}\forall j \in\mathcal{\mathcal{V}}_p^i\\
% & && v_{j-\frac{|\mathcal{V}|-1}{2}}\in \mathcal{\mathcal{V}}_p^i,\quad \forall i \in \{1,\ldots, M\}\forall j \in\mathcal{\mathcal{V}}_d^i\\

\newpage

\subsection{Min-max Multi-depot Vehicle Routing Problem}
\label{sub:MDVRP_form}
Multi-depot VRP is a multi-depot extension of mTSP where the agent of each route starts from an arbitrary depot in $D$ and must finally end at the selected starting depot. In addition, the set $\mathcal{Q}$ is additionally defined and $\mathcal{Q}_i$ indicates the set of agent (i.e., route) indexes assigned to the depot $i$. The MILP formulation is provided based on the description in \citet{kim2022learning}, which is as follows:
\begin{align}
\label{mDVRP_form}
&\text{minimize}&& L&&\\
&\text{subject to.}& & \sum_{i \in \mathcal{V}}\sum_{j \in \mathcal{V}}   d_{ij}x_{ijm} \leq L,  && \forall k\in   \{1,\ldots,M\} , i\neq j, & \\
&& & \sum_{j \in \mathcal{V} j\neq i} \sum_{m \in  \{1,\ldots,M\}} x_{ijm} = 1, &&\forall i \in \mathcal{V} \setminus \mathcal{D},& \\
&& &  \sum_{i \in \mathcal{V} j\neq i} \sum_{m \in  \{1,\ldots,M\}} x_{ijm} = 1, && \forall j \in \mathcal{V} \setminus \mathcal{D},& \\
&& & \sum_{i \in \mathcal{V} } x_{ijm}  -  \sum_{h \in \mathcal{V} } x_{jhm} = 0, &&    \forall m \in  \{1,\ldots,M\},\forall j \in \mathcal{V} \setminus \mathcal{D}, & \\
&& & u_{im} - u_{jm} + (|\mathcal{V}|-|\mathcal{D}|+1) x_{ijm} \leq |\mathcal{V}|-|\mathcal{D}|, && \forall m \in   \{1,\ldots,M\}, i,j \in \mathcal{V} \setminus \mathcal{D} , i \neq j,& \\
&& & 0 \leq u_{im} \leq |\mathcal{V}|-|\mathcal{D}|, && \forall m \in   \{1,\ldots,M\}, i \in \mathcal{V} \setminus \mathcal{D} ,& \\
&& & x_{ijm} \in \{0,1\}, && \forall m \in   \{1,\ldots,M\},  i,j \in \mathcal{V},& \\
&& & u_{im} \in \mathbb{Z}, && \forall m \in   \{1,\ldots,M\}, i \in \mathcal{V} ,&  \\
&& & \mathcal{Q}_{i} \subseteq \{1,2,\ldots,M\}, && \forall i \in \mathcal{D} ,&\\
&& & \sum_{j \in \mathcal{V} \setminus \mathcal{D}}  x_{ijm} \leq 1, && \forall m \in \mathcal{Q}_{i} ,\forall i \in \mathcal{D} ,&\label{eq:mdvrp:const5} \\
&& & \sum_{i \in \mathcal{V} \setminus \mathcal{D}}  x_{ijm} \leq 1, && \forall m \in \mathcal{Q}_{j} ,\forall j \in \mathcal{D}, & \label{eq:mdvrp:const6}
\end{align}
where Eq. \eqref{eq:mdvrp:const5} and Eq. \eqref{eq:mdvrp:const6} indicate that each vehicle starts and returns its depot at most once. 

As a multi-depot extension of min-max mTSP, min-max MDVRP maintains consistent partition and navigation requirements compared to the formulations in min-max mTSP (Eq. \eqref{partition-mtsp} and Eq. \eqref{navigation-mtsp}) \citep{carlsson2009solving}.

\newpage

\subsection{Min-max Flexible Multi-depot Vehicle Routing Problem}
\label{appendix:fmdvrp}
Flexible MDVRP is an extension of MDVRP, allowing the vehicle to return to any depot in $\mathcal{D}$. \citet{kim2022learning} also provides the FMDVRP formulation by extending the MDVRP formulation (\ref{sub:MDVRP_form}).
To account for the flexibility of depot returning, researchers introduce a dummy node for all depots and afterward, a depot is modeled with a start depot and an end depot. Besides $\mathcal{Q}$, min-max FMDVRP further uses $\mathcal{D}_s$ and $\mathcal{D}_e$ to represent the set of start depots and end depots and $s_m$ represents the start node of the agent $m\in \{1,\ldots,M\}$.
\begin{align}
\label{FMDVRP_form}
&\text{minimize}&& L&&\\
&\text{subject to.}& & \sum_{i \in \mathcal{V}}\sum_{j \in \mathcal{V}}   d_{ij}x_{ijm} \leq L,  && \forall k\in   \{1,\ldots,M\} , i\neq j, & \\
&& & \sum_{j \in \mathcal{V} j\neq i} \sum_{m \in  \{1,\ldots,M\},} x_{ijm} = 1, &&\forall i \in \mathcal{V} \setminus \mathcal{D},& \\
&& &  \sum_{i \in \mathcal{V} j\neq i} \sum_{ m \in  \{1,\ldots,M\},} x_{ijm} = 1, && \forall j \in \mathcal{V} \setminus \mathcal{D},& \\
&& & \sum_{i \in \mathcal{V} } x_{ijm}  -  \sum_{h \in \mathcal{V} } x_{jhm} = 0, && \forall m \in  \{1,\ldots,M\} ,   \forall j \in \mathcal{V} \setminus \mathcal{D},& \\
&& & u_{im} - u_{jm} + (|\mathcal{V}|-|\mathcal{D}|+1) x_{ijm} \leq |\mathcal{V}|-|\mathcal{D}|, && \forall m \in   \{1,\ldots,M\}, i,j \in \mathcal{V} \setminus \mathcal{D} , i \neq j,& \\
&& & 0 \leq u_{im} \leq |\mathcal{V}|-|\mathcal{D}|, && \forall m \in   \{1,\ldots,M\}, i \in \mathcal{V} \setminus \mathcal{D} ,& \\
&& & x_{ijm} \in \{0,1\}, && \forall m \in   \{1,\ldots,M\},  i,j \in \mathcal{V},& \\
&& & u_{im} \in \mathbb{Z}, && \forall m \in   \{1,\ldots,M\}, i \in \mathcal{V} ,&  \\
&& & \mathcal{Q}_{i} \subseteq \{1,2,\ldots,M\}, && \forall i \in \mathcal{Q} ,&\\
&& & \sum_{j \in \mathcal{V} \setminus \mathcal{D}}  x_{s_{m}jm} = 1, && \forall m \in \{1,\ldots,M\}  \label{eq:fmdvrp:const1},\\
&& & \sum_{j \in \mathcal{V} \setminus \mathcal{D}}  x_{ijm} = 0, && \forall i \in \mathcal{D} \setminus s_{m} , \forall m \in \{1,\ldots,M\}, \label{eq:fmdvrp:const2}\\
&& & \sum_{j \in \mathcal{V} \setminus \mathcal{D}}  x_{ijm} \leq 1, && \forall m \in \mathcal{Q}_{i} ,\forall i \in \mathcal{D}_s, &\label{eq:fmdvrp:const3} \\
&& & \sum_{i \in \mathcal{V} \setminus \mathcal{D}}  x_{ijm} \leq 1, && \forall m \in \mathcal{Q}_{j} ,\forall j \in \mathcal{D}_e, & \label{eq:fmdvrp:const4}\\
&& & \sum_{j \in \mathcal{V} \setminus \mathcal{D}}  x_{ijm} = 0, && \forall m \in \{1,\ldots,M\} , \forall i \in \mathcal{D}_e, &\label{eq:fmdvrp:const5}\\
&& & \sum_{j \in \mathcal{V} \setminus \mathcal{D}}  x_{ijm} = 0, && \forall m \in \{1,\ldots,M\} , \forall i \in \mathcal{D}_s, &\label{eq:fmdvrp:const6}\\
&& & \sum_{i \in \mathcal{D}_s} \sum_{j \in \mathcal{V} \setminus \mathcal{D}}  x_{ijm} = \sum_{i \in \mathcal{V} \setminus \mathcal{D}} \sum_{j \in \mathcal{D}_e}  x_{ijm}, && \forall m \in \{1,\ldots,M\}, && \label{eq:fmdvrp:const7}
\end{align}
where Eq. \eqref{eq:fmdvrp:const1} and Eq. \eqref{eq:fmdvrp:const2} indicate each vehicle starts at its depot. Eq. \eqref{eq:fmdvrp:const3} to Eq. \eqref{eq:fmdvrp:const6} indicate constraints about start and end depots. Eq. \eqref{eq:fmdvrp:const7} indicates the balance equation of the start depots and end depots.

Min-max FMDVRP inherits the partition function in Eq. \eqref{partition-mtsp} but changes the calculation of length function $L(\boldsymbol{\tau^i})$ in Eq. \eqref{navigation-mtsp} as follows due to there is no edge from the start depots to the end depots in each route:
\begin{align}
    L'(\boldsymbol{\tau}^i)=\sum_{i=2}^{|\boldsymbol{\tau}^i|}\Vert \boldsymbol{x}_i-\boldsymbol{x}_{i-1}\Vert_2.
\end{align}

\newpage

\section{DPN: Details}\label{Methods2}

This section provides additional details of the DPN method, including the motivation of the P\&N Encoder, the decoder structure for various min-max VRPs, the P\&N Encoder for multi-depot min-max VRPs, and a supplemental illustration of the adopted Rotation-based PE.

\subsection{Motivation of P\&N Encoder}\label{modelstructure}

The \textbf{encoder in Equity-Transformer} calculate layers of multi-head self-attentions $\text{MHA}(\text{Concat}[H_a^{(0)},H_c^{(0)}],\\\text{Concat}[H_a^{(0)},H_c^{(0)}])$ to the agent initial embedding $H_a^{(0)}$ and the customer initial embedding $H_c^{(0)}$. Ignoring the formulas of the multi-head mechanism, the attention score $\alpha$ in the $l$-th layer \textbf{encoder of Equity-Transformer} consists of the following four parts of relations (i.e., agent-agent, agent-customer, customer-agent, and customer-customer from upper left to bottom right):
\begin{equation}
    \begin{aligned}
    \alpha &=\text{Concat}[H_a^{(l)},H_c^{(l)}]W_Q\big(\text{Concat}[H_a^{(l)},H_c^{(l)}]W_K\big)^{\intercal}\\
    &=\begin{bmatrix}
    H_a^{(l)}W_Q(H_a^{(l)}W_K)^{\intercal} \in \mathcal{R}^{(M+1)\times (M+1)} & & H_a^{(l)}W_Q(H_c^{(l)}W_K)^{\intercal} \in \mathcal{R}^{(M+1)\times N}\\
    H_c^{(l)}W_Q(H_a^{(l)}W_K)^{\intercal} \in \mathcal{R}^{N\times (M+1)} & &H_c^{(l)}W_Q(H_c^{(l)}W_K)^{\intercal} \in \mathcal{R}^{N\times N}\\
    \end{bmatrix}.
    \end{aligned}
    \label{mat}
\end{equation}

In any single column of the attention score matrix, the embeddings of agents and customers are fused. Considering that agent embeddings mainly carry the features for the customer partition task and customer embeddings stand for navigation requirements. The normalization from the Softmax function may incline customers to focus on either the partition task or the navigation task thus resulting in less effective features for partition and navigation. Moreover, $M+1$ agent embeddings are inherently similar in values (i.e., only differ in PEs), so the self-attention calculation $H_a^{(l)}W_Q(H_a^{(l)}W_K)^{\intercal}$ might become inexplicable noise. 
\begin{figure}[H]
    \centering
    \subfigure[Before Softmax]{\includegraphics[width = 0.32\textwidth]{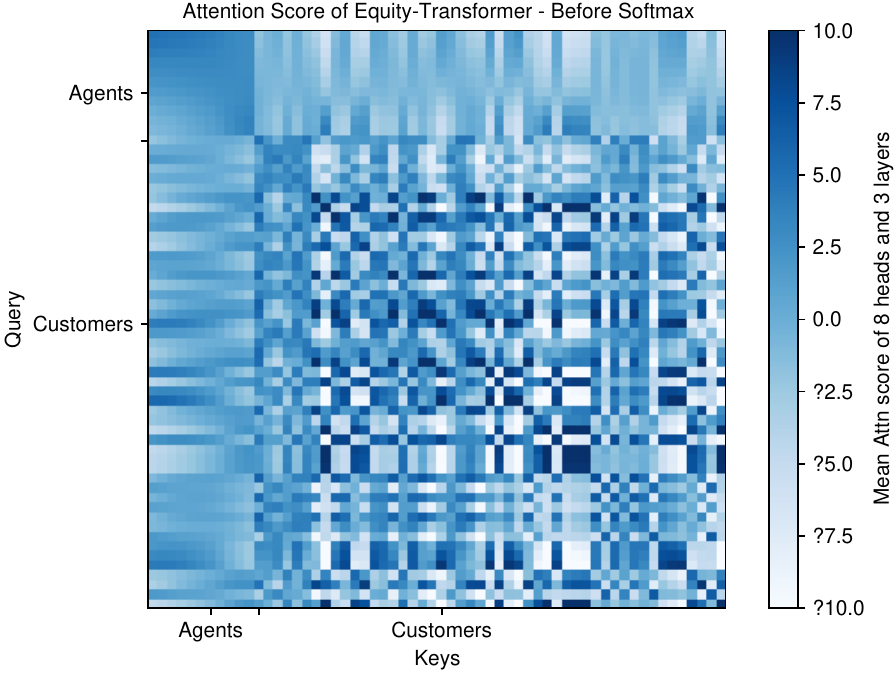}}
    \quad\quad\quad\quad
    \subfigure[After Softmax]{\includegraphics[width = 0.32\textwidth]{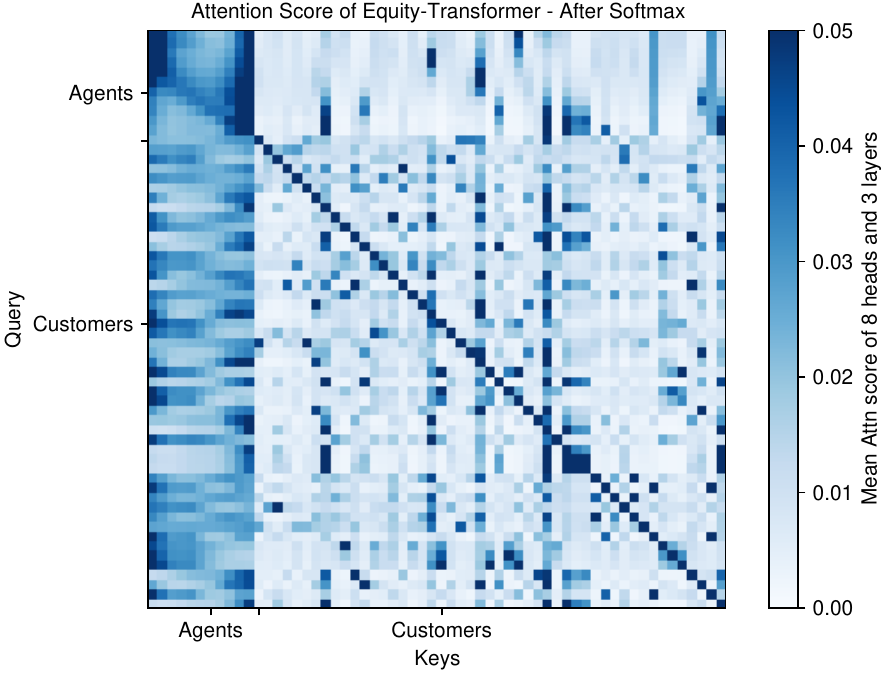}}
    \caption{Attention score of min-max mTSP50 ($M$=10), each data is the average score of 8 heads and 3 attention layers.}
    \label{fig:attn}
\end{figure}

Figure \ref{fig:attn} shows the attention score $\alpha$ in Equity-Transformer and the attention score normalized by the Softmax function processing a min-max mTSP instance with $M$=10. Refer to Eq. \eqref{mat}, the $H_a^{(l)}W_Q(H_a^{(l)}W_K)^{\intercal}$ (upper left) part displays irregular noise, which also affects the calculation of the upper right part by a Softmax normalization. Moreover, the Softmax function amplifies the attention score of the $H_c^{(l)}W_Q(H_a^{(l)}W_K)^{\intercal}$ part (bottom left), which originally lies around 0, thereby confusing the calculation of both the customer-customer relations (bottom right) and the customer-agent relations (bottom left). To ensure the independence of the three necessary parts of relations (bottom left, bottom right, and upper right), our P\&N Encoder designs a bi-part structure that calculates Softmax and the three correlations separately.

\textbf{Complexity.} In addition, the attention mechanisms in P\&N Encoder do not increase the time complexity of each layer, slightly reducing from $\mathcal{O}\big((N+M+1)^2d\big)$ to $\mathcal{O}\big((N^2+2NM)d\big)$. The practical training and testing time generally remain the same considering the impact of additional feed-forward layers. Furthermore, the time complexity of the encoder is not the bottleneck of overall time complexity in sequential planning methods \citep{bello2017neural}. The space complexity is also reduced from $\mathcal{O}\big((N+M+1)^2+(N+M+1)d\big)$ to $\mathcal{O}\big(N^2+2NM+(N+M)d\big)$, and together with the help of the APS-Loss, our DPN can adopt larger batch sizes in training compared to existing sequential planning methods.

\newpage
\subsection{Model Structure for Min-max mTSP}
The P\&N Encoder for min-max mTSP has been described in detail in the main text. The $L$-layer P\&N Encoder processes the agent embeddings $H_a^{(0)}$ and the customer embeddings $H_c^{(0)}$ to $H_a^{(L)}$ and $H_c^{(L)}$, respectively. Sequential planning methods use contextual embedding to process the information from the instance $\mathcal{G}$, the number of routes $M$, and the previously generated routes. For $H_a^{(L)}=[\boldsymbol{h}_{a,1}^{(L)},\boldsymbol{h}_{a,2}^{(L)},\ldots,\boldsymbol{h}_{a, M}^{(L)}]$ and $H_c^{(L)}=[\boldsymbol{h}_{c,1}^{(L)},\boldsymbol{h}_{c,2}^{(L)},\ldots,\boldsymbol{h}_{c, N}^{(L)}]$, in decoding the $m$-th route ($m\le M$) and with $n$ customers left, if the is current agent index is $o_c$ and the current node index is $node_c$, the contextual embedding $\boldsymbol{h}_{contex}\in \mathcal{R}^{d}$ is calculated as follows:
\begin{equation}
    \begin{aligned}
        \boldsymbol{h}_{contex}  =& \frac{W_{emb}}{N+M}\big(\sum_{i=1}^{M}\boldsymbol{h}_{a,i}^{(L)}+\sum_{i=1}^{N}\boldsymbol{h}_{m,i}^{(L)}\big)\\
        &+\text{Concat}\big[\boldsymbol{h}_{a,o_c}^{(L)},\boldsymbol{h}_{c,node_c}^{(L)},\frac{M-m+1}{M},\frac{n}{N}\big]W_{step}\\
        &+\text{Concat}\big[L'(\boldsymbol{\tau}^{m+1}),
        \underset{i\in\{1,\ldots,N\}}{\text{max}}\Vert \boldsymbol{x}_{d}-\boldsymbol{x}_i\Vert_2,
        \text{LD}\big]W_{length},
    \end{aligned}
\end{equation}
where LD represents the maximal distance to the depot among all the $n$ left customers. $W_{emb} \in \mathcal{R}^{d\times d}$, $W_{step} \in \mathcal{R}^{2d+2\times d}$, $W_{length} \in \mathcal{R}^{3\times d}$ are three projection matrices. When decoding the first action, $node_c$ is set to the depot in min-max mTSP. In training, DPN generates $K$ permutations of agents and concat the $K$ $\boldsymbol{h}_{contex}$ to $H_{contex}\in \mathcal{R}^{K\times d}$. Then the decoder of DPN contains a single 8-head glimpse attention-layer \citep{bello2017neural} which calculates the mutual attention between contextual embeddings, agent embeddings, and customer embeddings, the attention result $\boldsymbol{q}_{contex}$ for the following logit calculation is 
\begin{equation}
    \begin{aligned}
        \boldsymbol{q}_{contex}=\text{MHA}(H_{contex},\text{Concat}\big[\boldsymbol{h}^{(L)}_{a,1},\ldots,\boldsymbol{h}^{(L)}_{a,M},\boldsymbol{h}^{(L)}_{c,1},\ldots,\boldsymbol{h}^{(L)}_{c,N}\big]),
    \end{aligned}
\end{equation}
where $\boldsymbol{q}_{contex}\in \mathcal{R}^{K\times d}$. Afterward, given the last selected node $node_c$, the final probability $\pi_\theta(a_t|s_t,c)$ is computed as follows \citep{wang2024distance}:

\begin{equation}
    \begin{aligned}
    \text{dist}(node_c,j) = \alpha_d *\text{exp}\big(\frac{\Vert \boldsymbol{x}_{node_c}-\boldsymbol{x}_j\Vert_2}{\underset{i\in\{1,\ldots,N\}}{\text{max}}\Vert \boldsymbol{x}_i-\boldsymbol{x}_{node_c}\Vert_2}\big),
    \end{aligned}
    \label{prob}
\end{equation}
\begin{equation}
\begin{aligned}
    u_j&=\left\{
    \begin{aligned}
    & C\cdot \text{tanh}\Big(\frac{\boldsymbol{q}_{contex}(\boldsymbol{h}_jW^L)^{\intercal}}{\sqrt{d_k}}+\text{dist}(node_c,j)\Big) \odot M_a &j\in \mathcal{V} \cap j\notin  s_t\\
    & -\infty &\text{otherwise}
    \end{aligned}
    \right.,
    \end{aligned}
\end{equation}
where $j\notin  s_t$ represents the unvisited condition. $W_L$ is a learnable square projection matrix and $M_a$ is an agent feasibility mask, it will be $-\infty$ if the agent is not $o_c$, and 1 for unmasked actions. It will also mask the agent $o_c$ when generating the last route. $\alpha_d$ is a trainable parameter, C is set to 50, and the above probability is calculated invariantly in all min-max VRPs. For the single-depot min-max VRPs, selecting the first $M$ indexes corresponds to selecting the depot.

\subsection{Model Structure for Min-max mPDP}

Due to the special constraints introduced by the partition and navigation tasks of PDP (Appendix \ref{sub:mPDP_form}), in the navigation part for min-max mPDP, we adopt the Heterogeneous Attention \citep{li2021heterogeneous} instead of self-attention between customers. For the partition part of the P\&N Encoder, we use different linear projections to process pickup and delivery customers in the calculations of queries (i.e., Eq. \eqref{place2}). In the decoder, the contextual embedding is calculated as follows:
\begin{equation}
    \begin{aligned}
        \boldsymbol{h}_{contex}  =& \frac{W_{emb}}{N+M}\big(\sum_{i=1}^{M}\boldsymbol{h}_{a,i}^{(L)}+\sum_{i=1}^{N}\boldsymbol{h}_{m,i}^{(L)}\big)\\
        &+\text{Concat}\big[\boldsymbol{h}_{a,o_c}^{(L)},\boldsymbol{h}_{c,node_c}^{(L)},\frac{M-m+1}{M},\frac{2p}{N}\big]W_{step}\\
        &+\text{Concat}\big[L'(\boldsymbol{\tau}^{m+1}),\text{Longest-PD},
        \text{Longest-P},
        \text{Longest-D},\frac{\text{Sum-PD}}{M-m}\big]W_{length},
    \end{aligned}
\end{equation}

where $p$ is the number of left pickups, Longest-PD is the longest distance of visited paired pickups and deliveries in the current route, Longest-P is the maximal distance to the depot for unvisited pickups, Longest-D is the maximal distance to the depot for unvisited deliveries, and Sum-PD represents the sum of the unvisited distance of paired pickups and deliveries.

\subsection{Model Structure for Multi-depot Min-max VRPs}\label{mdvrpmodel}

\begin{figure}[H]
    \centering
    \includegraphics[width = 0.85\hsize]{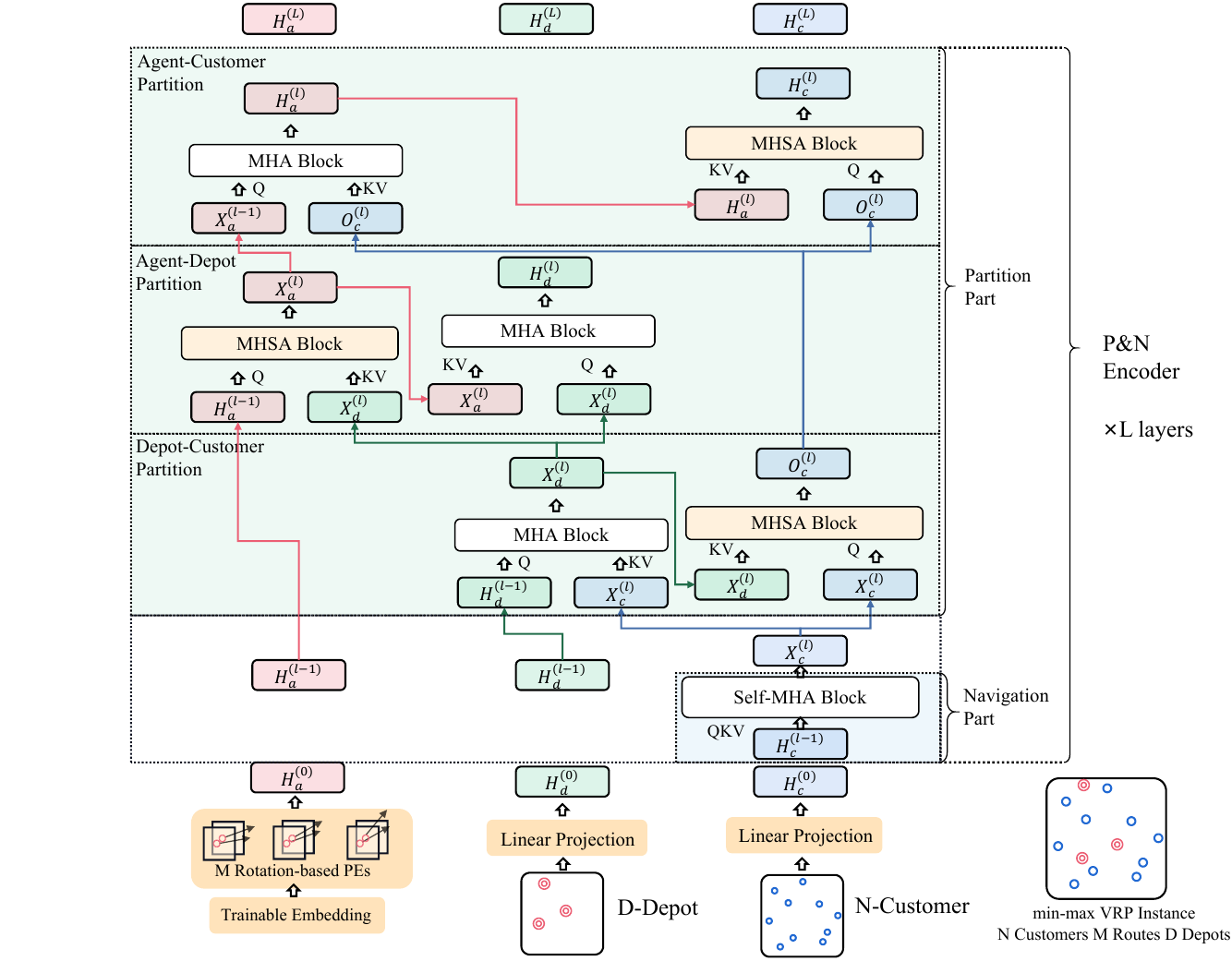}
    \caption{Model structure of the proposed P\&N Encoder in min-max MDVRP and min-max FMDVRP.}
    \label{fig:multi-net}
\end{figure}

Figure \ref{fig:multi-net} shows that P\&N Encoder designs additional partition parts to handle the $D$ distinct depots. The input to $M$ Rotation-based PEs is a trainable embedding representing a dummy depot. For initial embeddings, depot embeddings $H_d^{(0)}$, agent embeddings $H_a^{(0)}$, and customer embeddings $H_c^{(0)}$ are generated by linear projection.

The navigation part in the P\&N Encoder for multi-depot Min-max VRPs maintains the same as what for min-max mTSP as follows:
\begin{align}
    &\hat{X}_c^{(l)}=\alpha_1*\text{MHA}(H_c^{(l-1)},H_c^{(l-1)}))+H_c^{(l-1)},\\
    &X_c^{(l)}=\alpha_2*\text{FF}(\hat{X}_c^{(l)})\quad\quad\quad\quad\quad\quad+\hat{X}_c^{(l)}.
\end{align}

The partition part consists of the depot-customer partition part, the agent-depot partition part, and the agent-customer partition part, sequentially. The depot-customer partition part is as follows:
\begin{align}
    &\hat{X}_d^{(l)}=\alpha_3*\text{MHA}(H_d^{(l-1)},X_c^{(l)})+H_d^{(l-1)},\\
    &X_d^{(l)}=\alpha_4*\text{FF}(\hat{X}_d^{(l)})\quad\quad\quad\quad\ \ +\hat{X}_d^{(l)},\\
    &\hat{O}_c^{(l)}=\alpha_5*\text{MHSA}(X_c^{(l)},X_d^{(l)})\ \  +X_c^{(l)},\\
    &O_c^{(l)}=\alpha_6*\text{FF}(\hat{O}_c^{(l)})\quad\quad\quad\quad\ \ \ +\hat{O}_c^{(l)},
\end{align}
where MHSA is leveraged to assign only one (maybe two depots in min-max FMDVRP) depot to each customer. The agent-depot partition part is as follows:
\begin{align}
    &\hat{X}_a^{(1)}=\alpha_7*\text{MHSA}(H_a^{(l-1)},X_d^{(l)})+H_a^{(l-1)},\\
    &X_a^{(l)}=\alpha_8*\text{FF}(\hat{X}_a^{(l)})\quad\quad\quad\quad\quad\ +\hat{X}_a^{(l)},\\
    &\hat{H}_d^{(l)}=\alpha_9*\text{MHA}(X_d^{(l)},X_a^{(l)})\ \quad \ \ +X_d^{(l)},\\
    &H_d^{(l)}=\alpha_{10}*\text{FF}(\hat{H}_d^{(l)})\quad\quad\quad\quad \quad +\hat{H}_d^{(l)},
\end{align}
where MHSA is utilized to assign only one (maybe two depots for min-max FMDVRP) depot to each agent. The agent-customer partition part is as follows:
\begin{align}
    &\hat{X}_d^{(l)}=\alpha_{11}*\text{MHA}(X_d^{(l)},O_c^{(l)})\ \ +X_d^{(l)},\\
    &H_a^{(l)}=\alpha_{12}*\text{FF}(\hat{H}_a^{(l)})\quad\quad\quad\ \ \ \ +\hat{H}_a^{(l)},\\
    &\hat{H}_c^{(l)}=\alpha_{13}*\text{MHSA}(O_c^{(l)},H_a^{(l)}) +O_c^{(l)},\\
    &H_c^{(l)}=\alpha_{14}*\text{FF}(\hat{H}_c^{(l)})\quad\quad\quad\ \ \ \ +\hat{H}_c^{(l)},
\end{align}
where MHSA is introduced to assign only one agent to each customer. The decoder for min-max MDVRP and min-max FMDVRP only vary in the feasibility mask. If depot embeddings are $H_d^{(L)}=[\boldsymbol{h}_{d,1}^{(L)},\boldsymbol{h}_{d,2}^{(L)},\ldots,\boldsymbol{h}_{d, D}^{(L)}]$ and $x_{d_j}$ represent the coordinate of the $j$-th depot, the contextual embedding for both is as follows:
\begin{equation}
    \begin{aligned}
        \boldsymbol{h}_{contex}  =& \frac{W_{emb}}{N+M+D}\big(\sum_{i=1}^{M}\boldsymbol{h}_{a,i}^{(L)}+\sum_{i=1}^{N}\boldsymbol{h}_{m,i}^{(L)}+\sum_{i=1}^{D}\boldsymbol{h}_{d,i}^{(L)}\big)\\
        &+\text{Concat}\big[\boldsymbol{h}_{a,o_c}^{(L)},\boldsymbol{h}_{c,node_c}^{(L)},\frac{M-m+1}{M},\frac{n}{N}\big]W_{step}\\
        &+\text{Concat}\big[L'(\boldsymbol{\tau}^{m+1}),
        \underset{i\in\{1,\ldots,N\}}{\text{max}}\underset{j\in\{1,\ldots,D\}}{\text{min}}\Vert \boldsymbol{x}_{d_j}-\boldsymbol{x}_i\Vert_2,
        \text{LD}\big]W_{length},
    \end{aligned}
\end{equation}
where LD represents the maximal distance to the nearest depot among all the $n$ left customers. The $node_c$ in decoding the first action is set to a random depot. In decoder, the $\boldsymbol{q}_{contex}$ in multi-depot min-max VRPs is modified as follows:
\begin{equation}
    \begin{aligned}
        \boldsymbol{q}_{contex}=\text{MHA}(H_{contex},\text{Concat}\big[\boldsymbol{h}^{(L)}_{d,1},\ldots,\boldsymbol{h}^{(L)}_{d,D},\boldsymbol{h}^{(L)}_{c,1},\ldots,\boldsymbol{h}^{(L)}_{c,N}\big]).
    \end{aligned}
\end{equation}

\subsection{Motivation About the Rotation-base PE}\label{PEmotivation}

The sequential planning method uses positional encoding to distinguish different agents starting from the same depot. In P\&N Encoder, position encoding is first calculated in the partition part of the first layer of P\&N Encoder, ignoring the formulation of the multi-head mechanism, $H_a^{(0)}=W_ax_d+b_a+PE$, the PE is involved as follows (calculating Eq. \eqref{place}):
\begin{equation}
    \begin{aligned}
        \text{Attn}(H_a^{(0)},X_c^{(1)})=&\text{Softmax}\left( \frac{ H_a^{(0)} W_Q (X_c^{(1)} W_K)^{\intercal}}{\sqrt{d}}\right)X_c^{(1)} W_V,\\
        H_a^{(0)} W_Q (X_c^{(1)} W_K)^{\intercal}=&(W_ax_d+b_a+PE)W_Q (X_c^{(1)} W_K)^{\intercal}\\
        =&(W_ax_d+b_a)W_Q (X_c^{(1)} W_K)^{\intercal}+ PEW_Q (X_c^{(1)} W_K)^{\intercal}\\
        =&\text{position-based score}+\text{angle-based score}.\\
    \end{aligned}
\end{equation}

In the first layer, the attention calculation can be divided into a position-based score and an angle-based score. For the $M$ agents, their position-based score remains the same, so the PE is used to highlight the nodes with right and predetermined angles in calculating the angle-based score (i.e., being responsible for the angle-based customer assignments). In the subsequent blocks, residual connections enable the PE to maintain the above functions. 

\begin{figure}[H]
    \centering
    \includegraphics[width = 0.85\hsize]{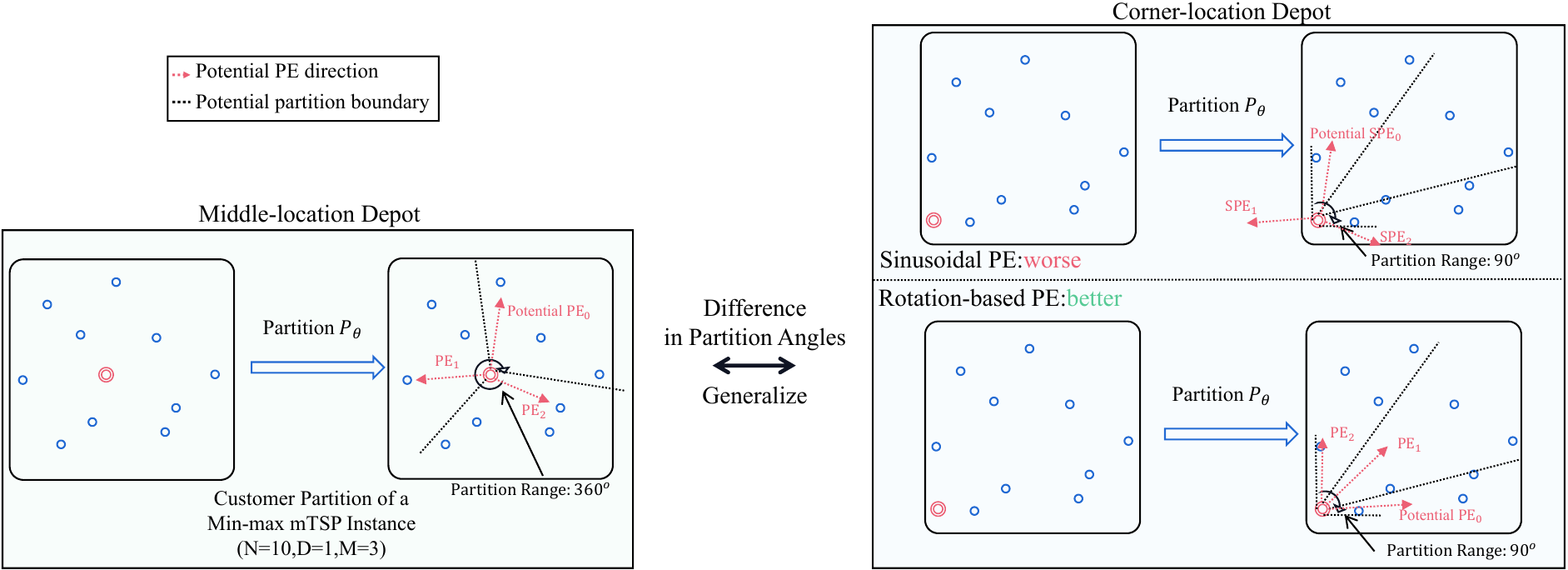}
    \caption{The PE required for the customer partition task varies to the depot locations. So the Rotation-based PE in DPN introduces the location of depots into calculation.}
    \label{fig:PEpartition}
\end{figure}

Given the number of routes $M$, sinusoidal position encoding represents a fixed angle relationship so the angle-based partition will remain the same for any instance. However, the angle-related distribution of the routes in optimal solution varies with the depot location. As shown in visual cases Figure \ref{lkh1} and Figure \ref{lkh2}, and Figure \ref{fig:PEpartition}, if the location of the depot is in the bottom left corner (e.g. the right case in Figure \ref{fig:PEpartition} and the instance in Figure \ref{lkh2}), it only needs to consider customer partition in an approximate 90-degree space. However, in other cases as Figure \ref{lkh1} and the left case in Figure \ref{fig:PEpartition}, all the directions are involved. Therefore, the sinusoidal PE meets challenges in generalizing to instances with different depot locations so it is necessary to introduce the depot coordinates to the calculation of PE. Experiments in Appendix \ref{ablationpe} demonstrate that SPE prefers instances with corner-location depot. The proposed Rotation-based positional encoding achieves better partition optimalities.

\subsection{Rotation-base PE implementation}\label{implementationpe}

The main text provides a calculation method based on complex numbers for the Rotation-based PE in DPN. In practice, we adopt a real-number implementation provided in \citet{su2023roformer}. Given $\boldsymbol{emb} = \boldsymbol{x}_d W_a + b_a, \boldsymbol{emb} \in \mathcal{R} ^{M \times d}$ and $\theta_i=\frac{1}{1,000}^{\frac{i-1}{d}}$ for $i\in\{1,\ldots,d/2\}$, $\theta_i \in \mathcal{R} ^{d/2}$, the $\text{PE}\in \mathcal{R}^{M \times d}$ in position $m$ $\text{PE}_m\in \mathcal{R}^{d}$ is as follows:

\begin{equation}
\begin{aligned}
\text{PE}_m &=(\left(\begin{array}{c}
emb_{m,1} \\
emb_{m,2} \\
emb_{m,3} \\
emb_{m,4} \\
\vdots \\
emb_{m,d-1} \\
emb_{m,d}
\end{array}\right) \otimes\left(\begin{array}{c}
\cos m \theta_1 \\
\cos m \theta_1 \\
\cos m \theta_2 \\
\cos m \theta_2 \\
\vdots \\
\cos m \theta_{d / 2-1} \\
\cos m \theta_{d / 2-1}
\end{array}\right)+\left(\begin{array}{c}
-emb_{m,2} \\
emb_{m,1} \\
-emb_{m,4} \\
emb_{m,3} \\
\vdots \\
-emb_{m,d} \\
emb_{m,d-1}
\end{array}\right) \otimes\left(\begin{array}{c}
\sin m \theta_1 \\
\sin m \theta_1 \\
\sin m \theta_2 \\
\sin m \theta_2 \\
\vdots \\
\sin m \theta_{d / 2} \\
\sin m \theta_{d / 2}
\end{array}\right))W_{PE}.
\end{aligned}
\end{equation}

The RoPE in natural language processing usually uses 10,000 as the base \citep{su2023roformer,su2024roformer}. However the number of agents in min-max VRPs usually does not exceed 100, so we adopt a smaller base of 1,000 for a shorter sequence. There is only a slight difference in the training stability between the two bases.

\subsection{Discussion: Application on General VRPs}\label{discussap}

Different from general VRPs (e.g., CVRP), min-max VRPs limit the total number of vehicles and minimize the longest route. Solvers for min-max VRPs need to process these requirements with special procedures. DPN adopts the P\&N Encoder and the APS-Loss to specifically process this constraint so the proposed DPN is unnecessary for VRPs without this constraint. Moreover, due to the existence of this constraint, together with min-max VRP emphasizing balancing the lengths between different routes, powerful heuristic approaches for min-sum problems are not well-generalized to the min-max case \citep{bertazzi2015min}, and existing constructive min-max VRP solving frameworks can not effectively generalize to general VRP \citep{son2023solving}.

We acknowledge the enormous value of a universal and effective framework for both VRPs and min-max VRPs. Therefore, as future work, we are eager to adopt the idea of DPN to design a universal constructive solver to process decoupled representations between routes and effectively handle both VRPs and min-max VRPs.

\newpage

\section{Experiments Details}

\subsection{Hyperparameters}\label{Seetings}

This paper trains 16 models on different scales and different settings. We will provide these models as pre-trained after being open-source, and their hyperparameters are listed in Table \ref{hyper}.

\begin{table}[htbp]
\caption{Training settings of the proposed DPN on different scales.}
\centering
\renewcommand\arraystretch{1}
\setlength{\tabcolsep}{8mm}
\small{
\begin{tabular}{lcc}
\hline\bottomrule
\multicolumn{3}{c}{Single-depot Training Settings}                                                                                  \\ \hline
\multicolumn{1}{l|}{Problem}                      & \multicolumn{1}{c|}{mPDP50 \&  mTSP50}        & mPDP100 \&  mTSP100        \\ \hline
\multicolumn{1}{l|}{Fintune or not}               & \multicolumn{1}{c|}{No}                   & No                     \\
\multicolumn{1}{l|}{number of agents($M$)}          & \multicolumn{1}{c|}{{[}2,10{]}}           & {[}2,10{]}             \\
\multicolumn{1}{l|}{number of depots($D$)}          & \multicolumn{1}{c|}{1}                    & 1                      \\
\multicolumn{1}{l|}{number of encoder layers (L)} & \multicolumn{1}{c|}{6}                    & 6                      \\
\multicolumn{1}{l|}{learning rate}                & \multicolumn{1}{c|}{1.00E-04}             & 1.00E-04               \\
\multicolumn{1}{l|}{learning rate decay}          & \multicolumn{1}{c|}{1}                    & 1                      \\
\multicolumn{1}{l|}{batch size}                   & \multicolumn{1}{c|}{256}                  & 256                    \\
\multicolumn{1}{l|}{epoches}                      & \multicolumn{1}{c|}{500}                  & 500                    \\
\multicolumn{1}{l|}{epoch size}                   & \multicolumn{1}{c|}{256000}               & 256000                 \\ 
\multicolumn{1}{l|}{number of permutations ($K$)}   & \multicolumn{1}{c|}{60}                   & 60                                                                                 \\ \hline
\multicolumn{1}{l|}{Problem}                      & \multicolumn{1}{c|}{mPDP200 \&  mTSP200}      & mPDP500 \&  mTSP500        \\ \hline
\multicolumn{1}{l|}{Fintune or not}               & \multicolumn{1}{c|}{From 100}             & From 100               \\
\multicolumn{1}{l|}{number of agents($M$)}          & \multicolumn{1}{c|}{{[}10,20{]}}          & {[}30,50{]}            \\
\multicolumn{1}{l|}{number of depots($D$)}          & \multicolumn{1}{c|}{1}                    & 1                      \\
\multicolumn{1}{l|}{number of encoder layers (L)} & \multicolumn{1}{c|}{6}                    & 6                      \\
\multicolumn{1}{l|}{learning rate}                & \multicolumn{1}{c|}{1.00E-05}             & 1.00E-05               \\
\multicolumn{1}{l|}{learning rate decay}          & \multicolumn{1}{c|}{1}                    & 1                      \\
\multicolumn{1}{l|}{batch size}                   & \multicolumn{1}{c|}{64}                   & 16                     \\
\multicolumn{1}{l|}{epoches}                      & \multicolumn{1}{c|}{20}                   & 20                     \\
\multicolumn{1}{l|}{epoch size}                   & \multicolumn{1}{c|}{64000}                & 16000                  \\
\multicolumn{1}{l|}{number of permutations ($K$)}   & \multicolumn{1}{c|}{60}                   & 60                     \\ \hline
\multicolumn{3}{c}{Multi-depot Training Settings}                                                                                  \\ \hline
\multicolumn{1}{l|}{Problem}                      & \multicolumn{1}{c|}{MDVRP50 \&  FMDVRP50}     & MDVRP100 \&  FMDVRP100     \\ \hline
\multicolumn{1}{l|}{Fintune or not}               & \multicolumn{1}{c|}{No}                   & No                     \\
\multicolumn{1}{l|}{number of agents($M$)}          & \multicolumn{1}{c|}{{[}2,10{]}}           & {[}2,10{]}             \\
\multicolumn{1}{l|}{number of depots($D$)}          & \multicolumn{1}{c|}{{[}2,10{]}}           & {[}2,10{]}             \\
\multicolumn{1}{l|}{number of encoder layers (L)} & \multicolumn{1}{c|}{6}                    & 6                      \\
\multicolumn{1}{l|}{learning rate}                & \multicolumn{1}{c|}{1.00E-04}             & 1.00E-04               \\
\multicolumn{1}{l|}{learning rate decay}          & \multicolumn{1}{c|}{1}                    & 1                      \\
\multicolumn{1}{l|}{batch size}                   & \multicolumn{1}{c|}{256}                  & 256                    \\
\multicolumn{1}{l|}{epoches}                      & \multicolumn{1}{c|}{500}                  & 500                    \\
\multicolumn{1}{l|}{epoch size}                   & \multicolumn{1}{c|}{256000}               & 256000                 \\
\multicolumn{1}{l|}{number of permutations ($K$)}   & \multicolumn{1}{c|}{60}                   & 60                                                                                    \\ \hline
\multicolumn{1}{l|}{Problem}                      & \multicolumn{1}{c|}{MDVRP50-F \&  FMDVRP50-F} & MDVRP100-F \&  FMDVRP100-F \\ \hline
\multicolumn{1}{l|}{Fintune or not}               & \multicolumn{1}{c|}{From 50}              & From 100               \\
\multicolumn{1}{l|}{number of agents($M$)}          & \multicolumn{1}{c|}{{[}3,7{]}}            & {[}5,10{]}             \\
\multicolumn{1}{l|}{number of depots($D$)}          & \multicolumn{1}{c|}{8}                    & 8                      \\
\multicolumn{1}{l|}{number of encoder layers (L)} & \multicolumn{1}{c|}{6}                    & 6                      \\
\multicolumn{1}{l|}{learning rate}                & \multicolumn{1}{c|}{1.00E-05}             & 1.00E-05               \\
\multicolumn{1}{l|}{learning rate decay}          & \multicolumn{1}{c|}{1}                    & 1                      \\
\multicolumn{1}{l|}{batch size}                   & \multicolumn{1}{c|}{128}                  & 128                    \\
\multicolumn{1}{l|}{epoches}                      & \multicolumn{1}{c|}{20}                   & 20                     \\
\multicolumn{1}{l|}{epoch size}                   & \multicolumn{1}{c|}{128000}               & 128000                 \\
\multicolumn{1}{l|}{number of permutations ($K$)}   & \multicolumn{1}{c|}{60}                   & 60                     \\ \toprule\hline
\end{tabular}}\label{hyper}
\end{table}

\subsection{Datasets}\label{datasets}
There are a total of 10 100-instance datasets used in min-max mTSP and min-max mPDP in Table \ref{50100} and Table \ref{more}. We used the provided 100-instance test set in \citet{son2023solving} for these datasets. \citet{son2023solving} also provides the results of closed-source methods ScheduleNet and NCE with their provided datasets. So the Obj. of these two learning-based methods provided in Table \ref{more} is accurate. We also generate a 100-instance validation size for the validation curve in Figure \ref{fig:curve} and Figure \ref{fig:curve10}. The testing dataset for multi-depot min-max VRPs is also uniformly generated. In addition, Table \ref{50100} also shows the number of customers $N$ and depots $D$ corresponding to each dataset. The use of $N+D$ to represent scale in mTSP problems comes from previous works \citep{cao2021dan,son2023solving,mahmoudinazlou2024hybrid}.

\subsection{Training Time}
The detailed training time of DPN on four involved min-max VRPs are listed in Table \ref{time}. All 100-node training tasks can be done within 5 days.
\begin{table}[H]
\centering
\caption{Training time of DPN on a single Nvidia Tesla V100S GPU.}
\renewcommand\arraystretch{1.1}
\setlength{\tabcolsep}{4mm}
\small{
\begin{tabular}{l|cccc}\hline\bottomrule
Problem    & min-max mTSP100 & min-max mPDP100 & min-max MDVRP100 & min-max FMDVRP100 \\\hline
Epoch Time & 11.0min         & 11.2min         & 12.6min          & 12.6min           \\
Total Time & 91.7h           & 93.3h           & 105h             & 105h      \\ \toprule\hline  
\end{tabular}}\label{time}
\end{table}

\section{More Experiments}\label{Experiment2}

\subsection{Generalization on Benchmark: Min-max mTSP SetI}\label{set1}
We use the benchmark mTSP SetI with less than 500 nodes (i.e., 50- to 500-scale) to validate the generalization ability of DPN. The performances of heuristic algorithms and the best-known solution (BKS) are reported in LKH3 and HGA \citep{mahmoudinazlou2024hybrid}, We use a special fine-tuned versions on min-max mTSP200 for both the Equity-Transformer and DPN. They follow the same setting of the normal min-max mTSP200 fine-tuned model except for setting the range of agent number $M$ to $[3,20]$. As shown in Table \ref{mtspset}, as neural solvers, the proposed DPN-$\times$8aug and DPN-$\times$8aug-$\times$16per versions narrow the optimal gap compared to the Equity-Transformer-$\times$8aug.

\begin{table}[htbp]
\caption{Performances (objective function values and gaps to the BKS) of DPN on mTSP SetI datasets with less than 500 nodes.}
\centering
\renewcommand\arraystretch{1}
\setlength{\tabcolsep}{2.7mm}
\small{
\begin{tabular}{c|c|ccc|cccc}
\hline\toprule
           &    & BKS   & LKH-3  & HGA    & Equity-Transformer-$\times$8aug & DPN-$\times$8aug    & \multicolumn{2}{c}{DPN-$\times$8aug-$\times$16per} \\ \cline{3-9} 
Instances  & $M$= & Obj.  & Obj.   & Obj.   & Obj.               & Obj.   & Obj.       & Gap        \\ \hline
mtsp51     & 3  & 160   & 160    & 160    & 201                & 174    & 173        & 8.30\%     \\
           & 5  & 118   & 118    & 118    & 125                & 120    & 120        & 1.91\%     \\
           & 10 & 112   & 112    & 112    & 112                & 112    & 112        & 0.00\%     \\ \hline
mtsp100    & 3  & 8509  & 8509   & 8509   & 10411              & 9835   & 9759       & 14.69\%    \\
           & 5  & 6766  & 6766   & 6771   & 7929               & 7318   & 7279       & 7.59\%     \\
           & 10 & 6358  & 6358   & 6358   & 6454               & 6365   & 6359       & 0.00\%     \\
           & 20 & 6358  & 6358   & 6358   & 6358               & 6358   & 6358       & 0.00\%     \\ \hline
rand100    & 3  & 3032  & 3032   & 3032   & 3638               & 3196   & 3196       & 5.43\%     \\
           & 5  & 2410  & 2410   & 2410   & 2537               & 2477   & 2453       & 1.79\%     \\
           & 10 & 2299  & 2299   & 2299   & 2299               & 2299   & 2299       & 0.00\%     \\
           & 20 & 2299  & 2299   & 2299   & 2299               & 2299   & 2299       & 0.00\%     \\ \hline
mtsp150    & 3  & 13038 & 13038  & 13093  & 15678              & 14993  & 14929      & 14.50\%    \\
           & 5  & 8450  & 8450   & 8487   & 9993               & 9746   & 9695       & 14.73\%    \\
           & 10 & 5557  & 5557   & 5587   & 6404               & 6196   & 6196       & 11.50\%    \\
           & 20 & 5246  & 5246   & 5246   & 5263               & 5285   & 5248       & 0.03\%     \\ \hline
gtsp150    & 3  & 2402  & 2402   & 2402   & 2740               & 2551   & 2508       & 4.44\%     \\
           & 5  & 1741  & 1741   & 1741   & 1950               & 1823   & 1823       & 4.71\%     \\
           & 10 & 1555  & 1555   & 1555   & 1562               & 1556   & 1555       & 0.00\%     \\
           & 20 & 1555  & 1555   & 1555   & 1555               & 1555   & 1555       & 0.00\%     \\ \hline
kroA200    & 3  & 10691 & 10691  & 10700  & 13342              & 13004  & 12924      & 20.89\%    \\
           & 5  & 7412  & 7412   & 7420   & 8971               & 8755   & 8754       & 18.10\%    \\
           & 10 & 6223  & 6223   & 6223   & 6382               & 6243   & 6223       & 0.00\%     \\
           & 20 & 6223  & 6223   & 6223   & 6223               & 6223   & 6223       & 0.00\%     \\ \hline
lin318     & 3  & 15701 & 15701  & 15714  & 19514              & 16879  & 16879      & 7.50\%     \\
           & 5  & 11289 & 11289  & 11297  & 13763              & 12292  & 12265      & 8.65\%     \\
           & 10 & 9731  & 9731   & 9731   & 10129              & 9765   & 9738       & 0.07\%     \\
           & 20 & 9731  & 9731   & 9731   & 9731               & 9731   & 9731       & 0.00\%     \\ \hline
Gap to BKS &    & -     & 0.00\% & 0.07\% & 10.26\%            & 5.67\% & 5.36\%     &            \\ \bottomrule\hline
\end{tabular}}
\label{mtspset}
\end{table}

\subsection{Generalization on Benchmark: Min-max mTSP Lib}
We also adopt the widely used \citep{kim2022learning} mTSP Lib \citep{reinelt1991tsplib} dataset for testing DPN. mTSP Lib contains four min-max mTSP instances (i.e., eil51, berlin52, eil76, and rat99). In testing both DPN and Equity-Transformer, We use both the model trained on 50-node and 100-node instances for validation and report the better result. As shown in Table \ref{mtsplib}, DPN variants also narrow the optimality gap on the mTSP Lib dataset.

\begin{table}[htbp]
\caption{Performances (objective function values and gaps to the BKS) of DPN on mTSP Lib dataset.}
\centering
\renewcommand\arraystretch{1}
\setlength{\tabcolsep}{2.7mm}
\small{
\begin{tabular}{c|c|cc|cccc}
\hline\toprule
\multicolumn{1}{l|}{}         & \multicolumn{1}{l|}{}   & BKS    & HGA    & Equity-Transformer-$\times$8aug & DPN-$\times$8aug    & \multicolumn{2}{c}{DPN-$\times$8aug-$\times$16per} \\ \cline{3-8} 
\multicolumn{1}{l|}{Instance} & \multicolumn{1}{l|}{M=} & Obj.   & Obj.   & Obj.               & Obj.   & Obj.              & Gap               \\ \hline
eil51                         & 2                       & 223    & 223    & 236                & 234    & 227               & 2.00\%            \\
                              & 3                       & 160    & 160    & 164                & 167    & 165               & 3.09\%            \\
                              & 5                       & 118    & 118    & 120                & 120    & 120               & 1.66\%            \\
                              & 7                       & 112    & 112    & 112                & 112    & 112               & 0.00\%            \\ \hline
berlin52                      & 2                       & 4110   & 4110   & 4425               & 4421   & 4186              & 1.86\%            \\
                              & 3                       & 3074   & 3074   & 3263               & 3328   & 3113              & 1.27\%            \\
                              & 5                       & 2441   & 2441   & 2495               & 2663   & 2441              & 0.02\%            \\
                              & 7                       & 2441   & 2441   & 2441               & 2441   & 2441              & 0.00\%            \\ \hline
eil76                         & 2                       & 281    & 282    & 295                & 291    & 291               & 3.65\%            \\
                              & 3                       & 197    & 197    & 206                & 201    & 201               & 2.05\%            \\
                              & 5                       & 143    & 144    & 146                & 144    & 144               & 0.71\%            \\
                              & 7                       & 128    & 128    & 129                & 129    & 128               & 0.30\%            \\ \hline
rat99                         & 2                       & 666    & 668    & 795                & 754    & 754               & 13.16\%           \\
                              & 3                       & 518    & 518    & 566                & 572    & 572               & 10.55\%           \\
                              & 5                       & 450    & 450    & 474                & 467    & 467               & 3.71\%            \\
                              & 7                       & 437    & 437    & 440                & 442    & 442               & 1.02\%            \\ \hline
Gap to BKS                    &                         & 0.00\% & 0.06\% & 4.60\%             & 2.87\% & 2.81\%            &                   \\ \bottomrule\hline
\end{tabular}}
\label{mtsplib}
\end{table}

\subsection{Generalization on Cross-distribution Datasets}\label{distri-appn}

The cross-distribution generalization ability of neural combinatorial optimization solvers is necessary \citep{jiang2022learning}. In Table \ref{50100} and Table \ref{more}, DPN has shown effectiveness in min-max mTSP instances with a uniform distribution. In this subsection, we evaluate the DPN on min-max mTSP200 with the Gaussian distribution, the Rotation distribution, and the Explosion distribution provided in \citep{zhou2023towards}. We use the provided dataset in \citet{zhou2023towards}, and the result is shown in Table \ref{distri}. DPN demonstrates outstanding robustness on different distributions.

\begin{table}[htbp]
\caption{The average objective function values(i.e., Obj.), gaps to the best algorithm (i.e., Gap) on 3 datasets with different distributions. HGA makes one run for these results, and the best result is in bold.}
\centering
\renewcommand\arraystretch{1}
\setlength{\tabcolsep}{3.85mm}
\small{
\begin{tabular}{lcccccc}
\hline\toprule
\multicolumn{7}{c}{Min-max mTSP200-Gaussian}                                                                                                                                       \\ \hline
\multicolumn{1}{c|}{$M$=}                              & \multicolumn{2}{c|}{10}                        & \multicolumn{2}{c|}{15}                       & \multicolumn{2}{c}{20}   \\ \hline
\multicolumn{1}{l|}{Methods}                           & Obj.            & \multicolumn{1}{c|}{Gap.}    & Obj.            & \multicolumn{1}{c|}{Gap.}   & Obj.            & Gap.   \\ \hline
\multicolumn{1}{l|}{HGA}                               & \textbf{1.5063} & \multicolumn{1}{c|}{-}       & 1.5031          & \multicolumn{1}{c|}{0.00\%} & 1.5031          & 0.00\% \\
\multicolumn{1}{l|}{Equity-Transformer-F-$\times$8aug} & 1.6426          & \multicolumn{1}{c|}{9.05\%}  & 1.5458          & \multicolumn{1}{c|}{2.84\%} & 1.5265          & 1.56\% \\
\multicolumn{1}{l|}{DPN-F-$\times$8aug}                & 1.5204          & \multicolumn{1}{c|}{0.93\%}  & 1.5032          & \multicolumn{1}{c|}{0.00\%} & 1.5031          & 0.00\% \\
\multicolumn{1}{l|}{DPN-F-$\times$8aug-$\times$16per}  & 1.5181          & \multicolumn{1}{c|}{0.78\%}  & \textbf{1.5031} & \multicolumn{1}{c|}{-}      & \textbf{1.5031} & -      \\ \hline
\multicolumn{7}{c}{Min-max mTSP200-Explosion}                                                                                                                                      \\ \hline
\multicolumn{1}{c|}{$M$=}                              & \multicolumn{2}{c|}{10}                        & \multicolumn{2}{c|}{15}                       & \multicolumn{2}{c}{20}   \\ \hline
\multicolumn{1}{l|}{Methods}                           & Obj.            & \multicolumn{1}{c|}{Gap.}    & Obj.            & \multicolumn{1}{c|}{Gap.}   & Obj.            & Gap.   \\ \hline
\multicolumn{1}{l|}{HGA}                               & \textbf{1.7651} & \multicolumn{1}{c|}{-}       & \textbf{1.7488} & \multicolumn{1}{c|}{-}      & \textbf{1.7486} & -      \\
\multicolumn{1}{l|}{Equity-Transformer-F-$\times$8aug} & 1.9269          & \multicolumn{1}{c|}{9.17\%}  & 1.8277          & \multicolumn{1}{c|}{4.51\%} & 1.8062          & 3.30\% \\
\multicolumn{1}{l|}{DPN-F-$\times$8aug}                & 1.7924          & \multicolumn{1}{c|}{1.55\%}  & 1.7668          & \multicolumn{1}{c|}{1.03\%} & 1.7663          & 1.01\% \\
\multicolumn{1}{l|}{DPN-F-$\times$8aug-$\times$16per}  & 1.7894          & \multicolumn{1}{c|}{1.37\%}  & 1.7667          & \multicolumn{1}{c|}{1.02\%} & 1.7663          & 1.01\% \\ \hline
\multicolumn{7}{c}{Min-max mTSP200-Rotation}                                                                                                                                       \\ \hline
\multicolumn{1}{c|}{$M$=}                              & \multicolumn{2}{c|}{10}                        & \multicolumn{2}{c|}{15}                       & \multicolumn{2}{c}{20}   \\ \hline
\multicolumn{1}{l|}{Methods}                           & Obj.            & \multicolumn{1}{c|}{Gap.}    & Obj.            & \multicolumn{1}{c|}{Gap.}   & Obj.            & Gap.   \\ \hline
\multicolumn{1}{l|}{HGA}                               & 1.7492          & \multicolumn{1}{c|}{2.79\%}  & 1.7393          & \multicolumn{1}{c|}{4.23\%} & 1.7391          & 4.28\% \\
\multicolumn{1}{l|}{Equity-Transformer-F-$\times$8aug} & 1.8871          & \multicolumn{1}{c|}{10.90\%} & 1.7524          & \multicolumn{1}{c|}{5.01\%} & 1.7151          & 2.84\% \\
\multicolumn{1}{l|}{DPN-F-$\times$8aug}                & 1.7064          & \multicolumn{1}{c|}{0.28\%}  & 1.6697          & \multicolumn{1}{c|}{0.05\%} & 1.6678          & 0.00\% \\
\multicolumn{1}{l|}{DPN-F-$\times$8aug-$\times$16per}  & \textbf{1.7017} & \multicolumn{1}{c|}{-}       & \textbf{1.6688} & \multicolumn{1}{c|}{-}      & \textbf{1.6677} & -      \\ \bottomrule\hline
\end{tabular}}\label{distri}
\end{table}
\newpage

\subsection{Extending to $N$=100,000}
The capability of extending to very large-scale instances \citep{luo2023neural} is also necessary for neural min-max VRP solvers. The original P\&N Encoder of DPN cannot manage large-scale problems such as $N=100,000$. When generating a single solution, the space complexity of DPN is $\mathcal{O}(N(N+M))$, so the proposed DPN can only solve up to 10,000-scale problems on a single Tesla V-100S GPU.

\textbf{DPN-w/o Navigation Part for $N$=100,000.} To handle large-scale data such as $N$=100,000, we further propose a variant of DPN. We design a variant of DPN by removing the navigation part in each P\&N Encoder layer (only preserving the partition part, marked as DPN-w/o Navigation Part in Table \ref{large}) and train this variant from scratch on 100-node min-max mTSP (i.e, $N+D$=100). The space complexity of the DPN-w/o Navigation Part reduces to $\mathcal{O}(NM)$ so this model can derive feasible solutions on 100,000 scale problems. Results in Table below \ref{large} show that this variant can effectively solve very large min-max mTSP($N$=100,000) on a single GPU with considerable time.

All the heuristic baselines (like HGA and LKH) with default settings cannot generate results within 5 days on 100,000-scale problems. However, the performance of solutions can be partially reflected by the comparison to DPN and Equity-Transformer on min-max mTSP10,000 shown in Table \ref{large} where this variant of DPN demonstrates better scaling performances.

\begin{table}[H]
\centering
\caption{The average objective function values(i.e., Obj.) and the total time (i.e., Time) of DPN and DPN-w/o navigation part on min-max mTSP 10,000 and very large-scale instances.}
\renewcommand\arraystretch{1}
\setlength{\tabcolsep}{5mm}
\small{
\begin{tabular}{l|cccccc}
\bottomrule
                            & \multicolumn{6}{c}{min-max mTSP10,000($N$=9,999,$D$=1,10 instances)}                                        \\\hline\bottomrule
\multicolumn{1}{c|}{$M$=}   & \multicolumn{2}{c}{500}           & \multicolumn{2}{c}{750}           & \multicolumn{2}{c}{1,000}          \\\hline
\multicolumn{1}{c|}{Method} & Cost   & \multicolumn{1}{c}{Time} & Cost   & \multicolumn{1}{c}{Time} & Cost   & \multicolumn{1}{c}{Time} \\ \hline
Equity-Transformer          & 4.5645 & 3m                       & 2.9245 & 3m                       & 2.8524 & 3m                       \\
DPN                         & 2.4640 & 2.8m                     & 2.4724 & 2.8m                     & 2.3853 & 2.8m                     \\
DPN-w/o Navigation Part     & 2.3487 & 2.8m                     & 2.3333 & 2.8m                     & 2.2568 & 2.8m                     \\ \bottomrule\bottomrule
                            & \multicolumn{6}{c}{min-max mTSP(10 instances)}                                                            \\ \hline
\multicolumn{1}{c|}{}       & \multicolumn{2}{c}{$N+D$=50,000}     & \multicolumn{2}{c}{$N+D$=70,000}     & \multicolumn{2}{c}{$N+D$=100,000}    \\\hline
\multicolumn{1}{c|}{$M$=}   & \multicolumn{2}{c}{500}           & \multicolumn{2}{c}{500}           & \multicolumn{2}{c}{500}           \\ \hline
\multicolumn{1}{c|}{Method} & Cost   & \multicolumn{1}{c}{Time} & Cost   & \multicolumn{1}{c}{Time} & Cost   & \multicolumn{1}{c}{Time} \\ \hline
DPN-w/o Navigation Part     & 3.4183 & \multicolumn{1}{c}{34m}  & 3.7832 & \multicolumn{1}{c}{72m}  & 4.2078 & \multicolumn{1}{c}{2h}   \\ \toprule\hline
\end{tabular}}\label{large}
\end{table}

\subsection{Ablation on $M$=10}

\begin{figure}[H]
    \centering
    \subfigure[min-max mTSP]{\includegraphics[width = 0.35\textwidth]{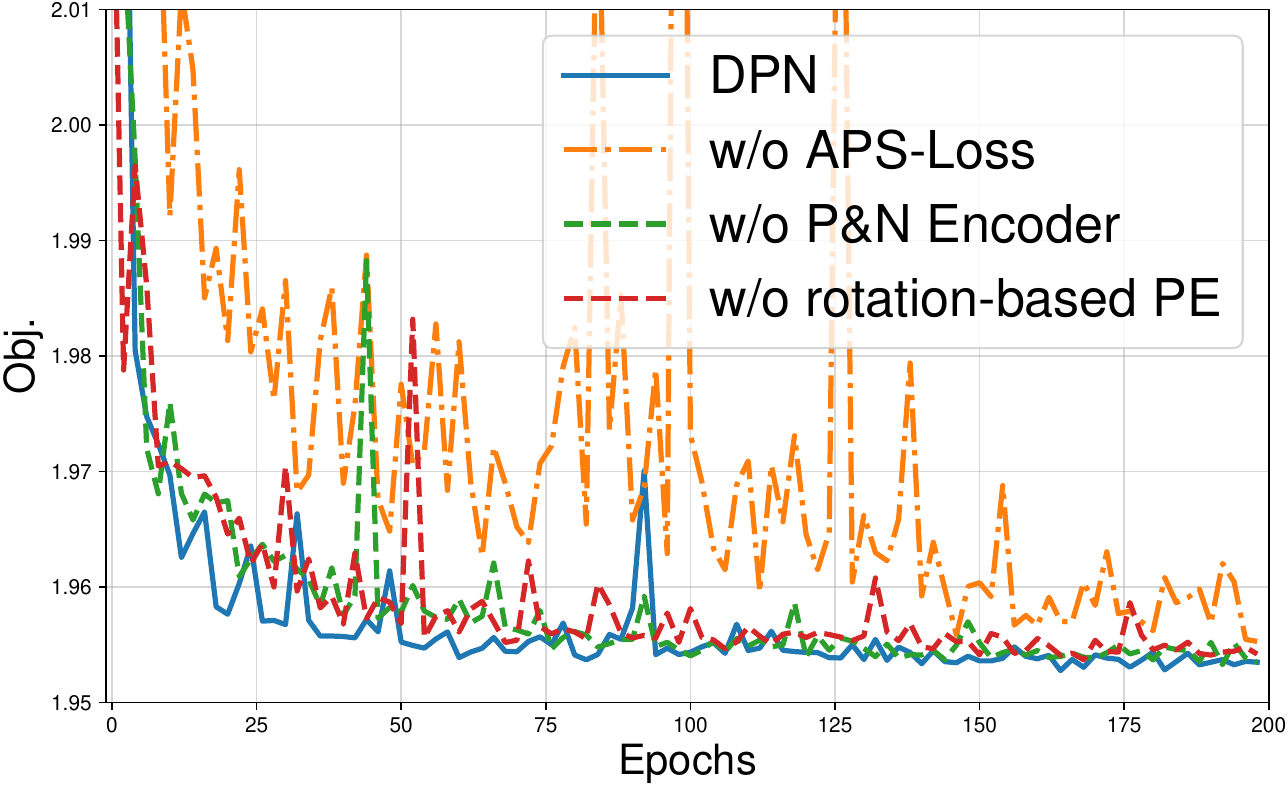}}
    \quad\quad\quad\quad
    \subfigure[min-max mPDP]{\includegraphics[width = 0.35\textwidth]{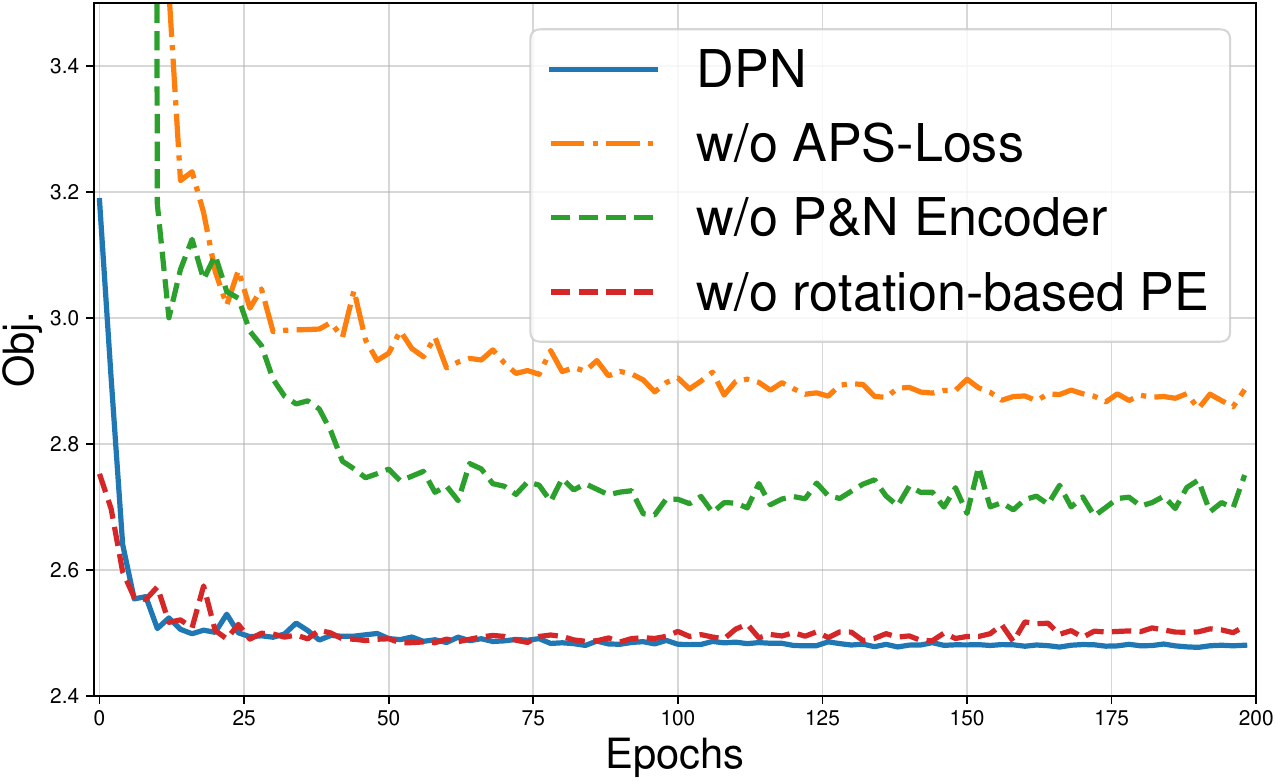}}
    \caption{Training curves for ablation study ($M$=10).}
    \label{fig:curve10}
\end{figure}

The ablation studies in Section \ref{ablation} focus on the setting of $M=5$, and the ablation studies on $M=10$ are provided in this part. As shown in Table \ref{ablationm10}, the efficiency of the proposed components in DPN is even more significant in problems with $M=10$. Due to the decision space of navigation reducing together with the average length of routes, min-max VRPs with $M=10$ agents might focus more on partition tasks rather than navigation tasks, so the use of Rotation-based PE demonstrates more significant superiority. The training curve in Figure \ref{fig:curve10} also confirms the conclusion. The data points in both Figure \ref{fig:curve} and Figure \ref{fig:curve10} are derived from evaluating the same validation set. These ablation models are equipped with both the $\times8 $aug and $\times16$per augmentations.

\begin{table}[htbp]
\caption{Ablation study on proposed components of DPN.}
\centering
\renewcommand\arraystretch{1}
\setlength{\tabcolsep}{3.85mm}
\small{\begin{tabular}{l|cc}
\hline\bottomrule
                               & min-max mTSP100 & min-max mPDP100 \\ \cline{2-3} 
\multicolumn{1}{c|}{$M$=}        & 10              & 10              \\ \hline
w/o P\&N Encoder               & 1.9537          & 3.2706          \\
w/o APS-Loss                   & 1.9571          & 2.8078          \\
w/o Rotation-based PE          & 1.9544          & 2.7123          \\
DPN-$\times8$aug-$\times16$per & \textbf{1.9532} & \textbf{2.6949} \\ \toprule\hline
\end{tabular}}\label{ablationm10}
\end{table}

\subsection{Ablation on PE}\label{ablationpe}
In Appendix \ref{PEmotivation}, we have shown that the Rotation-based PE implemented in DPN helps improve convergence and optimality on both $M$=5 and $M$=10. To further underscore the advantages of using Rotation-based PE, this section presents additional comparative experiments that substantiate the efficacy of Rotation-based PE in facilitating the learning of partition strategies, especially when considering various depot locations (outlined in Section \ref{PEmotivation}). These strategies are crucial for adapting to different spatial distributions of depots, thereby underlining the practical value of Rotation-based PE in adaptive partitioning contexts.

In this subsection, we generate 10,000 mTSP100 instances, the depot coordinate of the $i$-th instance is as follows: 
\begin{align}
    \textbf{x}_d = \big(\ 0.005+ 0.01(\lfloor \frac{i-1}{100}\rfloor)\ , \ 0.005+0.01(i\mod 100)\ \big),
\end{align}
where $(i\mod 100)$ represents the remainder.

\begin{figure}[H]
    \centering
    \includegraphics[width = 0.9\textwidth]{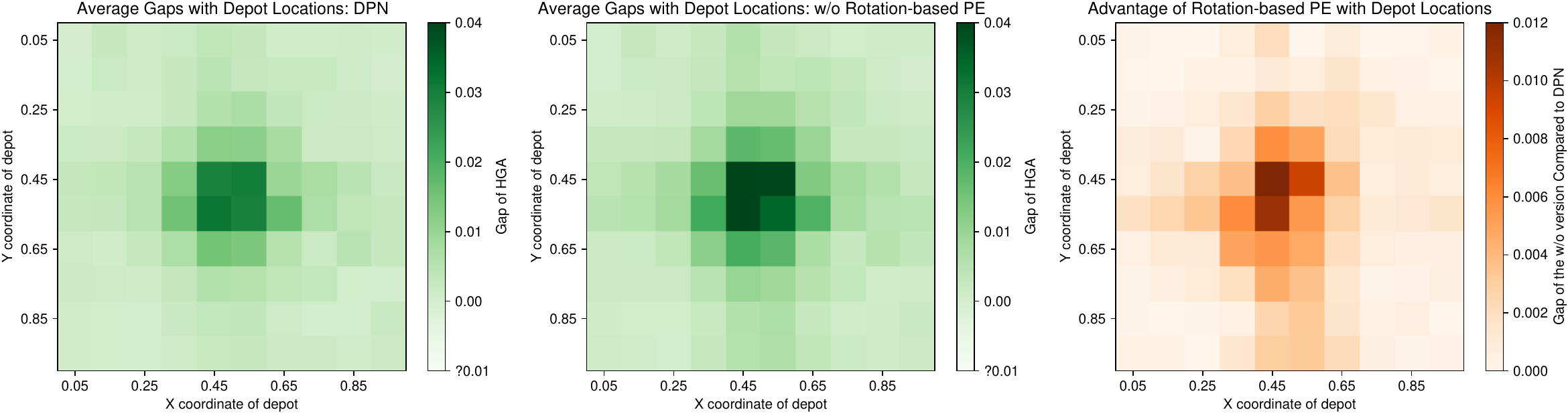}
    \caption{Plot of relations between depot locations and performances. These plots divide the xy-plate into 100 uniform blocks, and each block represents the average Gap of the instances whose depots are located in the block. The two green heatmaps on the left represent the relation between depot locations and the gaps to HGA \citep{mahmoudinazlou2024hybrid} when testing our DPN and the w/o Rotation-based PE ablation model respectively. The red heatmap on the right is the ratio of the objective function between the two versions of models (i.e., Gap of w/o Rotation-based PE to DPN).}
    \label{fig:pe}
\end{figure}

We run the model of DPN-$\times8$aug-$\times16$per, DPN-w/o-Rotation-based-$\times8$aug-$\times16$per, and the heuristic method HGA on these instances with $M$=10. HGA makes one run in these instances. Figure \ref{fig:pe} consists of 100 blocks and each block contains the average gap of the 100 instances whose depot is located in the block. The first two plots on the left in Figure \ref{fig:pe} demonstrate that for both the DPN (left 1) and the ablation model of DPN with SPE (without Rotation-based PE) (left 2), the gap to HGA relates to depot locations and the instances with middle-location depots perform relatively worse. Relatively, Rotation-based PE enables DPN to handle instances at any depot location with an even gap level and demonstrates the contribution of Rotation-based PE to learning a more robust customer partition strategy. 

The right plot shows that compared to Rotation-based PE, the original sinusoidal PE performs well in the instances with corner-location depots, but cannot generalize to the instances with middle-location depots. With the addition of the depot coordinate information, the Rotation-based PE improves the performance of these instances with middle-location depots.

\subsection{Ablation on the Fine-tuning Phase \& Zero-shot Generalization}

\begin{figure}[H]
    \centering
    \subfigure[min-max mTSP200 ($N$=199,$M$=10)]{\includegraphics[width = 0.35\textwidth]{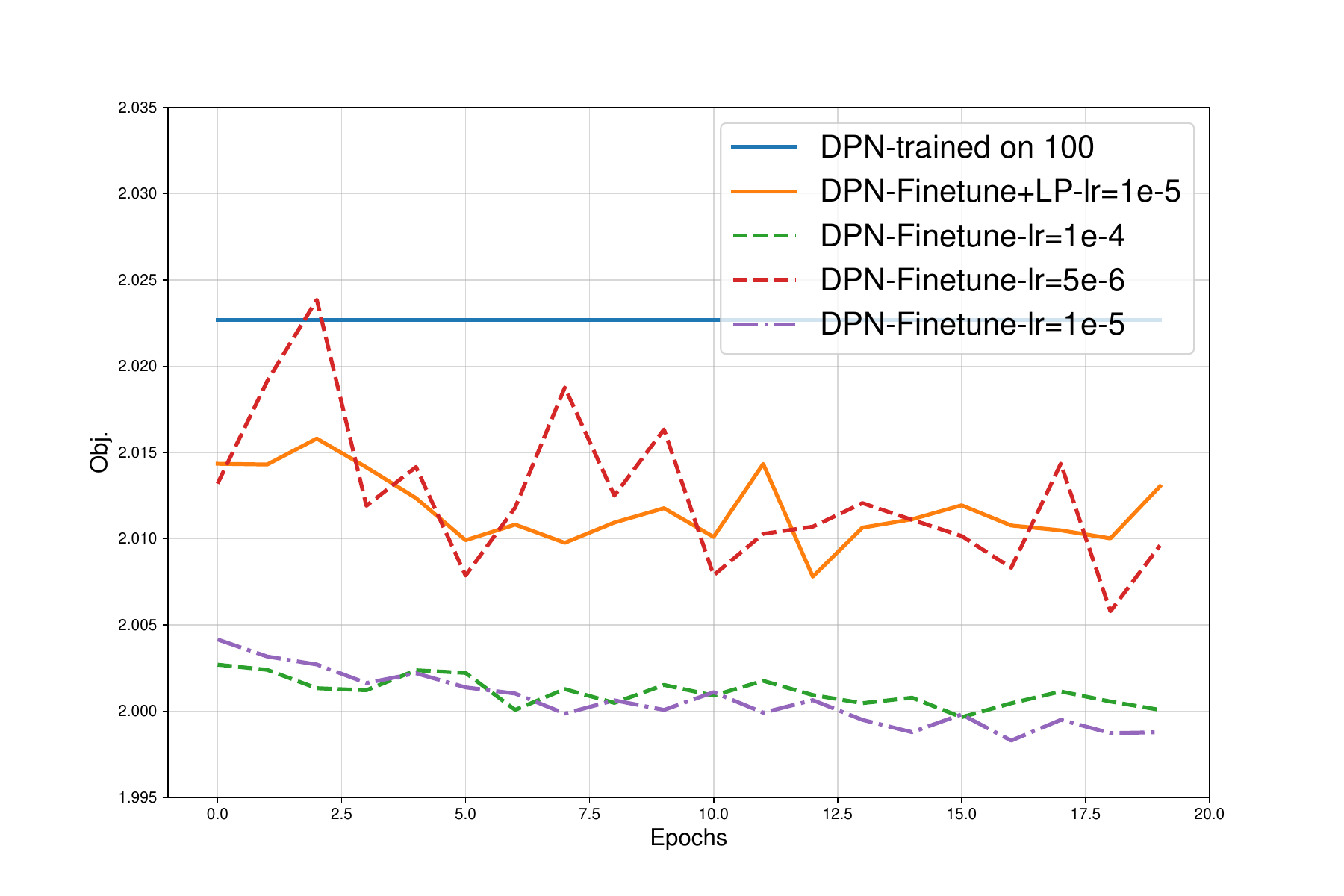}}
    \quad\quad\quad\quad
    \subfigure[min-max mTSP200 ($N$=199,$M$=15)]{\includegraphics[width = 0.35\textwidth]{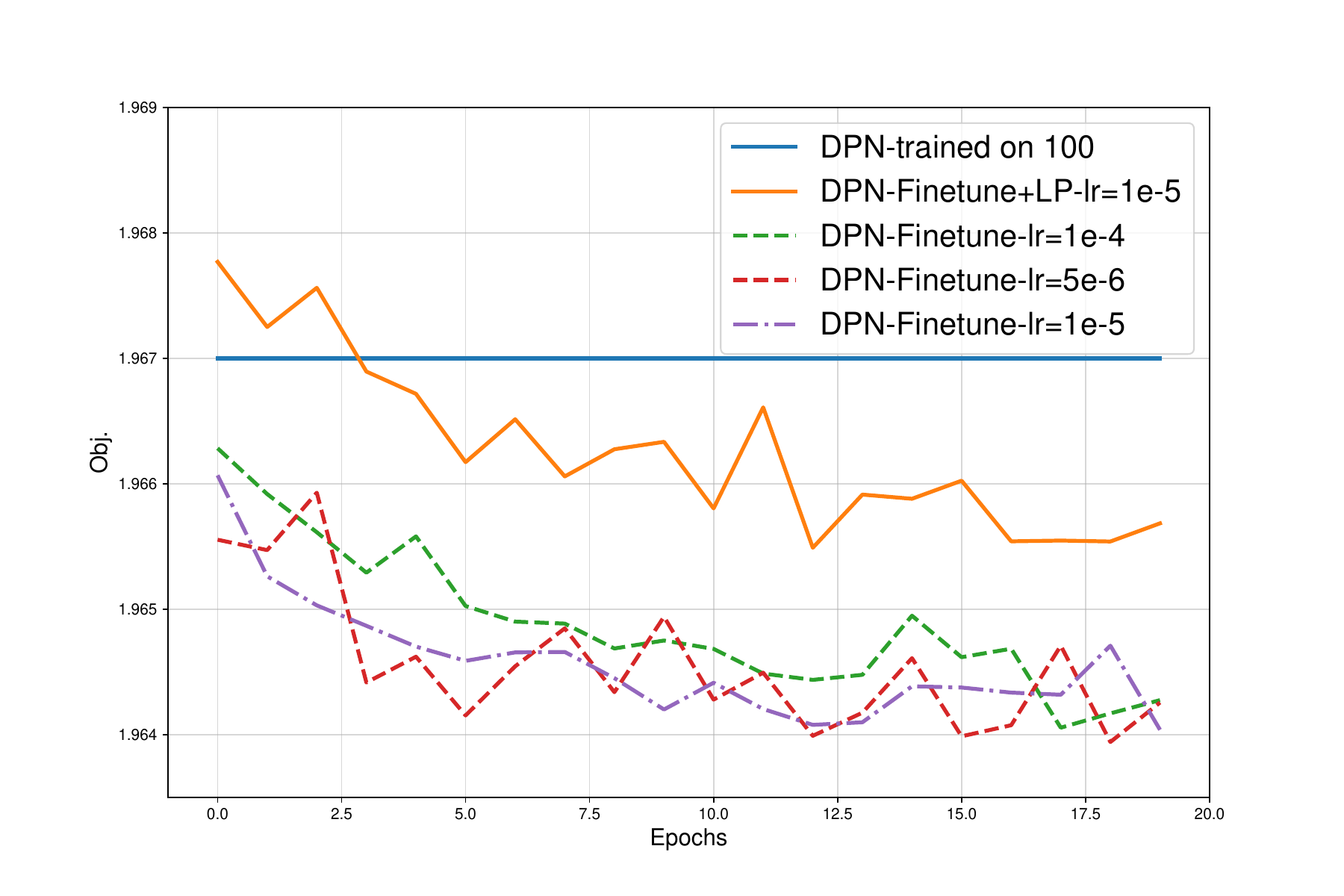}}
    \caption{Training curves on min-max mTSP200 with different learning rates and settings.}
    \label{fig:finetune}
\end{figure}

DPN uses a learning rate of 1e-5 in fine-tuning, and Figure \ref{fig:finetune} provides ablation studies to demonstrate the superiority of this setting. The DPN-Finetune-lr=1e-5 version represents fine-tuning with the learning rate being 1e-5 (i.e., the model for min-max mTSP in Table \ref{more}) and the DPN-Finetune-LP-lr=1e-5 is the fine-tuning setting adopted in Equity-Transformer \citep{son2023solving} which adds a 2-layer MLP for the contextual embedding $\boldsymbol{q}_{contex}$. Compared to models with other learning rates, the DPN-Finetune-lr=1e-5 consistently shows the best convergence speed among all versions.

Table \ref{zs} tests the ablation fine-tuned models with different settings. It proves the above conclusions as well, and in addition, columns of ``DPN-100'' with $\times8$aug or $\times16$per list the results of DPN trained on 100-scale min-max mTSP and tested on 200-scale min-max mTSP datasets which can demonstrate that our DPN exhibits advantages in zero-shot generalization compared to Equity-Transformer.

\begin{table}[H]
\caption{Ablation study on proposed components in DPN.}
\centering
\renewcommand\arraystretch{1.05}
\setlength{\tabcolsep}{8mm}
\small{\begin{tabular}{lccc}\hline\bottomrule
\multicolumn{4}{c}{Min-max mTSP200}                                                                                                                                  \\ \hline
\multicolumn{1}{c|}{$M$=}                                           & 10                             & 15                             & 20                             \\\hline
\multicolumn{1}{c|}{Methods}                                      & Obj.                           & Obj.                           & Obj.                           \\ \hline
\multicolumn{1}{l|}{HGA}                                          & 1.9861                         & \textbf{1.9628}                & \textbf{1.9627}                \\
\multicolumn{1}{l|}{LKH3}                                         & \textbf{1.9817}                & 1.9628                         & 1.9628                         \\
\multicolumn{1}{l|}{OR-Tools(600s)}                               & 2.3711                         & 2.3687                         & 2.3687                         \\ \hline
\multicolumn{1}{l|}{DAN}                                          & 2.3586                         & 2.1732                         & 2.1151                         \\
\multicolumn{1}{l|}{ScheduleNet*}                                 & 2.35                           & 2.13                           & 2.07                           \\
\multicolumn{1}{l|}{NCE*}                                         & 2.07                           & 1.97                           & 1.96                           \\
\multicolumn{1}{l|}{Equity-Transformer-$\times$8aug}            & 2.0750                         & 1.9947                         & 1.9658                         \\
\multicolumn{1}{l|}{Equity-Transformer-F-$\times$8aug}            & 2.0500                         & 1.9688                         & 1.9631                         \\ \hline
\multicolumn{1}{l|}{DPN-100-$\times$8aug}           & 2.0381                         & 1.9692                         & 1.9660                         \\
\multicolumn{1}{l|}{DPN-100-$\times$8aug-$\times16$per}           & 2.0227                         & 1.9670                         & 1.9648                         \\
\multicolumn{1}{l|}{DPN-F(lr=1e-4)-$\times$8aug-$\times16$per}    & 2.0107                         & 1.9643                         & \cellcolor[HTML]{D9D9D9}1.9628 \\
\multicolumn{1}{l|}{DPN-F(lr=5e-6)-$\times$8aug-$\times16$per}    & 2.0004                         & 1.9644                         & \cellcolor[HTML]{D9D9D9}1.9628 \\
\multicolumn{1}{l|}{DPN-F(lr=1e-5+LP)-$\times$8aug-$\times16$per} & 2.1064                         & 1.9673                         & 1.9645                         \\
\multicolumn{1}{l|}{DPN-F(lr=1e-5)-$\times$8aug-$\times16$per}    & \cellcolor[HTML]{D9D9D9}1.9993 & \cellcolor[HTML]{D9D9D9}1.9640 & \cellcolor[HTML]{D9D9D9}1.9628 \\ \toprule\hline
\end{tabular}}\label{zs}
\end{table}

\newpage
\section{Visualization}
\begin{figure}[H]
    \centering
    \subfigure[LKH3]{\includegraphics[width = 0.27\textwidth]{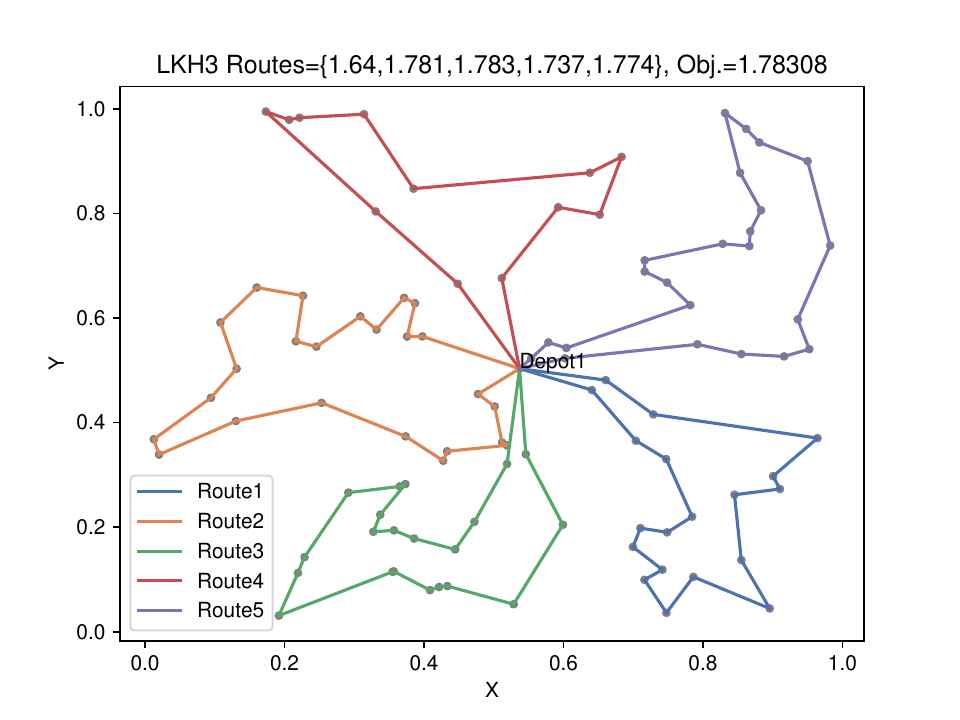}\label{lkh1}}    \subfigure[Equity-Transformer-$\times8$aug]{\includegraphics[width = 0.28\textwidth]{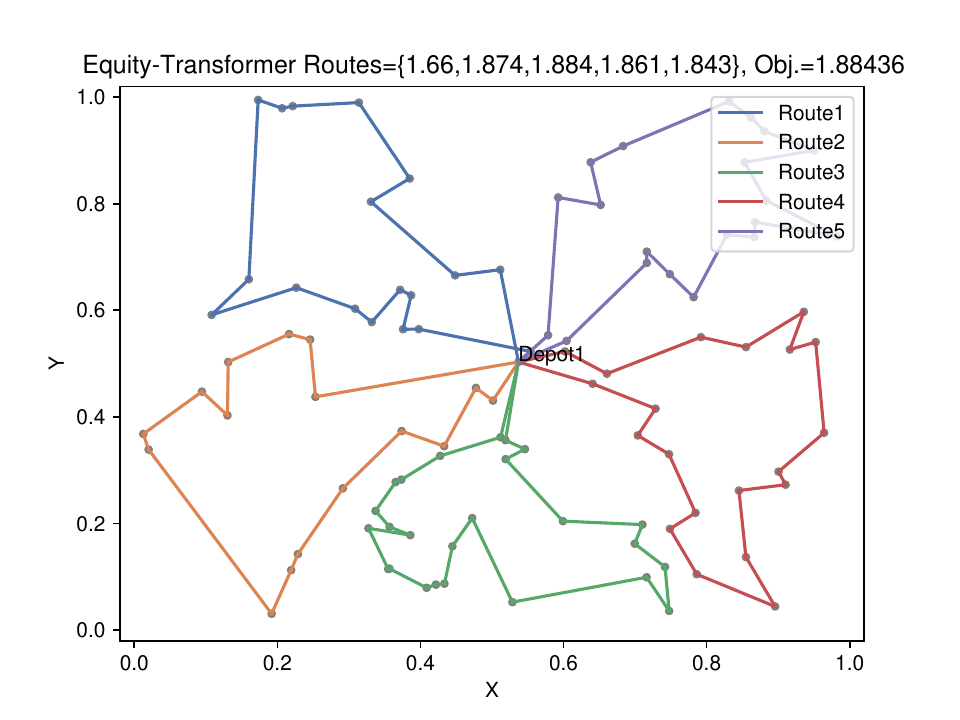}}    \subfigure[DPN-$\times8$aug]{\includegraphics[width = 0.27\textwidth]{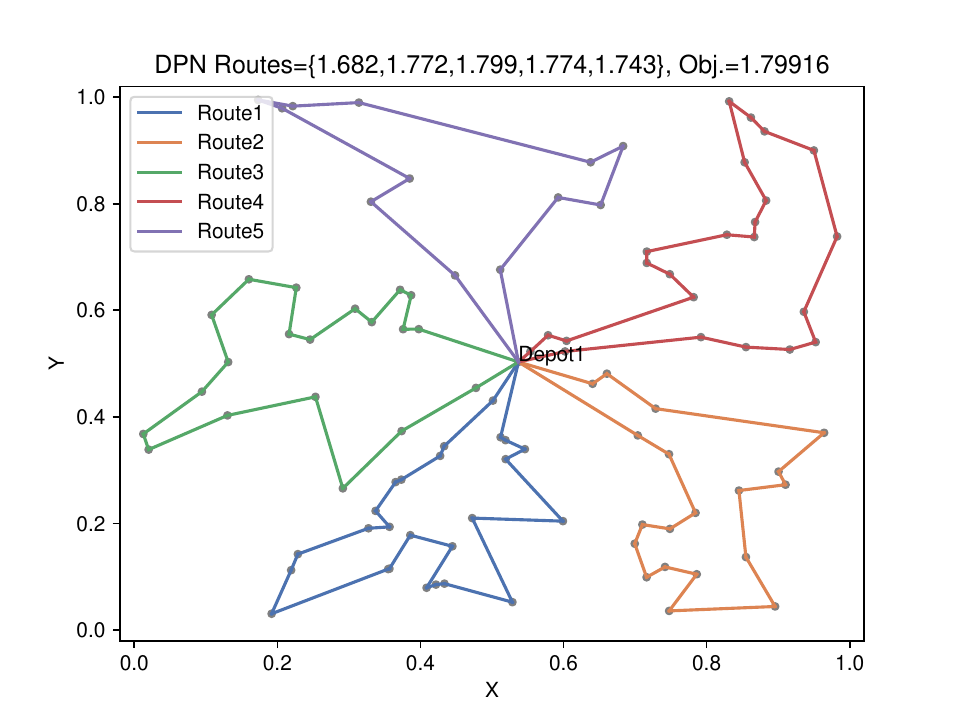}}    \subfigure[LKH3]{\includegraphics[width = 0.27\textwidth]{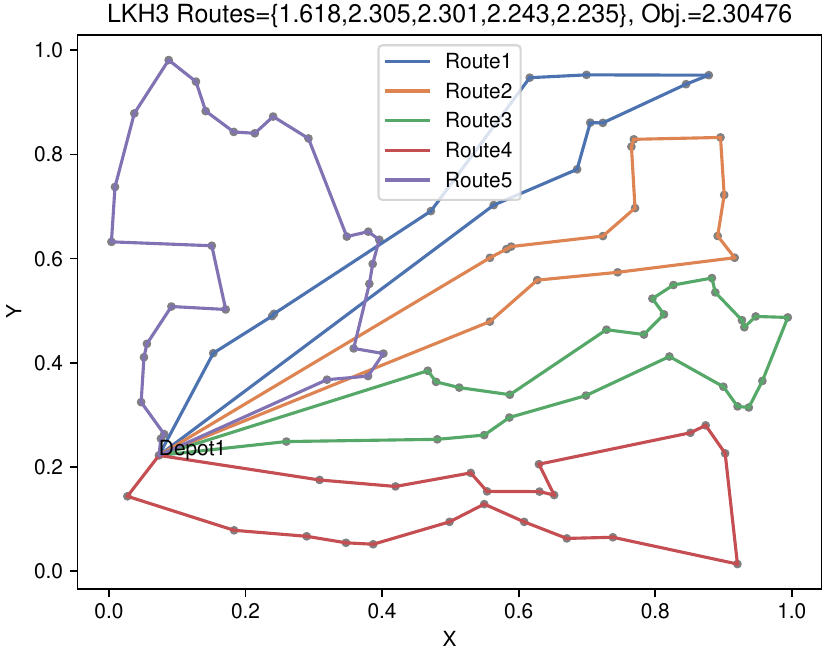}\label{lkh2}}    \subfigure[Equity-Transformer-$\times8$aug]{\includegraphics[width = 0.28\textwidth]{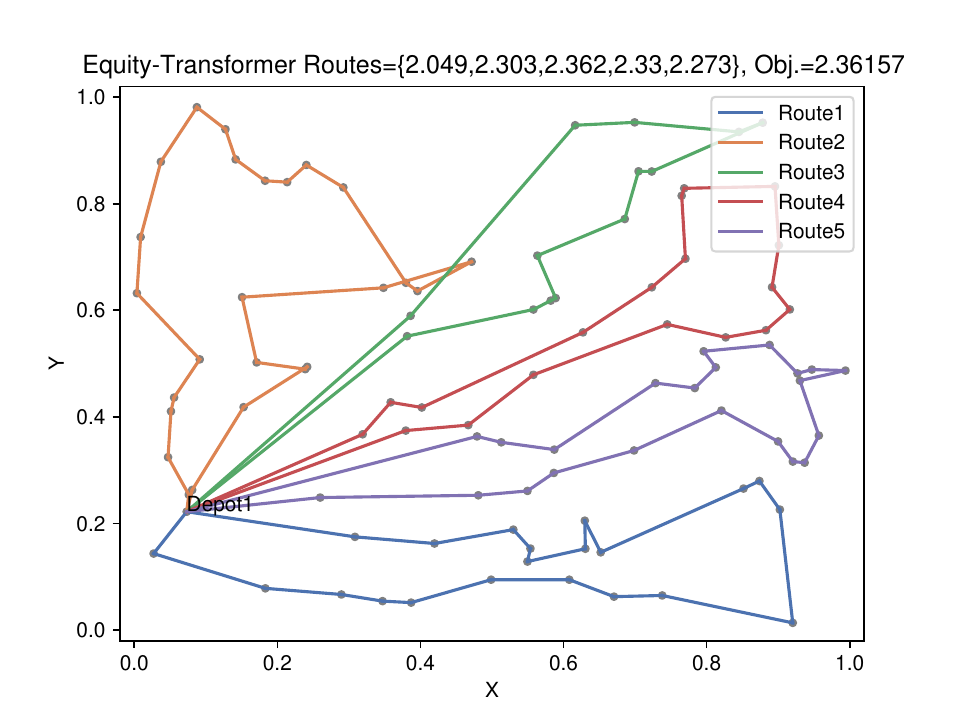}}    \subfigure[DPN-$\times8$aug]{\includegraphics[width = 0.27\textwidth]{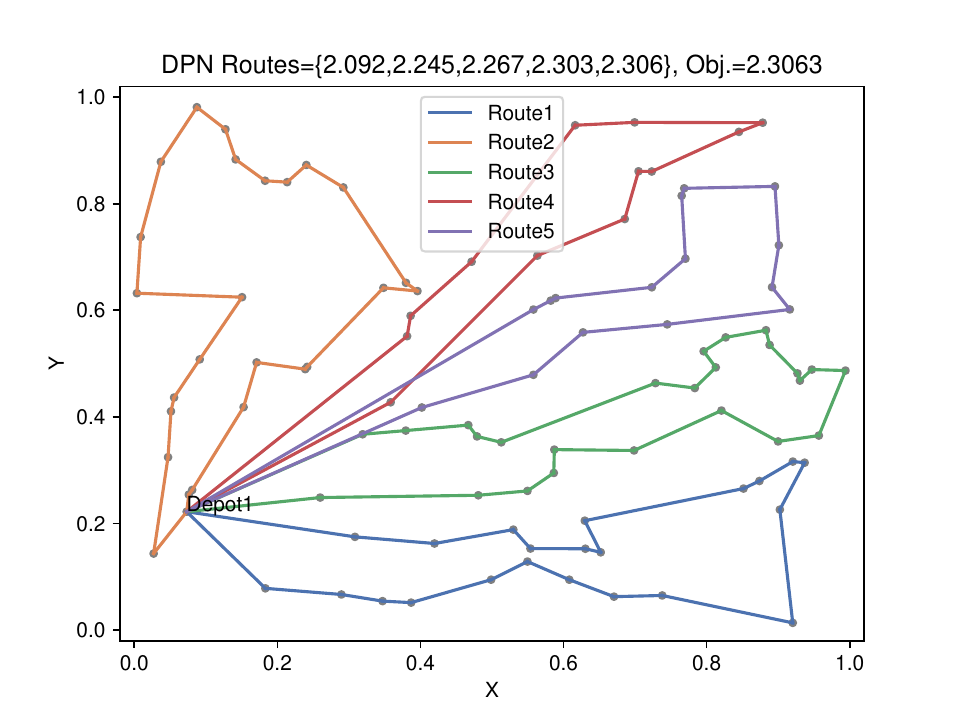}}
    \caption{Min-max mTSP instances ($M$=5), solving by LKH3, Equity-Transformer, and ours DPN.}
    \label{fig:mtsp-vis}
\end{figure}

\begin{figure}[H]
    \centering
    \subfigure[LKH3]{\includegraphics[width = 0.27\textwidth]{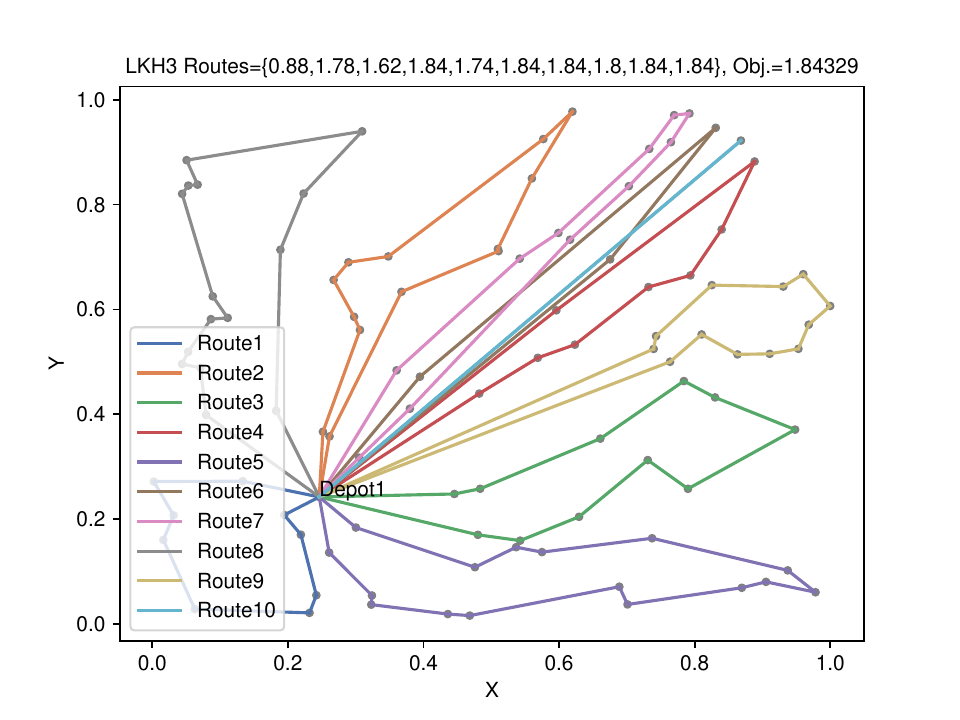}}    \subfigure[Equity-Transformer-$\times8$aug]{\includegraphics[width = 0.29\textwidth]{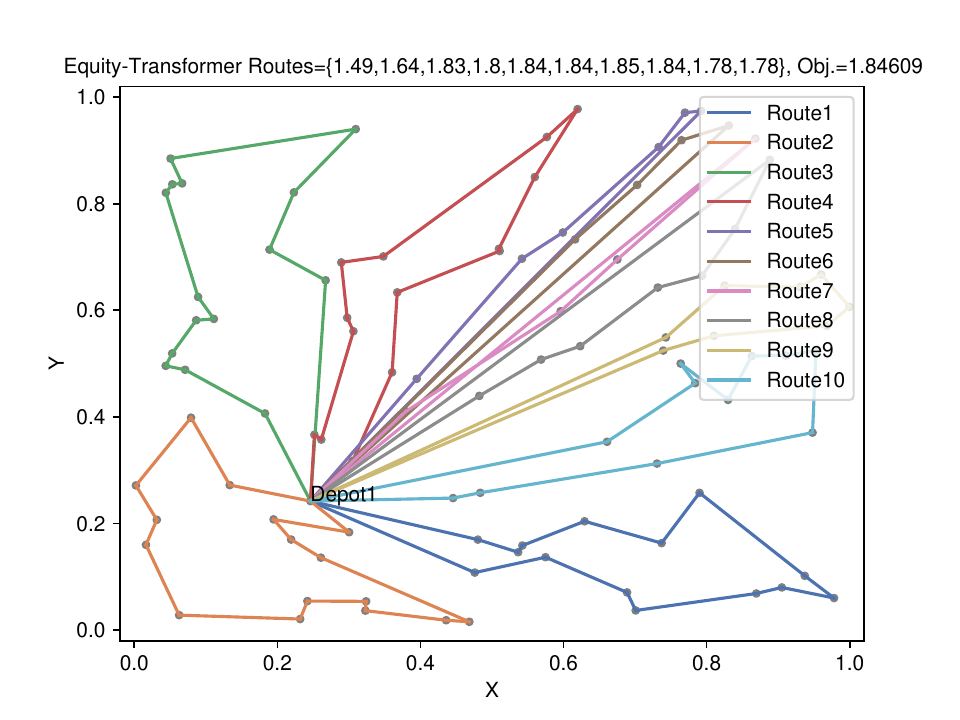}}    \subfigure[DPN-$\times8$aug]{\includegraphics[width = 0.27\textwidth]{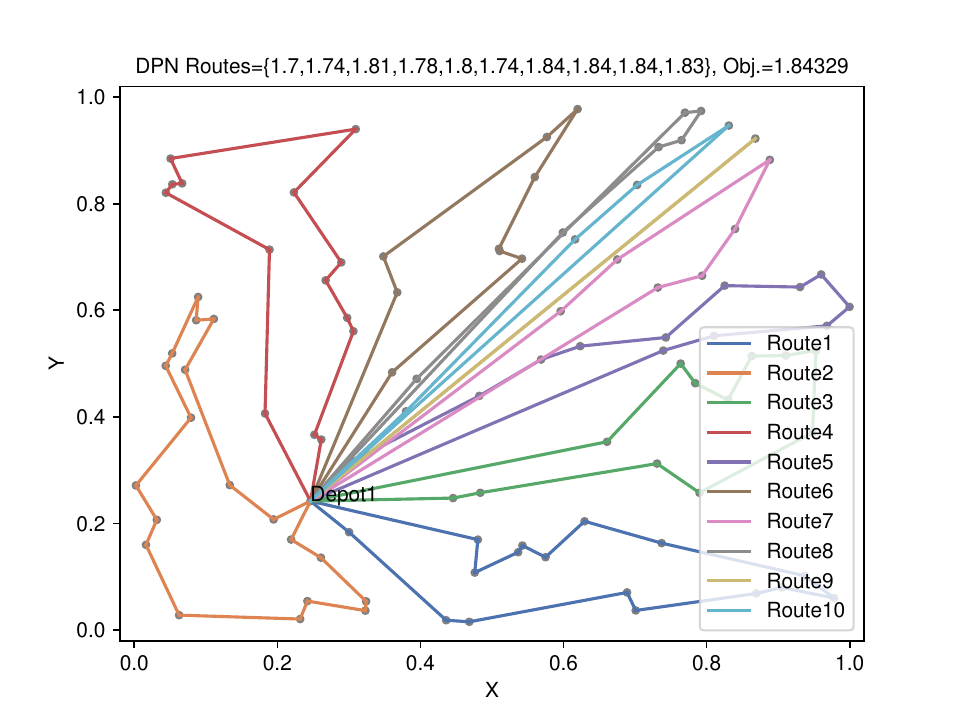}}    \subfigure[LKH3]{\includegraphics[width = 0.27\textwidth]{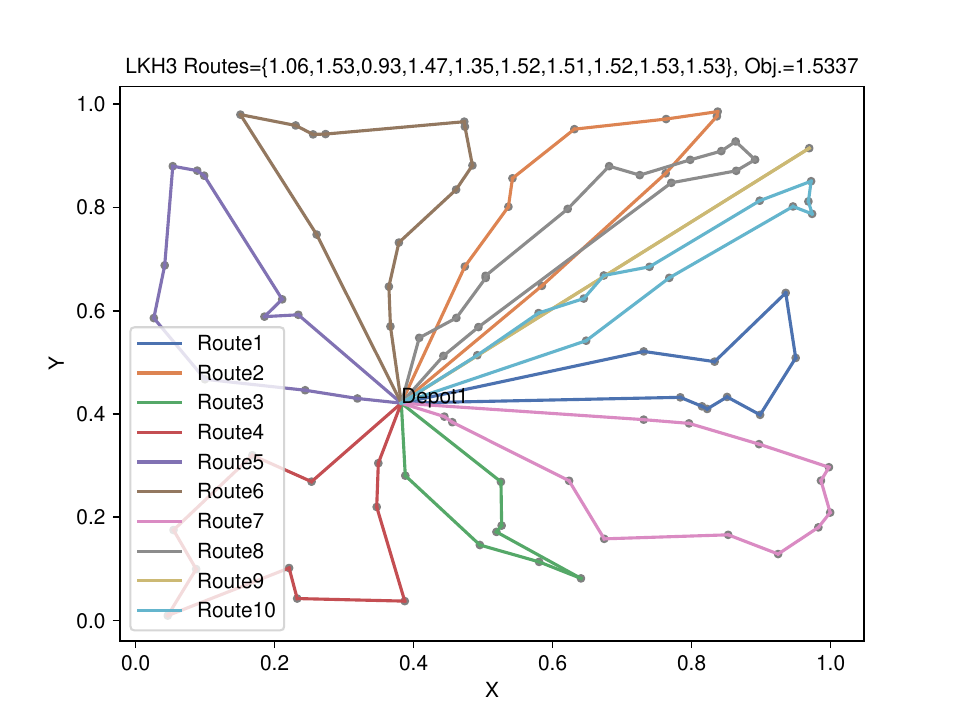}}    \subfigure[Equity-Transformer-$\times8$aug]{\includegraphics[width = 0.285\textwidth]{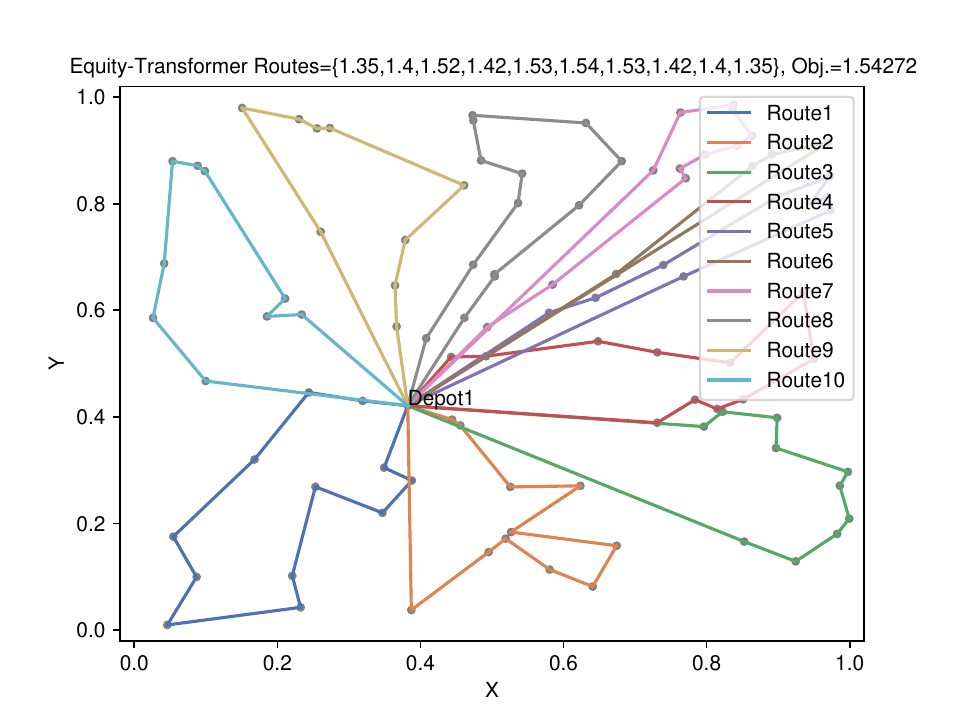}}    \subfigure[DPN-$\times8$aug]{\includegraphics[width = 0.27\textwidth]{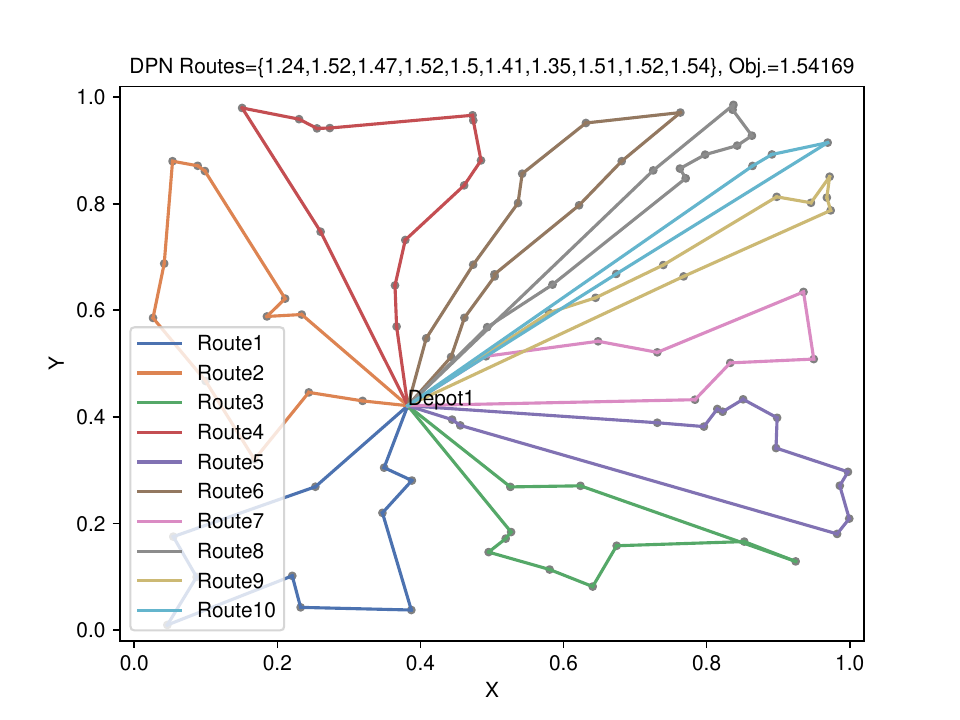}}
    \caption{Min-max mTSP instances ($M$=10), solving by LKH3, Equity-Transformer, and ours DPN.}
    \label{fig:mtsp-vis2}
\end{figure}

\begin{figure}[H]
    \centering\subfigure[Equity-Transformer-$\times8$aug]{\includegraphics[width = 0.4\textwidth]{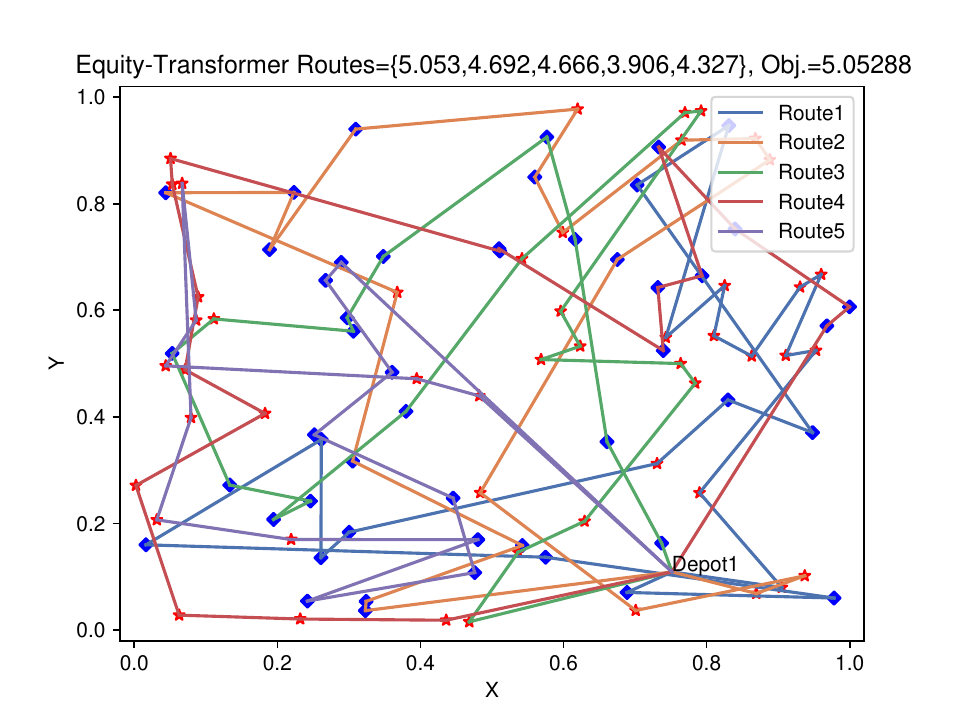}}   \quad\quad\quad \subfigure[DPN-$\times8$aug]{\includegraphics[width = 0.38\textwidth]{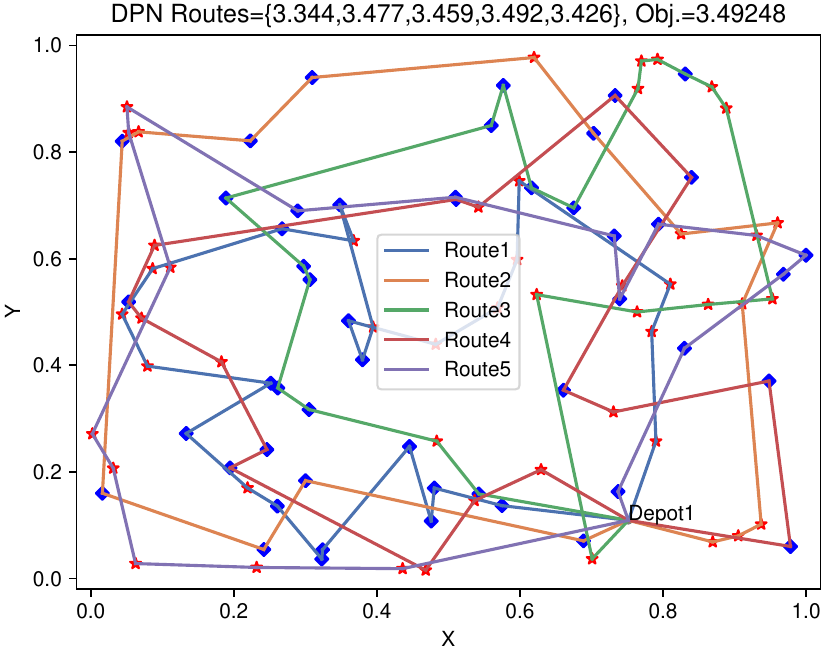}}    
    \caption{Min-max mPDP instances ($M$=5), solving by Equity-Transformer and ours DPN.  Different colors and shapes (blue diamonds for pickups and red stars for deliveries) distinguish customers.}
    \label{fig:mpdp-vis}
\end{figure}

\begin{figure}[H]
    \centering \subfigure[Equity-Transformer-$\times8$aug]{\includegraphics[width = 0.4\textwidth]{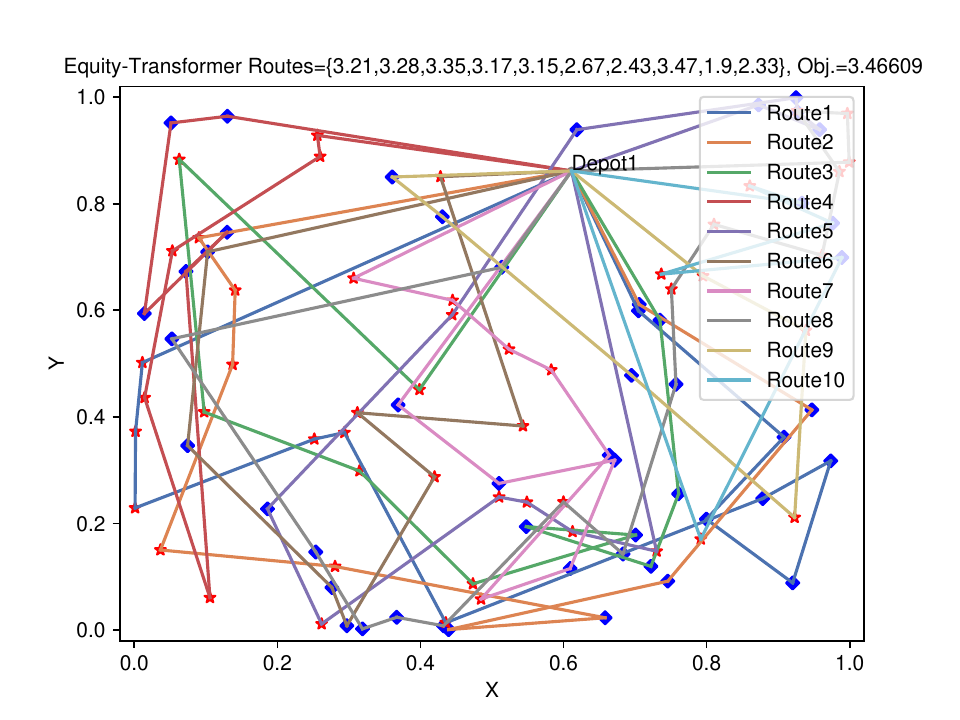}}  \quad\quad\quad   \subfigure[DPN-$\times8$aug]{\includegraphics[width = 0.38\textwidth]{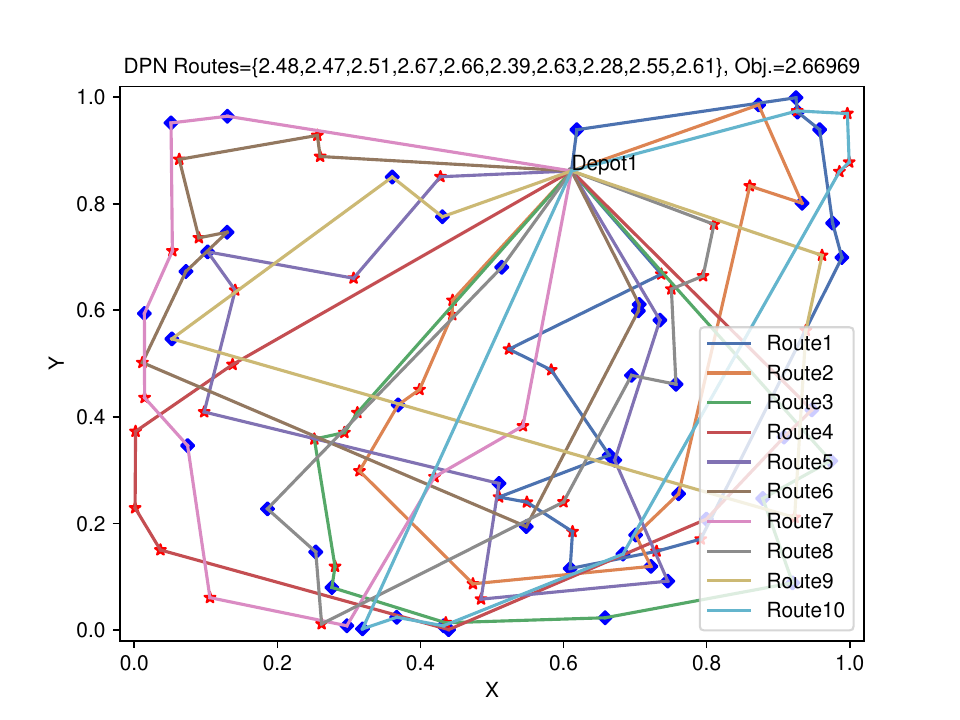}}
    \caption{Min-max mPDP instances ($M$=10), solving by Equity-Transformer and ours DPN. Different colors and shapes (blue diamonds for pickups and red stars for deliveries) distinguish customers.}
    \label{fig:mpdp-vis2}
\end{figure}

\begin{figure}[H]
    \centering \subfigure[DPN-$\times8$aug-$\times16$per]{\includegraphics[width = 0.38\textwidth]{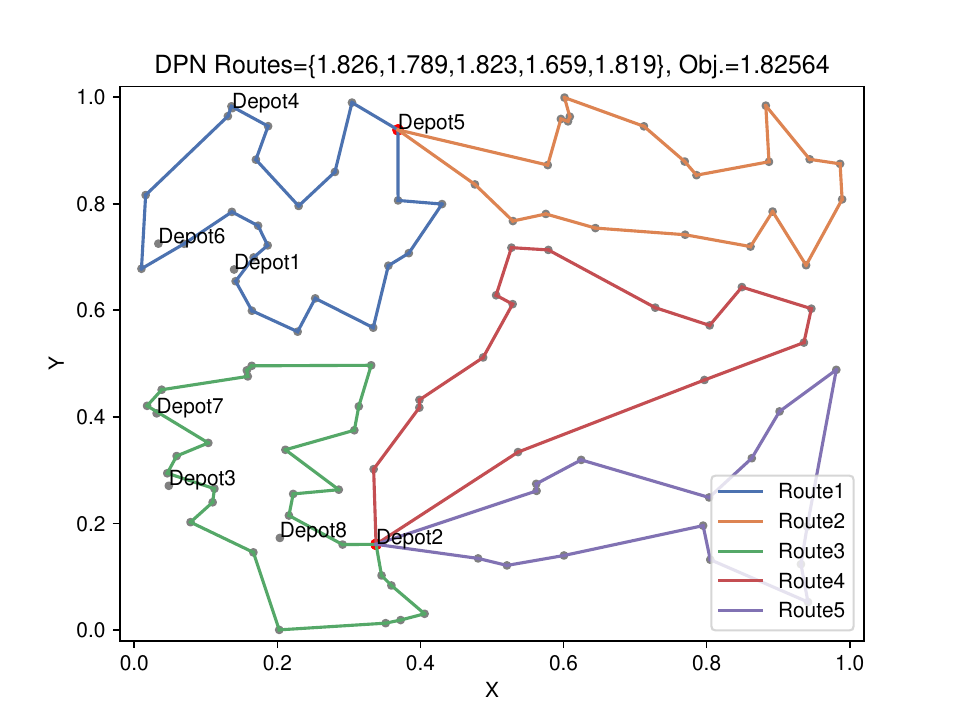}}  \quad\quad\quad   \subfigure[DPN-$\times8$aug-$\times16$per]{\includegraphics[width = 0.38\textwidth]{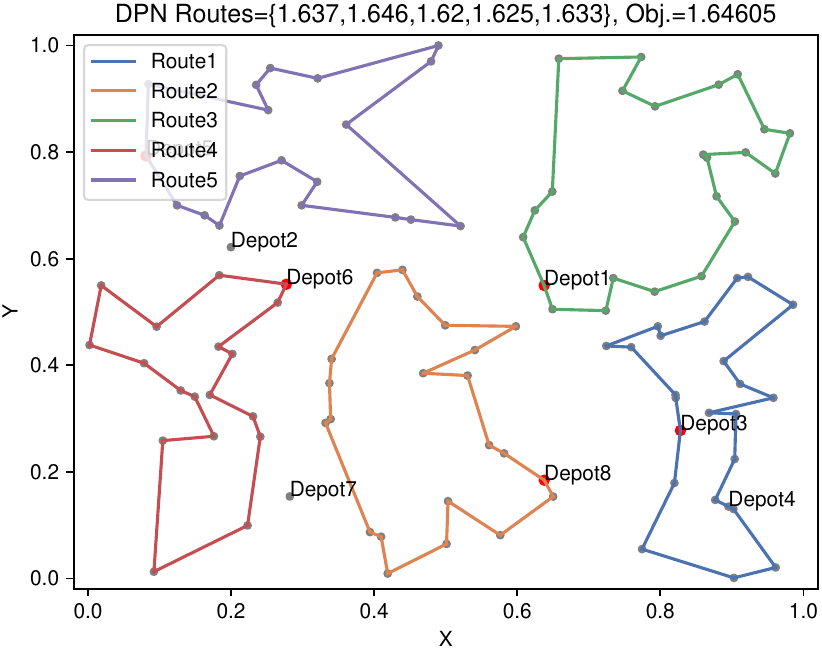}}
    \caption{Min-max MDVRP instances ($D$=8,$M$=5) solving by ours DPN. Selected depots are highlighted by red.}
    \label{fig:mdvrp-vis}
\end{figure}

\begin{figure}[H]
    \centering \subfigure[DPN-$\times8$aug-$\times16$per]{\includegraphics[width = 0.38\textwidth]{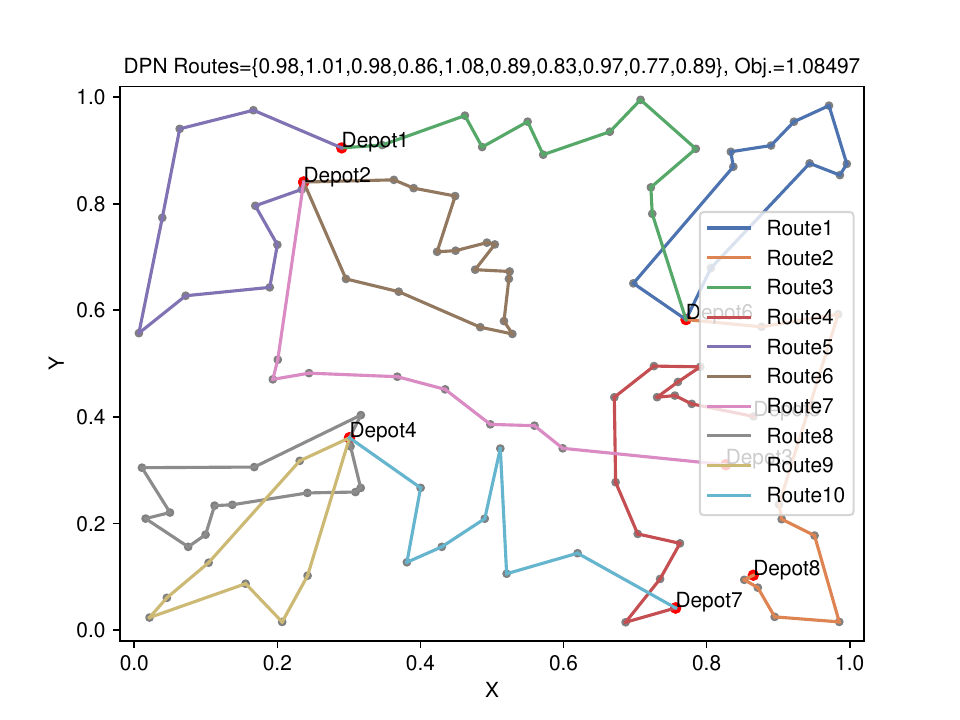}}  \quad\quad\quad   \subfigure[DPN-$\times8$aug-$\times16$per]{\includegraphics[width = 0.38\textwidth]{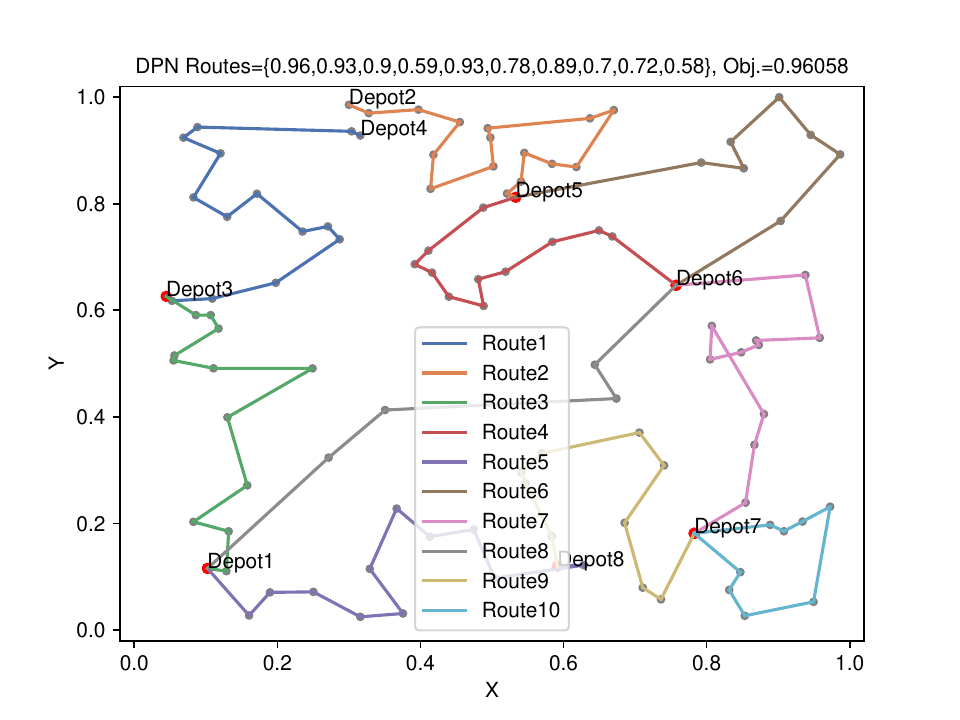}}
    \caption{Min-max FMDVRP instances ($D$=8,$M$=10) solving by ours DPN. Selected depots are highlighted by red.}
    \label{fig:mdvrp-vis2}
\end{figure}

\begin{figure}[H]
    \centering \subfigure[DPN-$\times8$aug-$\times16$per]{\includegraphics[width = 0.38\textwidth]{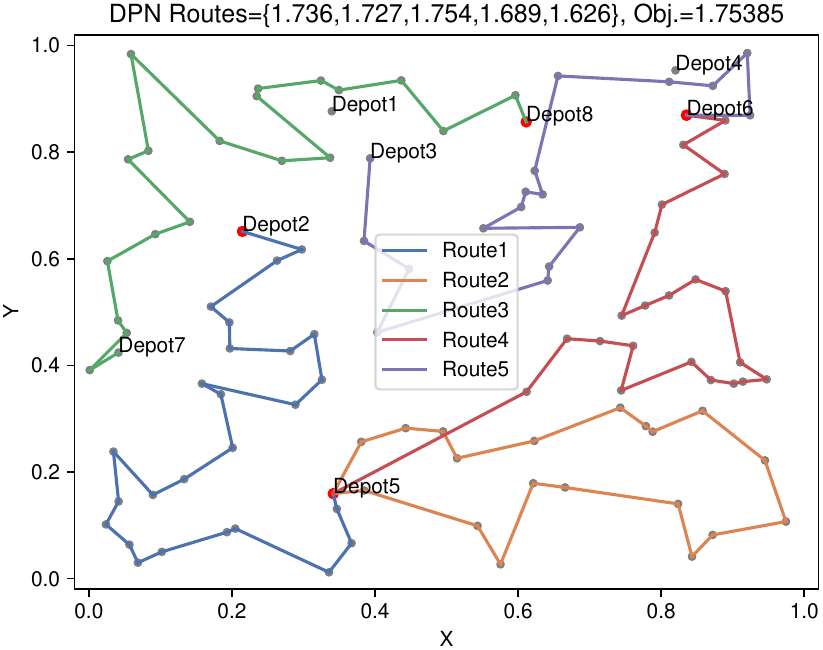}}  \quad\quad\quad   \subfigure[DPN-$\times8$aug-$\times16$per]{\includegraphics[width = 0.38\textwidth]{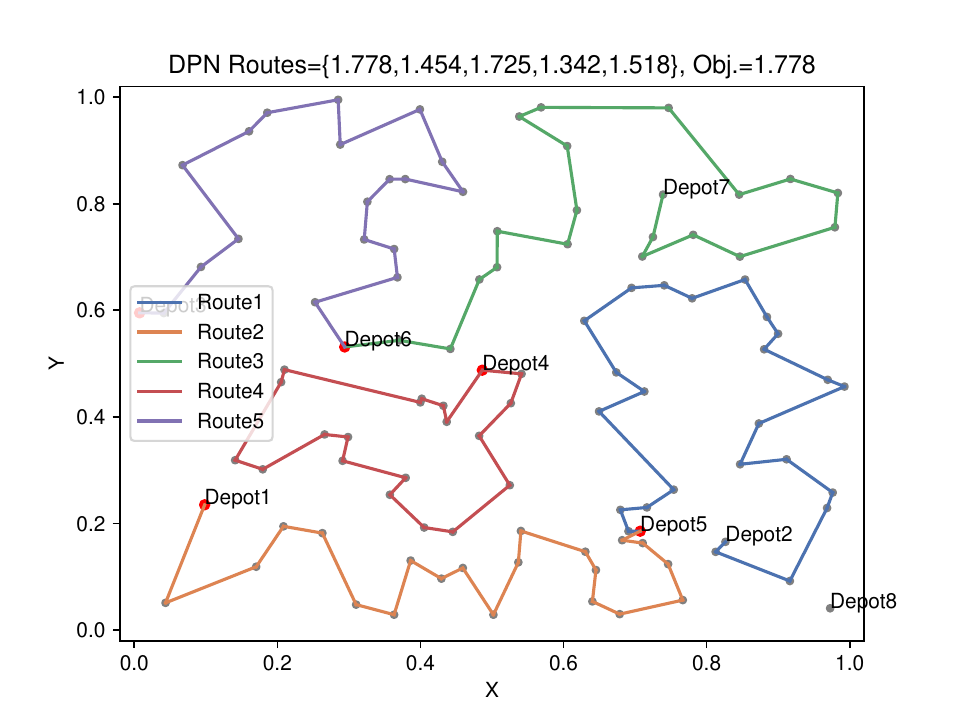}}
    \caption{Min-max FMDVRP instances ($D$=8,$M$=5) solving by ours DPN. Selected depots are highlighted by red.}
    \label{fig:fmdvrp-vis}
\end{figure}

\begin{figure}[H]
    \centering \subfigure[DPN-$\times8$aug-$\times16$per]{\includegraphics[width = 0.38\textwidth]{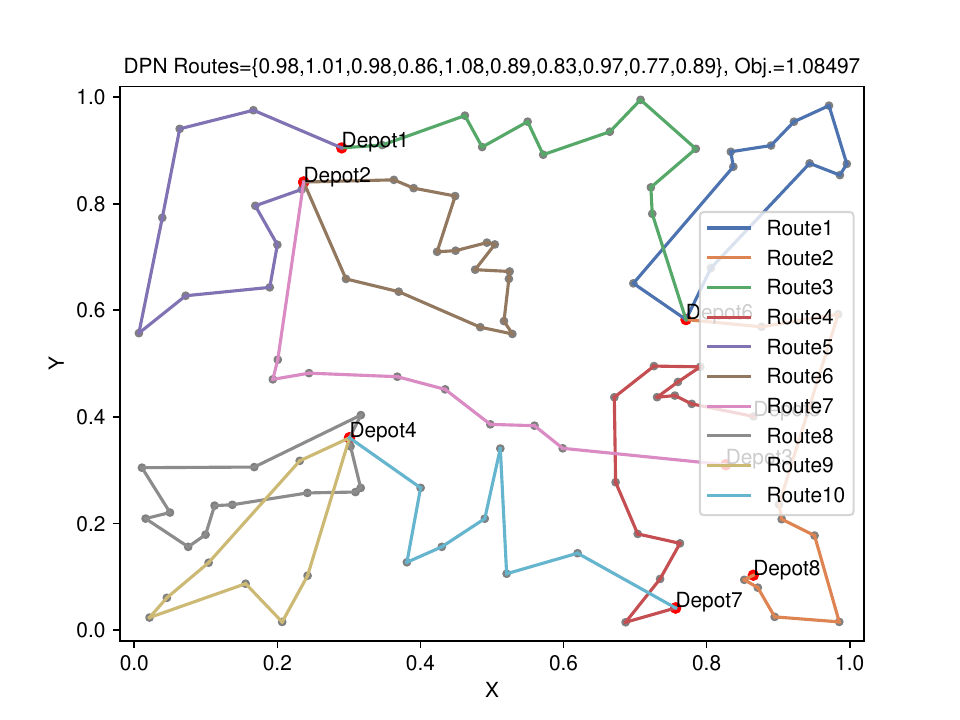}}  \quad\quad\quad   \subfigure[DPN-$\times8$aug-$\times16$per]{\includegraphics[width = 0.38\textwidth]{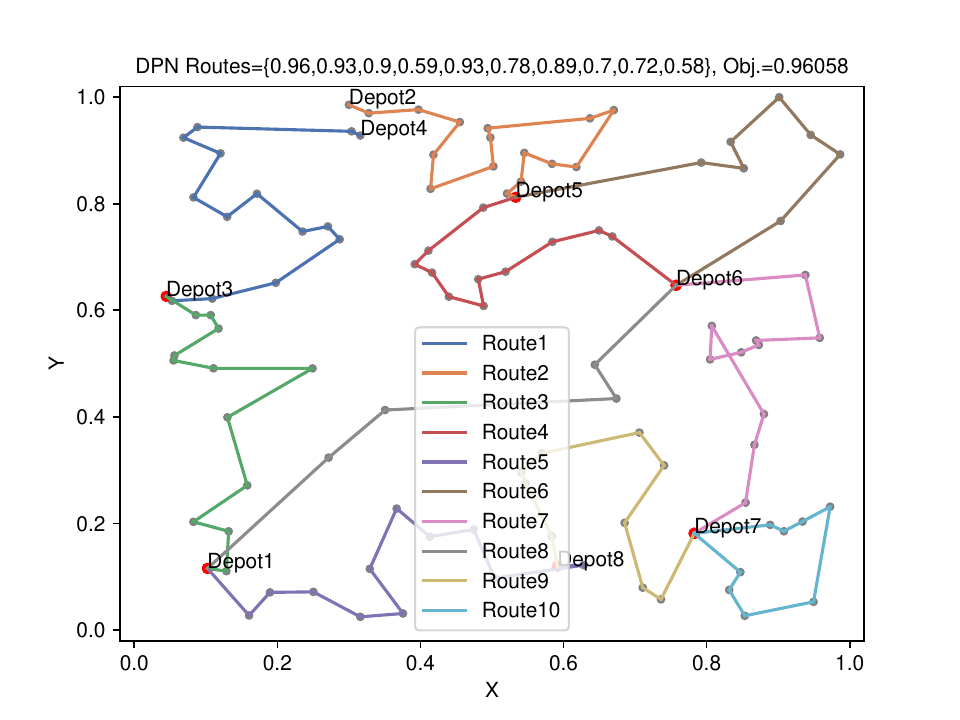}}
    \caption{Min-max FMDVRP instances ($D$=8,$M$=10) solving by ours DPN. Selected depots are highlighted by red.}
    \label{fig:fmdvrp-vis2}
\end{figure}

\begin{figure}[H]
    \centering
    \subfigure[OR-Tools*]{\includegraphics[width = 0.31\textwidth]{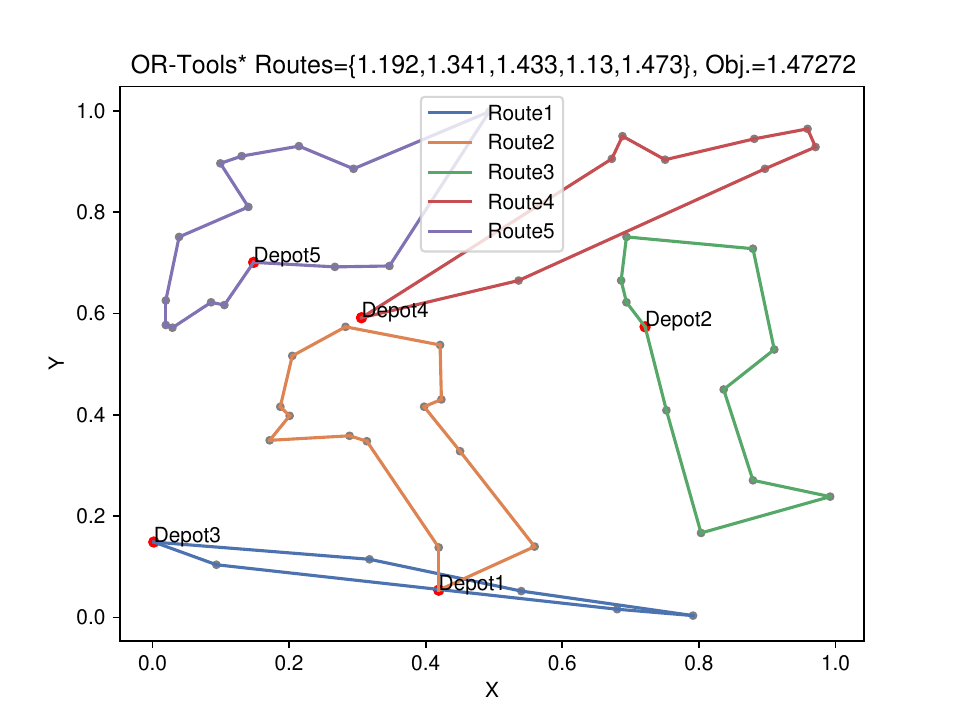}}    \subfigure[NCE*]{\includegraphics[width = 0.31\textwidth]{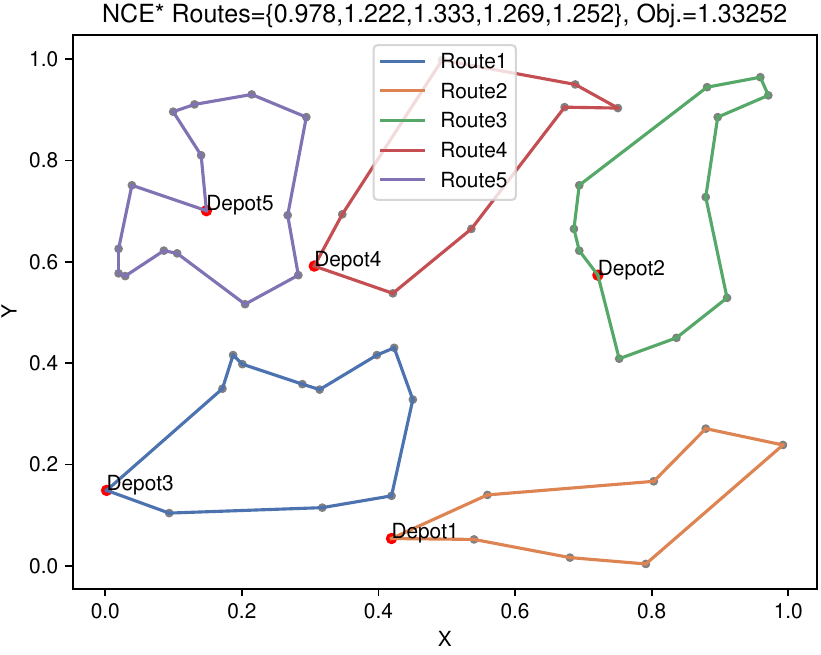}}    \subfigure[DPN-$\times8$aug-$\times16$per]{\includegraphics[width = 0.31\textwidth]{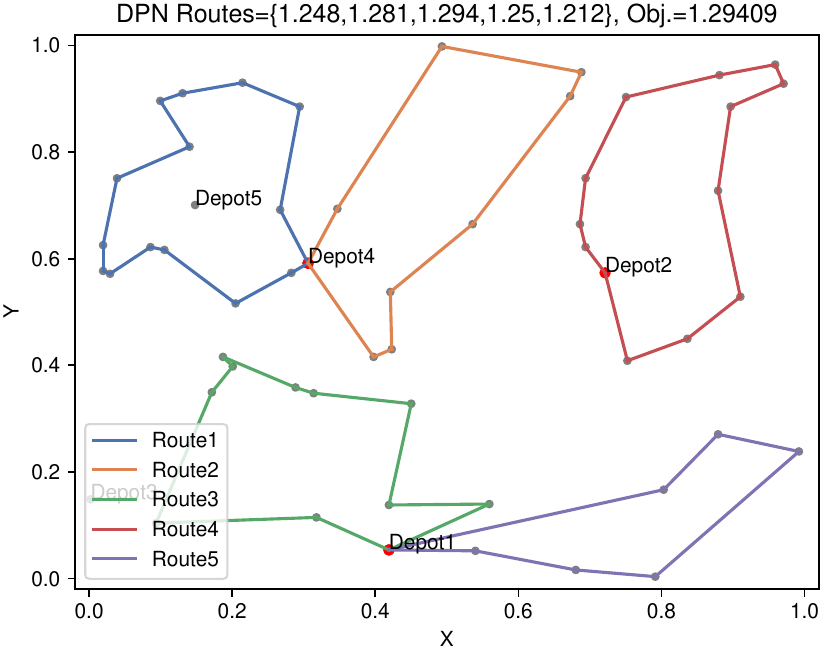}}
    \caption{Min-max MDVRP instances ($M$=5, $D$=5, $N$=50), solving by OR-Tools, NCE, and ours DPN. The solution of OR-Tools and NCE is reported in \citet{kim2022learning}. }
    \label{fig:mdvrp-rep}
\end{figure}

\begin{figure}[H]
    \centering
    \subfigure[OR-Tools*]{\includegraphics[width = 0.31\textwidth]{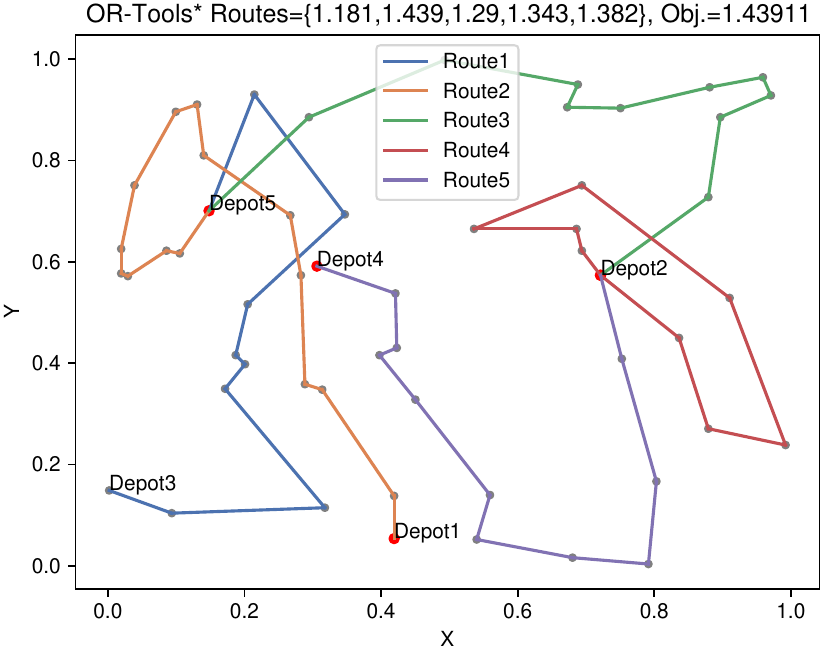}}    \subfigure[NCE*]{\includegraphics[width = 0.31\textwidth]{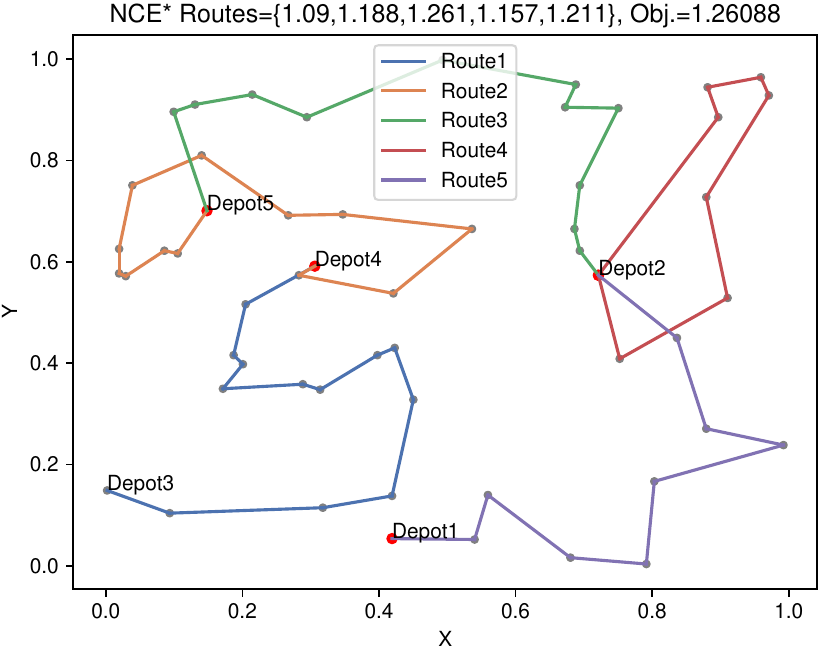}}    \subfigure[DPN-$\times8$aug-$\times16$per]{\includegraphics[width = 0.31\textwidth]{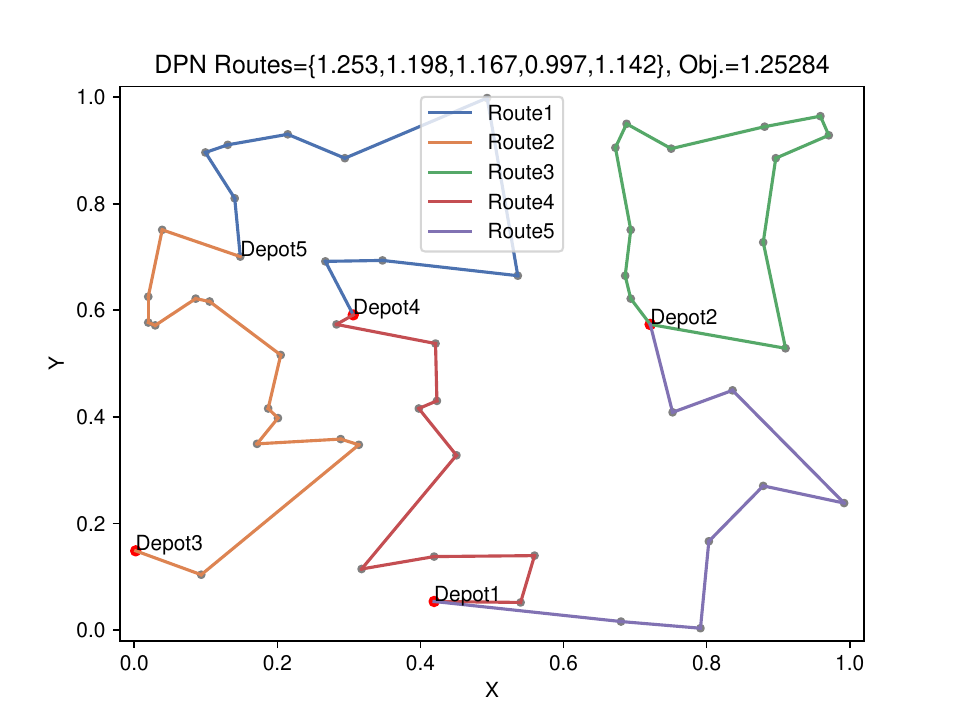}}
    \caption{Min-max FMDVRP instances ($M$=5, $D$=5, $N$=50), solving by OR-Tools, NCE, and ours DPN. The solution of OR-Tools and NCE is reported in \citet{kim2022learning}. }
    \label{fig:fmdvrp-rep}
\end{figure}

This section provides visualization of some random instances of min-max mTSP (Figure \ref{fig:mtsp-vis}, Figure \ref{fig:mtsp-vis2}), min-max mPDP (Figure \ref{fig:mpdp-vis}, Figure \ref{fig:mpdp-vis2}), and min-max MDVRP (Figure \ref{fig:mdvrp-vis}, Figure \ref{fig:mdvrp-vis2}, Figure \ref{fig:mdvrp-rep}), min-max FMDVRP (Figure \ref{fig:fmdvrp-vis}, Figure \ref{fig:fmdvrp-vis2}, Figure \ref{fig:fmdvrp-rep}) to display the advantage of the proposed DPN.


The solutions of OR-Tools and NCE in Figure \ref{fig:mdvrp-rep} and Figure \ref{fig:fmdvrp-rep} are reported in \citet{kim2022learning}. Due to their code being unavailable, we crawl the data and optimal solution through pixel coordinates. Therefore, the value of their objective function may be inconsistent. In these instances, DPN still outperforms other neural solvers.

\newpage

\section{Baseline \& License}\label{baseline}

The implement HGA algorithm is given in \url{https://github.com/Sasanm88/m-TSP}. For all uniform datasets with 100 instances, we set HGA to run 10 times. For the dataset with uniform depot location (adopted in Appendix \ref{ablationpe}) consisting of 10,000 instances and cross-distribution experiments (in Appendix \ref{distri-appn}), we only make HGA run once.

We implement the approximate min-max constraint and the pickup and delivery constraints for OR-Tools based on official tutorials \url{https://or-tools.github.io/docs/pdoc/ortools.html}. In solving min-max mTSP and min-max mPDP, we enable them to run 600s. For multi-depot min-max VRPs, due to various implementation alternations available, we use the results reported in \citet{kim2022learning}.

As to LKH3, We adopt the commonly used setting in the NCO methods \citep{kool2019attention} and implement the LKH3 in min-max mTSP based on the code from \url{https://github.com/wouterkool/attention-learn-to-route}. We do not set runtime limits, but instead, specify its maximal according to commonly used settings and set the MAX\_TRAILS parameter to 10,000, MAX\_CANDIDATES to 6, and RUNS to 10. So, the solution optimality of LKH3 in this paper may be different from the report results in \citet{son2023solving}.

As mentioned in Appendix \ref{datasets} in the supplementary File, in testing mTSP, we use the dataset provided by \citet{son2023solving}. Due to the learning methods ScheduleNet and NCE in mTSP are provided in \citet{son2023solving} with 2-digital results. So although they are not available, the results are accurate. However, with testing on a different dataset, the baseline results of min-max MDVRP and min-max FMDVRP in Table \ref{mdvrp} are not accurate.

For Equity-Transformer, we use the provided pre-trained model for experiments on 50-scale min-max VRPs and 200-scale min-max VRPs. We further train 100-scale models for the min-max mTSP100 and the min-max mPDP100 datasets for a fair comparison. Moreover, we train a min-max mPDP500 fine-tuned model based on the min-max mPDP100 model. In Table \ref{more}, we provide an ``Equity-Transformer-F-sample*'' version, which is the reported result in \citet{son2023solving} without illustrations about specific sample settings. 

We use the provided pre-trained model for every run with a greedy decoding setting for another open-source neural solver DAN \citep{cao2021dan}.

\begin{table}[H]
\centering
\setlength{\tabcolsep}{6mm}
\renewcommand\arraystretch{1.10}
% \caption{The average results on 1000 \textbf{Mixed 3-objective TSP} instances. The best results are in \textbf{bold}.}
% \usepackage{multirow}
\caption{A summary of licenses.}
\begin{tabular}{lll}
\hline\hline
Resources & Type    & License                              \\ \hline
HGA       & Code    & Available online                          \\
OR-Tools  & Code    & Apache License, Version 2.0          \\
LKH3      & Code    & Available for academic research use  \\
DAN        & Code    & MIT License                          \\
Equity-Transformer    & Code    & Available online                         \\
mTSP SetI    & Dataset & Available for any non-commercial use \\ 
mTSP Lib    & Dataset & Available for any non-commercial use \\ \hline\hline
\end{tabular}
\label{1}
\end{table}
The licenses for codes and datasets used in this work are listed in Table \ref{1}.

%%%%%%%%%%%%%%%%%%%%%%%%%%%%%%%%%%%%%%%%%%%%%%%%%%%%%%%%%%%%%%%%%%%%%%%%%%%%%%%
%%%%%%%%%%%%%%%%%%%%%%%%%%%%%%%%%%%%%%%%%%%%%%%%%%%%%%%%%%%%%%%%%%%%%%%%%%%%%%%

\end{document}